\documentclass{applemlr}

\usepackage{amsmath}
\usepackage{enumerate}
\usepackage{algorithm}
\usepackage{algpseudocode}
\usepackage{amsfonts}
\usepackage{amsthm}
\usepackage{cleveref}
\usepackage{diagbox}
\usepackage{colortbl}
\usepackage{amssymb}
\usepackage{xspace}
\usepackage{wrapfig}
\usepackage{adjustbox}
\usepackage{tabularx}
\usepackage{booktabs}
\usepackage{mathtools}
\usepackage{tikz}
\usepackage{enumitem}
\usepackage{silence}
\usepackage{dsfont}
\usepackage[table]{xcolor}
\usepackage[dvipsnames]{xcolor}
\usepackage{multirow}
\usepackage{makecell}
\usepackage{xfakebold}

\usepackage{amsmath,amsfonts,bm}

\def\eqref#1{equation~\ref{#1}}

\def\1{\bm{1}}

\DeclareMathAlphabet{\mathsfit}{\encodingdefault}{\sfdefault}{m}{sl}
\SetMathAlphabet{\mathsfit}{bold}{\encodingdefault}{\sfdefault}{bx}{n}

\definecolor{textgray}{HTML}{6E6E73}
\usetikzlibrary{positioning, calc}
\usetikzlibrary{decorations.pathmorphing}

\makeatletter
\patchcmd{\wrong@fontshape}{\@gobbletwo}{}{}{}
\makeatother
\numberwithin{equation}{section}
\makeatletter
\AtBeginDocument{
  \urlstyle{sf}
  
}
\makeatother

\definecolor{light}{RGB}{125, 125, 125}
\crefname{tcb@cnt@pbox}{code}{code}
\Crefname{tcb@cnt@pbox}{Code}{Code}
\crefname{assumption}{assumption}{assumption}
\Crefname{assumption}{Assumption}{Assumptions}

\newtcolorbox[auto counter]{pbox}[2][]{
  colback=white,
  title=Code~\thetcbcounter: #2,
  #1,fonttitle=\sffamily,
  fontupper=\sffamily,
  arc=2pt,
  colframe=bgcolor,
  coltitle=fgcolor,
  colbacktitle=bgcolor,
  toptitle=0.25cm,
  bottomtitle=0.125cm
}

\makeatletter
\newcommand\applefootnote[1]{%
  \begingroup
  \renewcommand\thefootnote{}%
  \renewcommand\@makefntext[1]{\noindent##1}%
  \footnote{#1}%
  \addtocounter{footnote}{-1}%
  \endgroup
}
\makeatother

\definecolor{cverbbg}{gray}{0.90}

\usepackage{eccvabbrv}
\usepackage{graphicx}

\usepackage{hyperref}

\usepackage{orcidlink}

\usepackage{url}
\usepackage[most]{tcolorbox}
\usepackage{times}
\usepackage{latexsym}
\usepackage{amsmath}
\usepackage{booktabs}
\usepackage{subcaption}
\usepackage{wrapfig}
\usepackage{colortbl}

\usepackage[T1]{fontenc}
\usepackage [utf8]{inputenc}

\usepackage{microtype}

\usepackage{inconsolata}
\usepackage{makecell}
\usepackage{multirow}
\usepackage{multicol}
\usepackage{xspace}
\usepackage{graphicx}
\usepackage{amsmath, amsfonts, amssymb}
\usepackage{xcolor}  
\usepackage{pifont} 
\usepackage{nicefrac}
\usepackage{fvextra}
\definecolor{myorange}{RGB}{255,165,0}
\definecolor{mybrown}{RGB}{150,75,0} 
\definecolor{myblue}{RGB}{30, 144, 255}
\definecolor{myred}{RGB}{255,0,0}
\definecolor{mypurple}{RGB}{128,0,128}
\newtcblisting{showcase}[1][]{
  enhanced,
  arc=0em,
  boxrule=.5pt,
  listing only,
  listing options={
    language=mycase,
    upquote=true,
    basicstyle=\showcasesize \ttfamily \setlength{\baselineskip}{1.1\baselineskip},
    breaklines=true,
    breakindent=8pt,
    xleftmargin=-8pt,
    xrightmargin=-8pt,
    aboveskip=-4pt,
    belowskip=-4pt,
    columns=fullflexible,
    escapeinside={|}{|},
  },
  colback=white,
  colframe=gray,
  colbacktitle=gray!10,        
  coltitle=black,              
  attach boxed title to top center={yshift=-3mm},
  #1
}

\lstdefinelanguage{myjson}{
    basicstyle=\ttfamily, 
    keywordstyle=\color{myorange}\bfseries, 
    morekeywords={parameters,returns,exceptions}, 
    literate=%
      {:}{{{\color{mybrown}{:}}}}1
      {,}{{{\color{mybrown}{,}}}}1
      {"}{{{\color{mybrown}{"}}}}1
      {\{}{{{\color{mycyan}{\{}}}}1
      {\}}{{{\color{mycyan}{\}}}}}1
      {[}{{{\color{mycyan}{[}}}}1
      {]}{{{\color{mycyan}{]}}}}1,
}

\newtcblisting[use counter=jsoncounter]{showjson}[1][]{
  enhanced,
  arc=0em,
  boxrule=.5pt,
  listing only,
  listing options={
    language=myjson,
    upquote=true,
    basicstyle=\tiny \ttfamily \setlength{\baselineskip}{1.1\baselineskip},
    breaklines=true,
    breakindent=8pt,
    xleftmargin=-8pt,
    xrightmargin=-8pt,
    aboveskip=-4pt,
    belowskip=-4pt,
    columns=fullflexible,
    escapeinside={|}{|},
  },
  colback=white,
  colframe=gray,
  colbacktitle=gray!10,        
  coltitle=black,              
  attach boxed title to top center={yshift=-3mm},
  #1
}

\lstdefinelanguage{prompt}{
    basicstyle=\scriptsize\ttfamily, 
    morestring = [s]{[}{]},
    stringstyle = \color{cyan},
    showstringspaces = false,
    morecomment = [f][\color{magenta}][0]{\#},
    moredelim = [s][\color{myred}]{**}{**},
    moredelim = [s][\color{myorange}]{\{}{\}},
    moredelim = [s][\color{mybrown}]{\{\{}{\}\}},
    moredelim = [s][\color{mybrown}]{'''}{'''},
    moredelim = [s][\color{mycyan}]{<}{>},
    moredelim = [s][\color{mybrown}]{"}{"},
    literate = %
        {:}{{\textcolor{mybrown}{:}}}1
        {-\ }{{\textcolor{myblue}{-\ }}}2
        {*\ }{{\textcolor{myblue}{*\ }}}2
        {0.\ }{{\textcolor{mypurple}{0.\ }}}3
        {1.\ }{{\textcolor{mypurple}{1.\ }}}3
        {2.\ }{{\textcolor{mypurple}{2.\ }}}3
        {3.\ }{{\textcolor{mypurple}{3.\ }}}3
        {4.\ }{{\textcolor{mypurple}{4.\ }}}3
        {5.\ }{{\textcolor{mypurple}{5.\ }}}3
        {6.\ }{{\textcolor{mypurple}{6.\ }}}3
        {7.\ }{{\textcolor{mypurple}{7.\ }}}3
        {8.\ }{{\textcolor{mypurple}{8.\ }}}3
        {9.\ }{{\textcolor{mypurple}{9.\ }}}3
        {\ \ a.\ }{{\textcolor{mypurple}{\ \ a.\ }}}5
        {\ \ b.\ }{{\textcolor{mypurple}{\ \ b.\ }}}5
        {\ \ c.\ }{{\textcolor{mypurple}{\ \ c.\ }}}5
        {\ \ d.\ }{{\textcolor{mypurple}{\ \ d.\ }}}5
        {\ \ e.\ }{{\textcolor{mypurple}{\ \ e.\ }}}5
        {\ \ f.\ }{{\textcolor{mypurple}{\ \ f.\ }}}5
        {\ \ g.\ }{{\textcolor{mypurple}{\ \ g.\ }}}5
        {\ \ h.\ }{{\textcolor{mypurple}{\ \ h.\ }}}5
        {\ I.\ }{{\textcolor{mypurple}{\ I.\ }}}4
        {\ II.\ }{{\textcolor{mypurple}{\ II.\ }}}5
        {\ III.\ }{{\textcolor{mypurple}{\ III.\ }}}6
        {\ IV.\ }{{\textcolor{mypurple}{\ IV.\ }}}5
        {\ V.\ }{{\textcolor{mypurple}{\ V.\ }}}4
}

\definecolor{vscKeyword}{HTML}{569CD6}   
\definecolor{vscParam}  {HTML}{DCDCAA}   
\definecolor{vscComment}{HTML}{6A9955}   
\definecolor{vscString} {HTML}{CE9178}   
\definecolor{vscNumber} {HTML}{B5CEA8}   
\definecolor{vscDefault}{HTML}{D4D4D4}   
\definecolor{vscBG}     {HTML}{1E1E1E}   

\usepackage{booktabs,tabularx,array}
\newcolumntype{L}[1]{>{\raggedright\arraybackslash}p{#1}}
\usepackage{ragged2e}
\usepackage{ltablex} 

\newtcblisting[use counter=promptcounter]{prompt}[1][]{
  enhanced,
  arc=0em,
  boxrule=.5pt,
  listing only,
  breakable,
  listing options={
    language=prompt,
    upquote=true,
    basicstyle=\scriptsize \ttfamily \setlength{\baselineskip}{1.1\baselineskip},
    breaklines=true,
    breakindent=8pt,
    xleftmargin=-8pt,
    xrightmargin=-8pt,
    aboveskip=-4pt,
    belowskip=-4pt,
    columns=fullflexible,
    escapeinside={|}{|},
  },
  colback=white,
  colframe=gray,
  colbacktitle=gray!5,        
  coltitle=black,              
  attach boxed title to top center={yshift=-3mm},
  #1
}
\usepackage[table]{xcolor}
\usepackage{xspace}
\newcommand{\cmark}{\textcolor{green!70!black}{\checkmark}}
\newcommand{\xmark}{\textcolor{red}{\ding{55}}}
\newcommand{\data}{\textsc{NarrativeTrack}\xspace}
\definecolor{tzBlueHeader}{RGB}{78,160,205}
\definecolor{tzBlueHeader2}{RGB}{105,185,225}
\definecolor{tzBlueBorder}{RGB}{115,190,225}
\definecolor{tzBlueFill}{RGB}{232,246,252}
\definecolor{rqBlueBorder}{HTML}{6AADE4}

\DeclareTextCommand{\textquotedbl}{OT1}{\char`\"}

\lstdefinestyle{jsonTiny}{
  basicstyle=\ttfamily\scriptsize,
  breaklines=true,
  breakindent=0pt,
  columns=fullflexible,
  keepspaces=true,
  showstringspaces=false,
  upquote=true,
  frame=none
}

\NewDocumentCommand{\ember}
{ mO{} }{\textcolor{purple}{\textsuperscript{\textit{Ember}}\textsf{\textbf{\small[#1]}}}}

\NewDocumentCommand{\brian}
{ mO{} }{\textcolor{orange}{\textsuperscript{\textit{Brian}}\textsf{\textbf{\small[#1]}}}}

\NewDocumentCommand{\kaixin}
{ mO{} }{\textcolor{red}{\textsuperscript{\textit{Kaixin}}\textsf{\textbf{\small[#1]}}}}

\NewDocumentCommand{\gargi}
{ mO{} }{\textcolor{blue}{\textsuperscript{\textit{Gargi}}\textsf{\textbf{\small[#1]}}}}

\NewDocumentCommand{\jinjin}
{ mO{} }{\textcolor{green}{\textsuperscript{\textit{Jinjin}}\textsf{\textbf{\small[#1]}}}}

\NewDocumentCommand{\zhenhailong}
{ mO{} }{\textcolor{cyan}{\textsuperscript{\textit{Zhenhailong}}\textsf{\textbf{\small[#1]}}}}
\usepackage{array}
\usepackage{tabularx}
\usepackage{booktabs}
\definecolor{cvprblue}{rgb}{0.21,0.49,0.74}

\title{NarrativeTrack: Evaluating Entity-Centric Reasoning for Narrative Understanding}

\author{%
 Hyeonjeong Ha$^{1,2\dagger}$, Jinjin Ge$^{1}$, Bo Feng$^{1}$, 
 Kaixin Ma$^{1}$, Gargi Chakraborty$^{1}$
 }
\affiliation{$^{1}$Apple, $^{2}$University of Illinois Urbana-Champaign
}

\abstract{
Multimodal large language models (MLLMs) have achieved impressive progress in vision-language reasoning, yet their ability to understand temporally unfolding narratives in videos remains largely underexplored. Narrative understanding requires more than recognizing isolated events: models must maintain coherent representations of who is doing what, when, and where across scene transitions and temporal gaps.
We introduce \data, the first benchmark to evaluate narrative understanding in MLLMs through fine-grained entity-centric reasoning. Unlike existing benchmarks limited to short clips or coarse scene-level semantics, we decompose videos into constituent entities and evaluate models using a Compositional Reasoning Progression (CRP), a structured framework that progressively increases narrative complexity across three dimensions: entity existence, entity changes, and entity ambiguity. This progression requires models to move beyond local perception to reasoning about entities' temporal persistence, state changes, and fine-grained perceptual disambiguation. To enable scalable benchmark construction, we develop a fully automated entity-centric pipeline that extracts temporally grounded entity representations and provides the foundation for CRP.
Evaluations of state-of-the-art MLLMs reveal that existing models struggle to maintain coherent entity representations under visual transitions and temporal dynamics. Open-source general-purpose MLLMs exhibit strong perceptual grounding but weak temporal continuity, while video-specialized MLLMs capture temporal context yet frequently hallucinate entities' contexts. These findings uncover a fundamental trade-off between perceptual grounding and temporal reasoning, indicating that narrative understanding emerges only from their integration. \data provides the first systematic framework to diagnose and advance temporally grounded narrative comprehension in MLLMs.}

\metadata[Code]{\url{https://github.com/apple/ml-NarrativeTrack}}
\metadata[Dataset]{\url{https://huggingface.co/datasets/hjha/NarrativeTrack}}
\metadata[Correspondence]{\sffamily \url{hh38@illinois.edu}, \url{{jinjin_ge, bfeng2, kaixin_ma, g_chakraborty}@apple.com}}

\begin{document}
\maketitle
\renewcommand{\thefootnote}{\fnsymbol{footnote}}

\footnotetext[2]{Work done during an internship at Apple.}

\section{Introduction}
Narrative understanding involves perceiving how entities evolve and interact over time to form coherent events, a fundamental aspect of human cognition. When watching a video, humans naturally build a mental model of the unfolding story by tracking who is present, what they are doing, and how their states change across scenes~\citep{zacks2007event}. This process relies on maintaining entity representations: temporally grounded structures that bind an entity's identity and state across time, enabling coherent reasoning even under occlusion, viewpoint changes, or scene transitions. Entities thus serve as the basic units of narrative structure, organizing events into meaningful temporal and causal relationships.

\begin{figure}[t]
     \centering
     \begin{subfigure}[b]{0.49\textwidth}
         \centering
         \includegraphics[width=\textwidth]{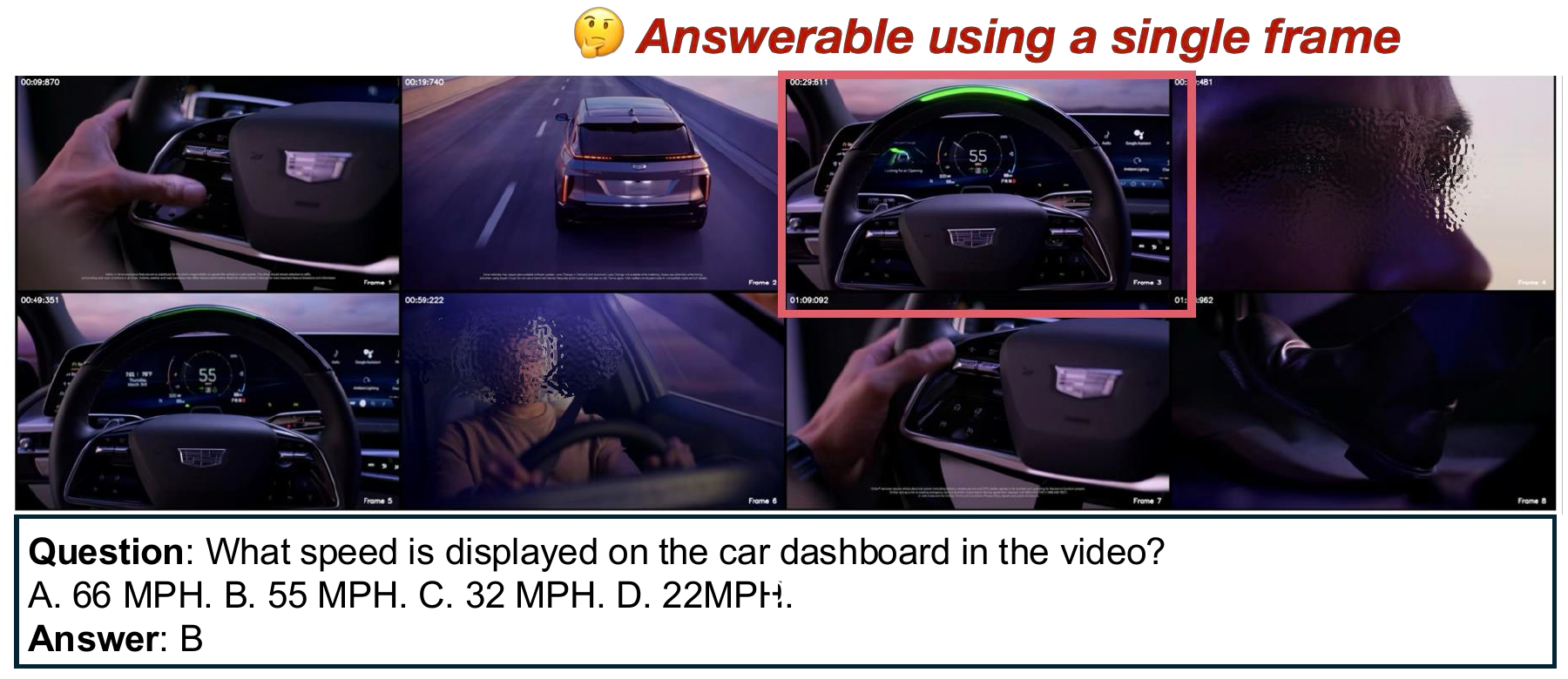}
         \caption{Video-MME~\citep{fu2025video}}
         \label{fig:ex_videomme}
     \end{subfigure}
     \hfill
     \begin{subfigure}[b]{0.49\textwidth}
         \centering
         \includegraphics[width=\textwidth]{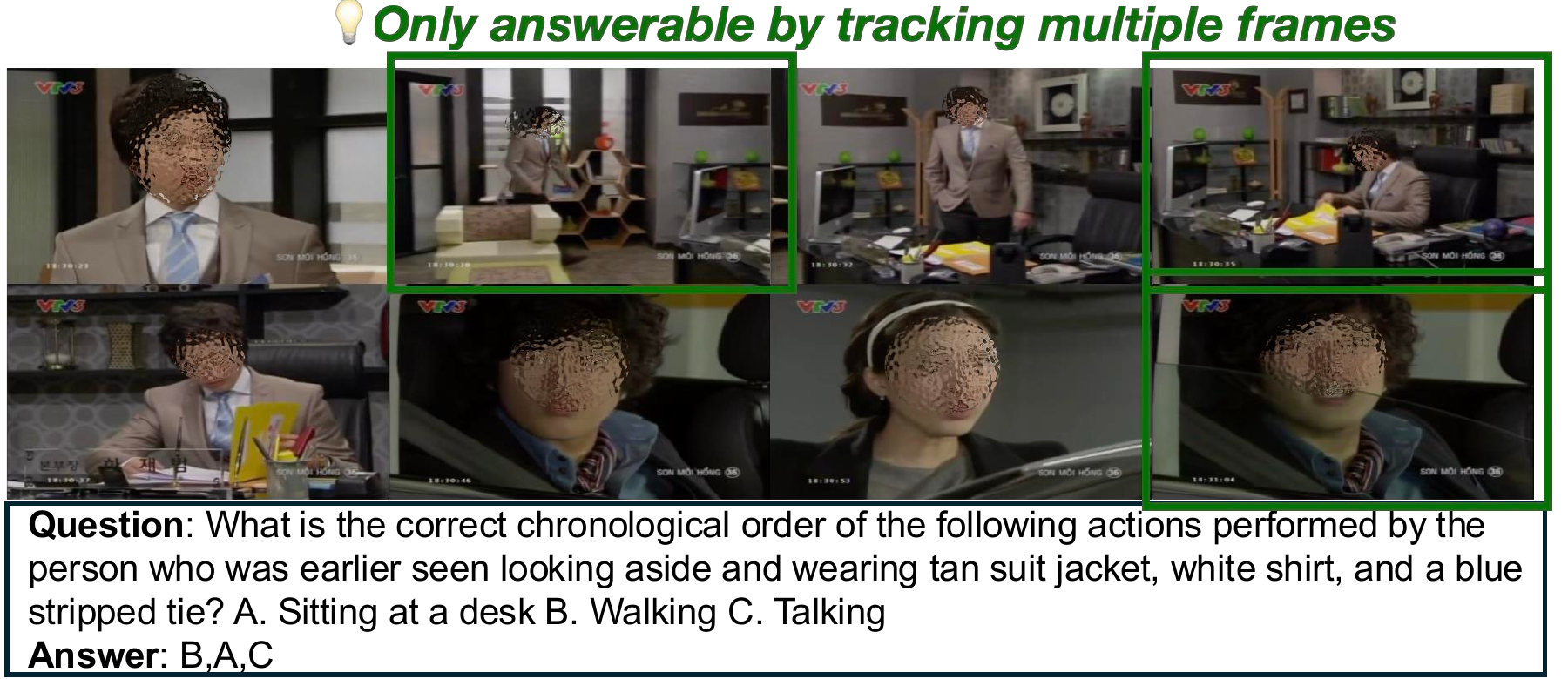}
         \caption{\data (Ours)}
         \label{fig:ex_ours}
     \end{subfigure}
     
    \caption{\small \textbf{Examples of existing benchmark and \data.} While existing benchmarks can often be answered from a single frame, ours requires reasoning by tracking entities over time. 
    }
    \label{fig:example_comparison}
\end{figure}

Despite remarkable advances in multimodal large language models (MLLMs) across both static~\citep{wang2024qwen2, liu2023visual, zhu2025internvl3, tsimpoukelli2021multimodal} and temporally evolving modalities~\citep{videollava, videochat2, maaz2023videochatgpt, cheng2024videollama, zhang2024llavanextvideo}, their entity-centric reasoning ability for narrative understanding remains largely unexamined. Existing benchmarks primarily assess either local recognition (e.g., object or action recognition) or global summarization (e.g., story-level understanding), leaving a critical gap in assessing whether models track entities consistently throughout a narrative. Datasets incorporating temporal information typically use short clips with minimal scene variation, emphasizing localized semantics~\citep{MSRVTT-QA, yu2019activitynet, fu2025video, xiao2021next, mangalam2023egoschema} (Fig.~\ref{fig:ex_videomme}). In contrast, long-form benchmarks capture coarse global context~\citep{wang2024lvbench, wu2024longvideobench}, but rarely test long-range temporal dependencies where entities evolve and interact across disjoint scenes (Fig.~\ref{fig:existing_examples}). Consequently, many tasks can be solved using static visual cues or language priors~\citep{feng2025breaking}, without genuine temporal reasoning, while heavy reliance on manual annotations further limits scalability and diagnostic granularity. 

To address these limitations, we introduce \data, a novel benchmark designed to evaluate narrative understanding in MLLMs through an entity-centric, bottom-up formulation. Rather than assessing video understanding from top-down summaries or scene-level semantics, we decompose each video into its constituent entities, which serve as the fundamental building blocks of events, and evaluate how models reason about their temporal continuity and evolution. 
This design is grounded in cognitive and narratological theory~\citep{zacks2007event, cutting2016narrative}: narrative comprehension emerges from maintaining entity continuity and interpreting their evolving actions and contextual roles. Evaluating whether models can track how entities persist, change, and become confusable directly probes the ability to construct temporally grounded narrative structure.
 
We formalize this by defining a \emph{Compositional Reasoning Progression} (CRP), which systematically increases narrative complexity across three interdependent reasoning dimensions:
(1) \textbf{entity existence} tests basic temporal continuity, evaluating whether models can track entities persistently across time; (2) \textbf{entity changes} evaluates reasoning about evolving actions, scenes, and outfits that signal narrative development~\citep{feng2025narrlv, feng2024tc}; (3) \textbf{entity ambiguity} introduces visually similar entities to challenge fine-grained perceptual disambiguation under temporal dynamics. This progression transforms narrative reasoning from single-skill testing into a compositional diagnostic of entity-centric reasoning, enabling a systematic evaluation of narrative understanding. To enable scalable construction of the entity representations that underpin CRP, we further present a \emph{fully automated entity-centric pipeline} that extracts temporally grounded trajectories augmented with fine-grained attributes directly from raw videos without human supervision.


We evaluate a diverse suite of state-of-the-art MLLMs on \data, spanning open-source general-purpose, open-source video-specialized, and proprietary models. Gemini-2.5-Pro achieves the highest average accuracy of 83.80\%, while open-source models exhibit substantial gaps. Notably, general-purpose MLLMs outperform video-specialized ones (e.g., Qwen2.5-VL-32B: 56.96\% vs. Video-LLaMA2-72B: 49.70\%), revealing a fundamental trade-off between perceptual grounding and temporal reasoning. General-purpose MLLMs show strong perceptual grounding, accurately recognizing static visual cues, yet frequently fail to maintain temporal consistency. Conversely, video-specialized MLLMs capture temporal continuity more reliably but lack perceptual robustness, often hallucinating under entity ambiguity or appearance changes. Further analysis shows that neither scaling model size nor increasing frame density improves narrative understanding, and it also exposes weaknesses in reasoning about backward temporal scenarios, highlighting a persistent difficulty in maintaining coherent entity representations. We leave architectural innovation, such as bidirectional temporal modeling and entity-centric training objectives, as future work. Overall, our results indicate that narrative understanding is a compositional capability, emerging from the integration of temporal reasoning and perceptual precision. By centering evaluation on fine-grained entity tracking, \data offers the first systematic framework to diagnose how and where MLLMs fail to maintain coherent, temporally grounded narrative structure.

\section{Related Work}
\subsection{Multimodal Large Language Models (MLLMs)}
MLLMs augment large language models with visual encoders, enabling joint reasoning over text and images for tasks such as chart understanding and visual question answering~\citep{wang2024qwen2, liu2023visual, zhu2025internvl3, wang2025perceptionawarepolicyoptimizationmultimodal, blume2025partonomylargemultimodalmodels, ha2025mm}. Recent advances have scaled these models toward unified input-output representations, long-context reasoning, and cross-modal generalization~\citep{zheng2025deepeyes, wang2024emu3, team2024chameleon, chen2024solo, dalal2025constructive}, bringing them closer to general-purpose visual-language understanding. Building on these developments, recent work has extended MLLMs to the video domain, enabling them to process sequential visual inputs and reason over temporally evolving scenes~\citep{videollava, cheng2024videollama, videochat2, maaz2023videochatgpt, cho2025perceptionlmopenaccessdatamodels}. However, most MLLMs remain optimized for static image understanding. Their visual encoders typically compress spatial and temporal information into coarse global embeddings, sacrificing fine-grained perceptual detail~\citep{li2024seed, kim2024finer}. Consequently, while they excel at high-level semantic alignment~\citep{jia2021scaling,radford2021learning}, they struggle to maintain consistent reasoning about individual entities or events over time. Moreover, how current MLLMs ground their responses in fine-grained elements such as entities remains underexplored, leaving a critical gap in understanding multimodal reasoning.

\subsection{Video Understanding Benchmarks}

Existing video understanding benchmarks evaluate MLLMs from diverse perspectives, including global semantic comprehension~\citep{MSRVTT-QA, yu2019activitynet}, causal and temporal reasoning~\citep{xiao2021next, videochat2}, action and event recognition~\citep{fu2025video}, spatial reasoning~\citep{mangalam2023egoschema}, and long-video understanding~\citep{wu2024longvideobench, li2024llamavid, song2024moviechat}. While each benchmark targets specific reasoning skills, recent work~\citep{feng2025breaking} has shown that many can be solved without genuine temporal reasoning. In many cases, model performance remains largely unchanged when frame order is shuffled, suggesting reliance on language priors or static visual cues rather than true temporal reasoning. These findings raise a concern that current evaluations often measure recognition of isolated events rather than the ability to maintain coherent spatio-temporal representations over time. More recently, compositional reasoning in videos has been studied through structured evaluation protocols. VELOCITI~\citep{saravanan2025velociti} introduces a strict video-language entailment, isolating agent–action binding errors via carefully constructed positive and negative captions. This design provides a controlled testbed for event-level compositional grounding. However, its scope remains temporally localized: entities are typically continuously visible within the clip, and the primary challenge lies in correctly associating agents with actions within short temporal windows. As such, the reasoning required is largely confined to within-clip binding rather than cross-scene continuity.

In contrast, narrative understanding in long-form videos introduces a fundamentally different challenge: characters may disappear and reappear across scenes, undergo state changes, and participate in causally linked events separated by substantial temporal gaps~\citep{mangalam2023egoschema, zacks2007event, cutting2016narrative}. Our benchmark moves beyond short-horizon entailment to entity-centric question answering over long videos, requiring models to maintain persistent identity and state representations across discontinuities. By explicitly evaluating long-range entity continuity under scene transitions, entity's state evolutions, and temporal gaps, our benchmark advances video understanding beyond localized grounding and static frame reasoning.
\section{Evaluating VideoLLMs Beyond the Frame}

\paragraph{\textbf{Problem Statement.}} Each input video \(V\) is represented as a sequence of frames sampled at a fixed frame rate \(f\), producing \(N+1\) frames with timestamps \(\{t_0, t_1, \dots, t_N\}\), where \(t_j = j / f\) for \(j \in [0, N]\). These timestamps define the temporal axis over which visual events unfold. For each entity \(e_i\), we define a temporally grounded \textit{entity representation}: 
\begin{align}
\tau_{e_i} = \{ &(t_{i0}, b_{i0}, a_{i0}, s_{i0}, o_{i0}), (t_{i1}, b_{i1}, a_{i1}, s_{i1}, o_{i1}), \nonumber\\
&\dots, (t_{iM}, b_{iM}, a_{iM}, s_{iM}, o_{iM})\},
\end{align}
where \(t_{ij} \in [t_0, t_N]\) denotes a timestamp $t_j$ at which \(e_i\) appears, and \(M \leq N\) is the total number of such observations. At each timestamp $t_j$, the entity representation includes: (1) \(b_{ij}\), the bounding box of \(e_i\); (2) \(a_{ij}\), the action performed by the entity; (3) \(s_{ij}\), the scene context; and (4) \(o_{ij}\), the entity's outfit.
This structured representation captures both \textit{temporal dynamics} and \textit{perceptual context}, serving as the foundation for evaluating whether MLLM can reason consistently about entity representation, tracking their continuity, appearances, and narrative roles across scenes, to achieve coherent and fine-grained narrative understanding.

In this work, we focus exclusively on \textit{human entities}, as they are the primary agents in narrative videos and exhibit rich temporal variations in action and appearances. This makes humans particularly representative subjects for evaluating entity-centric reasoning across temporal and perceptual grounding. In contrast, many non-human or background entities exhibit limited visual variation, which can reduce evaluation to simple identity matching, while also introducing ambiguity that may confound the assessment of compositional reasoning. Restricting the scope to visually salient human subjects enables a more controlled and challenging testbed for rigorously evaluating models' ability to maintain coherent entity-level understanding over time. Extending the framework to non-human entities remains a promising direction for future work.

\subsection{Automated Entity-Centric Pipeline}

\begin{table}[t]
\centering
\resizebox{0.9\linewidth}{!}{
    \begin{tabular}{lcccccccccc}
    \toprule
    \textbf{Benchmarks} & \textbf{Len.(s)} & \textbf{\#QA Pairs} & \textbf{Anno.} & \textbf{BBox} & \textbf{Action} & \textbf{Scene} & \textbf{Outfit} & \textbf{Entity-Centric} \\
    \midrule
    AVA~\citep{gu2018ava}          &  900 & - & M &\cmark & \cmark  &    \xmark  & \xmark   & \xmark   \\
    TVQA~\citep{lei2018tvqa}            & 11.2  & 15,253  & M & \xmark  &    \xmark     & \xmark         & \xmark & \xmark \\
    MovieQA~\citep{tapaswi2016movieqa}          &  205.5 & 2,144 & M & \xmark & \xmark  & \xmark   &    \xmark  & \xmark   \\
    How2QA~\citep{li2020hero}           & 15.3  & 2,852  & M &\xmark & \xmark  &     \xmark   &   \xmark   & \xmark   \\
    NExT-QA~\citep{xiao2021next}          &  39.5  & 8,564   & A & \xmark & \xmark&   \xmark    &   \xmark   & \xmark   \\
    Video-MME~\citep{fu2025video}          &  1,017.9 & 2,700  & M &  \xmark & \xmark &  \xmark  &   \xmark   & \xmark   \\
    PerceptionTest~\citep{patraucean2023perception}  & 23.0 & 44,000 &  A \& M & \cmark & \cmark &  \xmark  &    \xmark  & \xmark   \\
    LongVideoBench~\citep{wu2024longvideobench}  & 473 & 6,678 &  M &\xmark & \xmark &  \xmark  &   \xmark   & \xmark   \\
    LVBench~\citep{wang2024lvbench}  & 4,101 & 1,549 &  M &\xmark & \xmark &  \xmark  &    \xmark  & \xmark   \\
    \midrule
    \textbf{\data} &  55.3 & 1,006 & A & \cmark& \cmark& \cmark& \cmark& \cmark \\
    \bottomrule
    \end{tabular}%
}
\caption{\small \textbf{Comparison between existing video understanding benchmarks and \data.} \textit{Len.} denotes the average duration of clips, and \textit{Anno.} indicates the annotation type, where \textit{A} denotes automated annotation and \textit{M} refers to manual annotation. \looseness=-1}
\label{tab:videoqa_benchmarks}
\end{table}

\begin{figure*}[t]
    \centering
    \includegraphics[width=\linewidth]{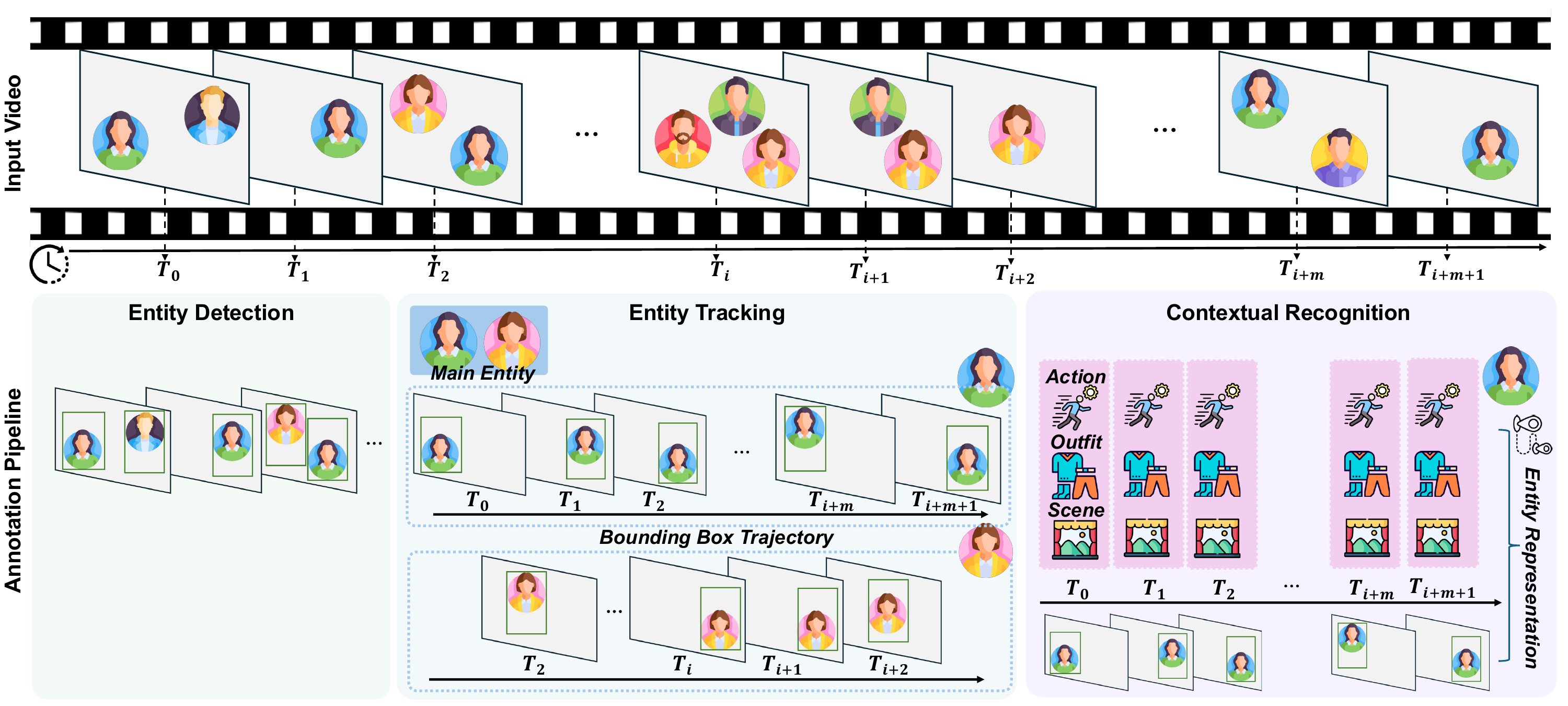}
    \caption{\textbf{Overview of Automated Entity-Centric Pipeline.} Our pipeline consists of three key stages: (1) Entity Detection, (2) Entity Tracking, (3) Contextual Recognition, extracting entity representations capturing bounding-box trajectories, actions, outfits, and scene contexts associated with the target entity from raw video without human supervision. 
    }
    \label{fig:concept_figure}
\end{figure*}

\label{sec:pipeline}
Building a benchmark for evaluating narrative understanding requires rich, temporally aligned representations that capture how entities evolve over time. Existing datasets lack such structured, entity-centric annotations, as manually labeling long, unconstrained videos is costly and often inconsistent (Table~\ref{tab:videoqa_benchmarks}). To overcome this limitation, 
we introduce a novel, fully automated pipeline that extracts structured entity representations \(\tau_{e_i}\) directly from raw videos without human supervision. The pipeline comprises three stages: \textit{entity detection}, \textit{entity tracking}, and \textit{contextual recognition} (Fig~\ref{fig:concept_figure}). 

\paragraph{\textbf{Entity Detection.}} We detect all visible entities in each frame using off-the-shelf multi-object detection models. Accurate detection is critical: missing or imprecise detections can fragment trajectories or merge distinct entities, cascading errors into tracking and contextual recognition. To improve reliability, we introduce an entity-centric ensemble detection strategy that combines predictions from Detectron2~\citep{wu2019detectron2} and Owlv2~\citep{owlv2} through spatial consensus rather than naive union. Let $\mathcal{B}^{(1)}$ and $\mathcal{B}^{(2)}$ denote the sets of bounding boxes predicted by the two detectors for a given frame. We compute pairwise IoU between boxes across detectors and consider two boxes to correspond to the same entity if IoU $\ge 0.5$, a standard threshold in detection matching and tracking that balances over-merging with duplicate retention. When two boxes meet this criterion, we keep only the higher-confidence score:
\[
\mathcal{B}_{\mathrm{final}} =
\mathcal{B}^{(1)} \cup \mathcal{B}^{(2)}
\setminus \{ b^{(2)}_j \mid \exists\, b^{(1)}_i, \ \mathrm{IoU}(b^{(1)}_i, b^{(2)}_j) \ge 0.5 \},
\]
This process preserves one representative box per entity while retaining unique detections from both models, yielding comprehensive yet non-redundant coverage of visible entities in each frame. The resulting per-frame detections are subsequently linked across time to form bounding box trajectories. Each trajectory is inherently spatio-temporal: it encodes the spatial localization of an entity within individual frames as well as its temporal evolution across frames, enabling consistent entity-level tracking throughout the video.

\paragraph{\textbf{Entity Tracking.}} Narrative understanding requires identifying \textit{who persists} over time. However, naively tracking every detected entity is inefficient and error-prone, as some may appear briefly or contribute little to the narrative. To focus on salient narrative participants, we identify main characters by clustering bounding-box embeddings extracted with state-of-the-art re-identification (ReID) models~\citep{zhou2019omni, ye2024transformer, niu2025chatreid}. We use OSNet-x1.0\footnote{https://kaiyangzhou.github.io/deep-person-reid/MODEL\_ZOO.html} as the ReID model. Each cluster corresponds to a distinct entity and is ranked by size, with the top four selected as main characters under the assumption that recurrent presence correlates with narrative centrality. Empirically, central characters tend to occupy larger spatial regions ($\geq$50\% of the frame height or width) and appear consistently over time. Across 100 sampled videos, the top-four clusters satisfy this criterion for at least 50\% of their instances, motivating our choice of four entities as a stable heuristic for capturing salient characters. To enhance identity consistency, we refine trajectories with face recognition for precise alignment and apply ensemble verification across multiple MLLMs with majority voting to suppress false identities. Detailed prompt is provided in Appendix \S\ref{appendix:mv_prompt}. This consensus-based strategy emulates human agreement, minimizing identity drift in a fully unsupervised setting. The resulting trajectories provide temporally consistent spatial localization for each target entity throughout the video. 

\paragraph{\textbf{Contextual Recognition.}}
Beyond spatial localization, narrative comprehension requires understanding \textit{what} each entity is doing, \textit{how} and \textit{when} it appears, and \textit{where} it is situated. For each entity trajectory, we augment every timestep \(t_{ij}\) with contextual attributes: \textit{actions} \(a_{ij}\), \textit{outfits} \(o_{ij}\), and \textit{scenes} \(s_{ij}\). 
We use Gemini-2.5-Pro~\citep{comanici2025gemini} to infer these contextual attributes (prompt is provided in \S\ref{appendix:recog_prompt}), highlighting the target entity in overlaid clips to preserve context and ensure the model attends to the correct individual. Attributes are annotated per segment in which the entity appears in clips, and these predictions populate \(\tau_{e_i}\), yielding a structured, temporally aligned representation of how each entity evolves throughout the narrative. To support QA generation for entity changes and entity ambiguity dimensions, Gemini-2.5-Pro is used to identify videos with significant attribute changes or visually similar entities based on predicted attributes.

\paragraph{\textbf{Pipeline Evaluation.}}
We evaluate the quality of our detection and tracking pipeline using the AVA dataset~\citep{gu2018ava}, which provides frame-level bounding boxes $\mathcal{G}=\{g_1, g_2, \dots, g_{N_G}\}$ but lacks entity identities. Since the correspondence between ground-truth and predicted boxes is unknown, each predicted box $p_i \in \mathcal{P}=\{p_1, p_2, \dots, p_{N_P}\}$ at frame $t$ is greedily matched to a ground-truth box: 
\[
g^*(p_i) = \operatorname*{arg\,max}_{g \in \mathcal{G} \setminus \mathcal{G}_{\mathrm{matched}}} \mathrm{IoU}(p_i, g)
\]
and if $\mathrm{IoU}(p_i,g^*(p_i)) \ge 0.5$, we record the match, updating $\mathcal{M}$ and $\mathcal{G}_{\mathrm{matched}}$ as: $\mathcal{M} \leftarrow \mathcal{M} \cup \{(p_i, g^*(p_i))\}$ and $\mathcal{G}_{\mathrm{matched}} \leftarrow \mathcal{G}_{\mathrm{matched}} \cup \{g^*(p_i)\}$. The recall is then computed as $\mathrm{Recall} = |\mathcal{M}|/|\mathcal{G}|$. Our ensemble detection improves recall from 0.780 (Detectron2 alone) and 0.801 (OWLv2 alone) to 0.848, confirming that spatial consensus effectively reconciles missed or inconsistent detections. 

To assess the reliability of our ensemble-based tracking verification, human experts from the authors label each tracked entity as \textit{kept} if it consistently matches the same entity across frames, or \textit{deleted} if an incorrect identity is assigned. We compare human-majority vote labels with the predictions of our model-based majority voting method on randomly sampled AVA video clips comprising 1,108 detections. Notably, two sources agree on 96.08\% of cases, indicating that our consensus approach reliably filters erroneous tracks and accurately preserves coherent identities. This confirms that our entity-centric pipeline produces high-quality tracking results in a fully automated manner.

\subsection{Compositional Reasoning Progression}
\begin{figure*}[t]
    \centering
    \begin{subfigure}[t]{0.52\linewidth}
        \centering
        \includegraphics[width=\linewidth]{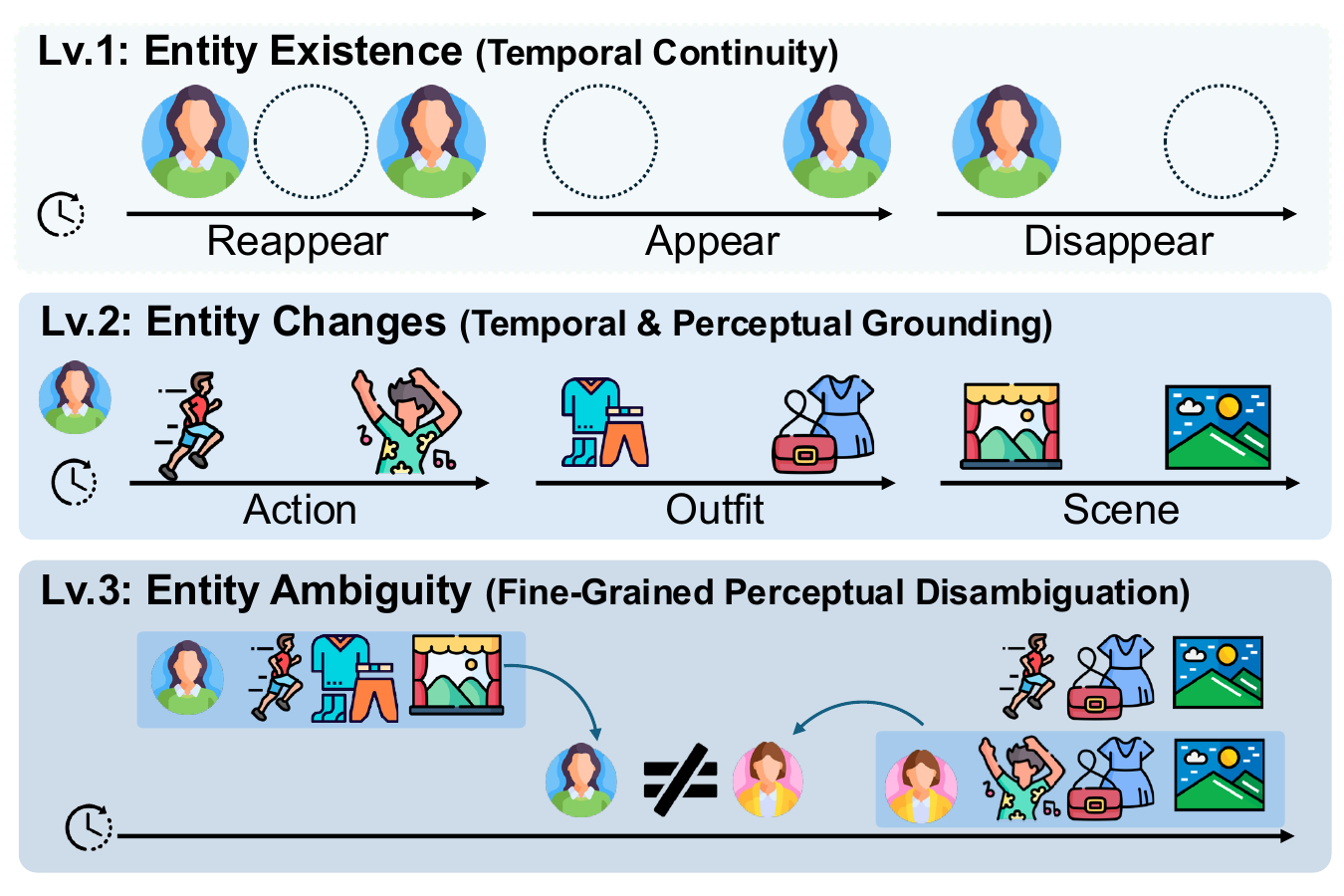}
        \caption{Compositional Reasoning Progression.}
        \label{fig:question_dist}
    \end{subfigure}
    \hfill
    \begin{subfigure}[t]{0.47\linewidth}
        \centering
        \includegraphics[width=\linewidth]{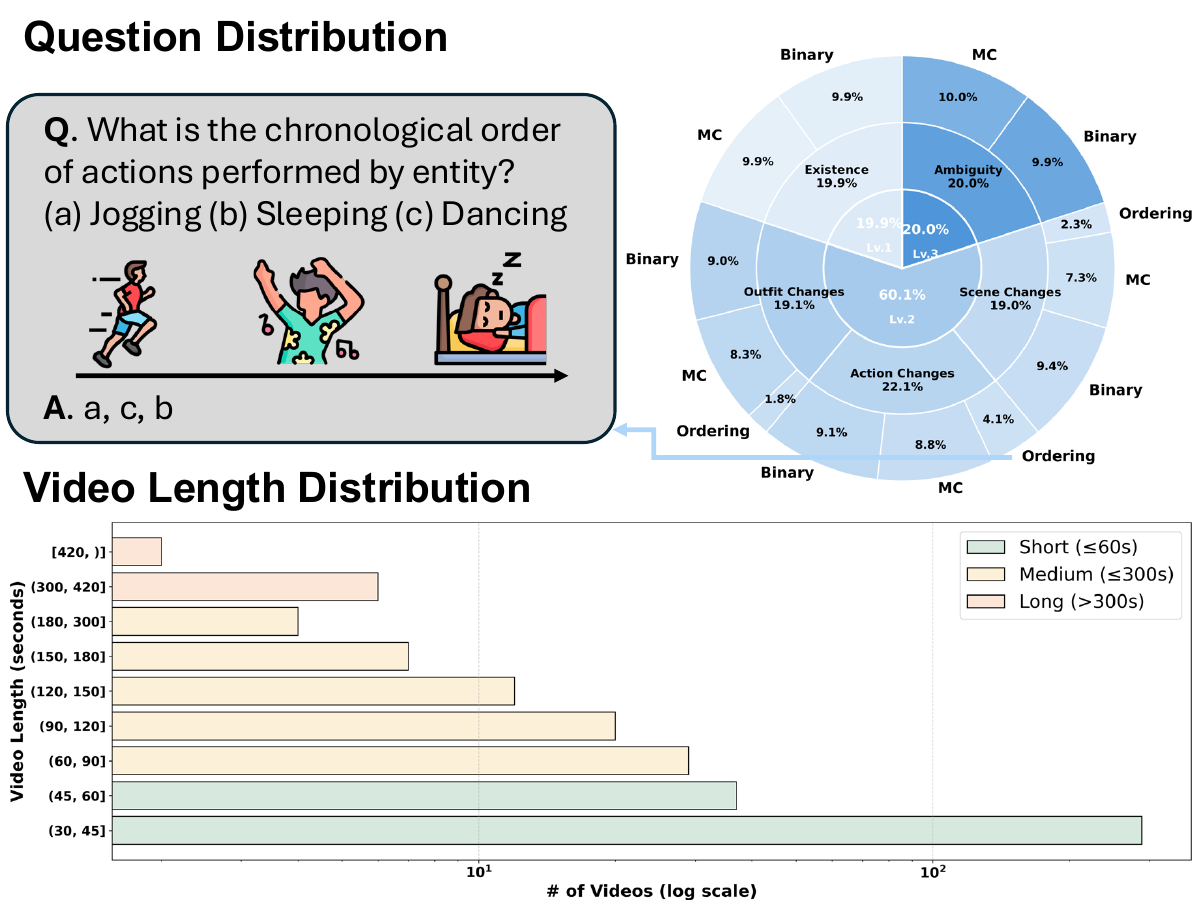}
        \caption{Data Statistics of \data.}
        \label{fig:video_length}
    \end{subfigure}
    \caption{\textbf{Overview of \data.} (a) Our benchmark is grounded in Compositional Reasoning Progression that introduces three levels of increasing complexity in entity-centric reasoning: entity existence, entity changes, and entity ambiguity. (b) The benchmark covers three question types: binary, multiple-choice (MC), and ordering, with diverse temporal scales from short to long (average of 55.3 seconds).}
    \label{fig:combined}
    \label{fig:benchmark}
\end{figure*}

To systematically evaluate narrative understanding, we introduce Compositional Reasoning Progression (CRP), an evaluation framework that decomposes entity-centric reasoning into progressively more complex dimensions (Fig.~\ref{fig:question_dist}). Each dimension isolates a distinct capability, enabling fine-grained analysis of model behavior. 

\textbf{Entity existence} evaluates \textit{temporal continuity}: the ability to track an entity's presence across its appearances, disappearances, and reappearances. This dimension isolates the most fundamental temporal requirement of maintaining a consistent identity over time. Failures indicate breakdowns in long-range temporal tracking. \textbf{Entity changes} assess whether a model can detect and ground state transitions of a target entity. Beyond continuity, the model must integrate dynamic visual and contextual cues. We decompose this dimension into: (i) \textbf{Action changes}, testing \textit{temporal grounding} of dynamic events; (ii) \textbf{Outfit changes}, probing \textit{fine-grained visual grounding} of perceptual transitions; (iii) \textbf{Scene changes}, evaluating \textit{spatial-contextual grounding} by linking entity with environmental context. The failures here reflect a weak grounding of evolving visual or contextual signals. \textbf{Entity ambiguity} introduces visually similar entities, requiring joint \textit{temporal reasoning and fine-grained perceptual disambiguation} to ensure reliable entity tracking. Errors reveal weak compositional reasoning.

Overall, CRP progresses from entity persistence to grounded state transitions, to joint temporal–perceptual disambiguation. This design enables precise diagnosis of whether failures arise from disrupted temporal continuity, weak grounding of evolving attributes, or confusion among similar entities.

\subsection{QA Generation}
\label{sec:qa_gen}
\begin{figure}[t]
    \centering
    \includegraphics[width=\linewidth]{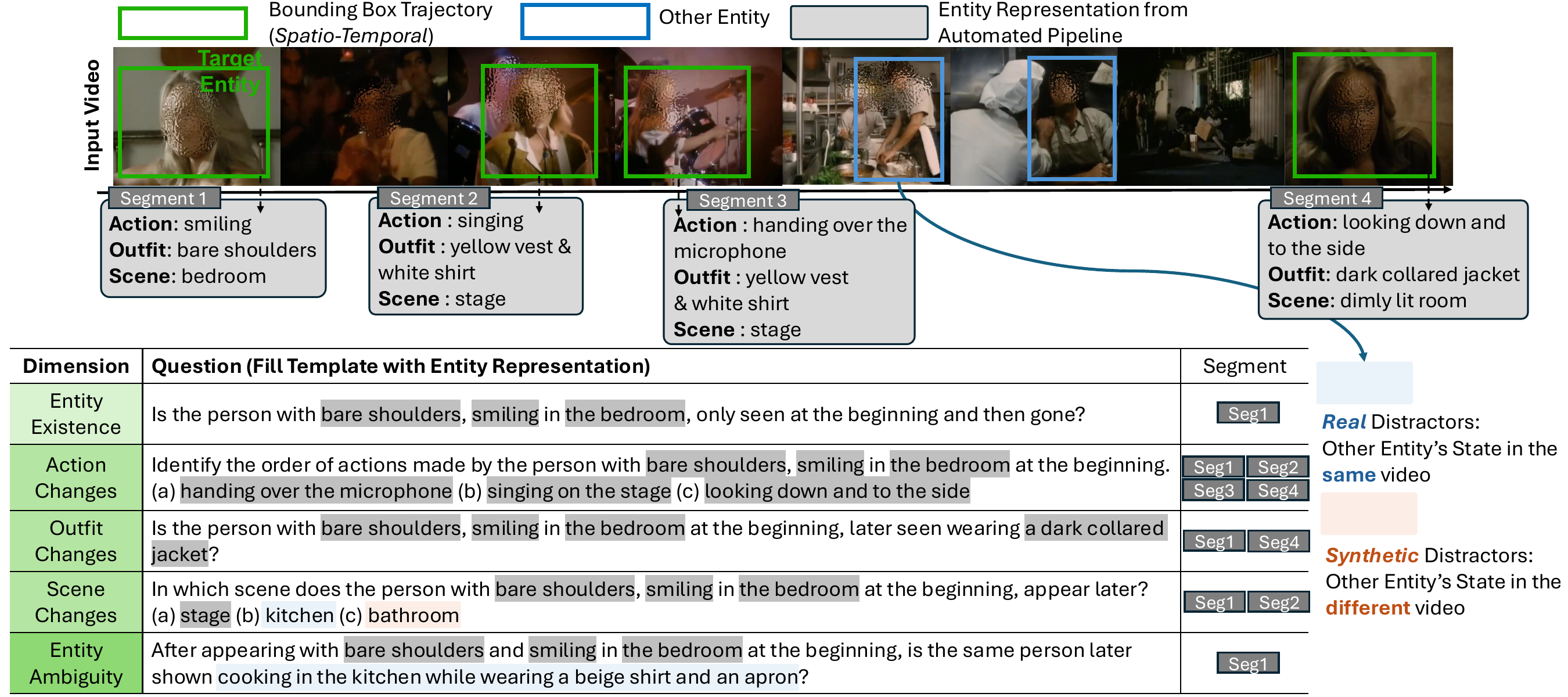}
    \caption{\textbf{QA Generation \& Qualitative Examples.} QA pairs are constructed by instantiating CRP-aligned reasoning templates using extracted entity representations. Distractors include real ones from other entities in the same video and synthetic ones from different videos.}
    \label{fig:qa_generation}
\end{figure}

Building on the structured entity-centric representations produced by our automated pipeline, we construct question-answer (QA) pairs fully programmatically from extracted metadata (Fig.~\ref{fig:qa_generation}). Questions are constructed by instantiating predefined reasoning templates aligned with CRP (Table~\ref{tab:template_ee}-\ref{tab:template_ea}). Each representation contains temporally evolving states, including bounding box trajectories, actions, scenes, outfits, and their transitions. Template slots are deterministically filled with corresponding attributes. For example, given an entity transition ($\{a_1, o_1, s_1\} \rightarrow \{a_2, o_2, s_2\}$), we instantiate: ``Is the person {$a_1$} in {$s_1$} with {$o_1$} at the beginning later seen wearing {$o_2$}?'' to probe entity outfit change reasoning. Ground-truth answers are directly derived from the same structured metadata to ensure semantic consistency. After automatic generation, GPT-4o is used for linguistic refinement (e.g., grammar check and fluency improvement) and to filter out invalid cases, such as questions generated from entity representations without valid state changes. The detailed prompt is provided in Appendix (\S\ref{sec:qa_quality}). 

\paragraph{\textbf{QA Formats and Reasoning Patterns.}} We design three QA formats: binary, multiple-choice, and ordering. The ordering format requires a chronological arrangement of the entity’s attribute transitions, explicitly evaluating fine-grained temporal compositionality beyond recognition. To probe temporal directional bias (\S\ref{sec:pos_bias}), we define three reasoning patterns: (1) \textit{forward} (tracking start $\rightarrow$ end), (2) \textit{backward} (reasoning end $\rightarrow$ start), (3) \textit{agnostic} (reasoning a mid-point $\rightarrow$ start \& end; bidirectional), which are produced programmatically by varying temporal reference points during template instantiation. 

\paragraph{\textbf{Distractor Construction.}} We construct two types of distractors for binary and multiple-choice formats. \textit{Real Distractors (Intra-Clip Negatives)} are sampled from other entities within the \textit{same} video clip. Since these entities co-exist temporally and spatially with the target entity, the model must accurately track and distinguish the target entity. This setting tests fine-grained entity discrimination and tracking under realistic multi-entity scenarios. In contrast, \textit{Synthetic Distractors (Cross-Clip Negatives)} are sampled from entities in \textit{different} video clips, and thus do not appear in the queried clip. This setup evaluates hallucination robustness (\S\ref{sec:distractor}): models should reject entities that are not present in the clip rather than relying on semantic plausibility. Intuitively, real distractors stress entity-level discrimination within context, while synthetic distractors test presence verification and hallucination avoidance. 

\paragraph{\textbf{Quality Review.}}
To assess the quality of the automatically generated QA pairs, we randomly sample 100 items and perform a triple-annotator review (valid/invalid). A QA pair is marked \textit{valid} only if both the question and its ground-truth answer are correctly grounded to the target entity and its states in the video. It is marked \textit{invalid} if either component is incorrect or misaligned, or if distractors contain obvious, non-grounded attributes that make the answer trivial.
Among the sampled QA pairs, 70\% are unanimously judged valid with substantial inter-annotator agreement (Fleiss’ $\kappa$ = 0.767). Most invalid cases arise from overly obvious synthetic distractors with implausible attribute assignments, rather than errors in entity tracking or contextual grounding. This indicates that the core entity-centric pipeline is reliable, consistent with \S\ref{sec:pipeline}. 

To further ensure correctness, human annotators manually verify the attributes inferred by proprietary MLLMs during the contextual recognition step against the corresponding video evidence. Based on this verification, we refine all retained QA pairs by: (1) correcting inconsistent ground-truth answers; (2) revising ambiguous or weakly grounded attributes in the questions; (3) replacing trivial synthetic distractors with visually confusable alternatives. This process yields 1,006 high-quality QA pairs from 406 video clips, spanning diverse genres (e.g., documentary, news, TV drama) and temporal scales up to 659 seconds (Fig.~\ref{fig:video_length}). The final distribution across CRP dimensions is balanced: entity existence (200), action changes (222), outfit changes (192), scene changes (191), and entity ambiguity (201).

To mitigate answer-choice bias, we balance the ground-truth answer distribution across candidates after manual verification. Finally, we randomly re-evaluate the cleaned benchmark with three annotators, achieving 96\% average human accuracy, confirming dataset reliability and clarity.


\section{Experiments}

\newcolumntype{C}{>{\centering\arraybackslash}p{2.4em}}
\newcolumntype{G}{>{\centering\arraybackslash\columncolor{gray!15}}p{2.4em}}
\begin{table*}[t] 
    \small
    \setlength{\tabcolsep}{2pt}
    \renewcommand{\arraystretch}{1.45}
    \centering 
        \resizebox{\linewidth}{!}{%
            \begin{tabular}{l|CCCCCCCCCCCCCG}
            \toprule Model & \multicolumn{2}{c}{Existence (EE)} & \multicolumn{3}{c}{Action Changes (AC)} & \multicolumn{3}{c}{Outfit Changes (OC)} & \multicolumn{3}{c}{Scene Changes (SC)} & \multicolumn{2}{c}{Ambiguity (EA)} & \cellcolor{gray!15}\textbf{Avg.} \\ \cmidrule(lr){2-3} \cmidrule(lr){4-6} \cmidrule(lr){7-9} \cmidrule(lr){10-12} \cmidrule(lr){13-14} & B & MC & B & MC & O & B & MC & O & B & MC & O & B & MC &\cellcolor{gray!15} \\
            \midrule
            \midrule
            \multicolumn{7}{l}{\textit{Open-Source General-Purpose MLLMs}} \\
            \midrule
            Qwen-2.5-VL-7B~\cite{bai2025qwen2} &57.00&36.00&48.91&38.20&17.07&{65.93}&55.42&22.22&46.32&46.58&34.78&70.00&45.55&48.81\\
            Qwen-2.5-VL-32B~\cite{bai2025qwen2} &68.00&42.00&{71.74}&42.70&19.51&61.54&{63.86}&5.56&68.42&{58.90}&30.44&{79.00}&{46.54}&{56.96}\\
    
            InternVL3-8B~\cite{zhu2025internvl3} &{71.00}&55.00&55.44&49.44&29.27&46.15&31.33&22.22&55.79&46.58&39.13&63.00&26.73&48.81\\
            InternVL3-38B~\cite{zhu2025internvl3} &70.00&55.00&56.52&{50.56}&21.96&56.04&51.81&22.22&{66.32}&{58.90}&{52.71}&74.00&36.63&55.47\\
            \midrule
            \multicolumn{7}{l}{\textit{Open-Source Video-Specialized MLLMs}} \\
            \midrule
            mPLUG-Owl3-7B~\cite{ye2023mplug} &60.00&37.00&63.04&35.96&{24.39}&42.86&22.89&16.67&53.68&45.21&21.74&62.00&22.77&42.94\\
            VideoChat2-7B~\cite{videochat2} &45.00&35.00&60.87&38.20&12.20&58.24&30.12&5.56&57.90&35.62&47.83&60.00&21.78&42.55\\
            
            Video-LLaMA2-7B~\cite{cheng2024videollama} &53.00&42.00&58.70&35.96&12.20&50.55&28.92&16.67&55.79&52.06&26.09&51.00&25.74&43.04\\
            Video-LLaMA2-72B~\cite{cheng2024videollama} & 59.00&46.00&52.17&49.44&19.51&42.86&42.17&11.11&62.10&57.53&43.48&71.00&36.63&49.70\\
            
            Video-LLaVA-7B~\cite{videollava} &42.00&35.00&57.61&33.71&9.76&32.97&14.46&{33.33}&52.63&28.77&17.39&32.00&12.87&33.00\\
            
            LLaVA-Video-7B~\cite{zhang2024llava} &67.00&2.00&70.65&17.98&17.07&43.96&25.30&5.56&57.90&20.55&21.74&40.00&11.88&34.39\\
            LLaVA-NeXT-Video-7B~\cite{zhang2024llavanextvideo} &65.00&41.00&68.48&33.71&2.44&47.25&18.07&0.00&54.74&38.36&13.04&63.00&27.72&42.94\\
            LLaVA-NeXT-Video-34B~\cite{zhang2024llavanextvideo} &53.00&47.00&55.44&39.33&7.32&48.35&21.69&0.00&56.84&39.73&13.04&39.00&19.80&39.36\\
            
            VILA-8B~\cite{lin2024vila} &59.00&56.00&60.87&37.08&4.88&51.65&28.92&5.56&60.00&53.43&21.74&53.00&23.76&45.33\\
            
            \midrule
            \multicolumn{7}{l}{\textit{Proprietary MLLMs}} \\
            \midrule
            Gemini-2.0-Flash &67.00&69.00&76.09&52.81&34.15&64.84&49.40&11.11&64.21&60.27&52.17&85.00&34.65&60.24\\
            Gemini-2.5-Flash &73.00&59.00&80.44&61.80&\textbf{63.42}&86.81&75.90&66.67&{76.84}&61.65&65.22&\underline{94.00}&78.22&74.25\\
            Gemini-2.5-Pro  &\textbf{88.00}&\underline{75.00}&{83.70}&\underline{75.28}&\underline{60.98}&\textbf{95.60}&\textbf{86.75}&{83.33}&\textbf{82.11}&\underline{83.56}&\underline{91.30}&\underline{94.00}&\underline{82.18}&\textbf{83.80}\\
            Gemini-3-Flash 
            & \underline{87.00} & \textbf{77.00} 
            & 83.70 & 69.66 & \textbf{63.42} 
            & 89.01 & 57.83 & \textbf{94.44} 
            & \textbf{82.11} & 68.49 & \textbf{95.65} 
            & \underline{94.00} & 78.22 
            & 79.32 \\
            
            Gemini-3.1-Pro 
            & 82.00 & 74.00 
            & \underline{85.87} & \textbf{78.65} & 48.78 
            & \underline{91.21} & \underline{84.34} & \underline{88.89} 
            & \underline{80.00} & \textbf{87.67} & \textbf{95.65} 
            & \textbf{97.00} & \textbf{85.15} 
            & \underline{83.40} \\
            GPT-4.1 
            & 78.00 & 63.00 
            & \textbf{86.96} & 68.54 & 46.34 
            & 82.42 & 74.70 & 66.67 
            & 77.90 & 73.97 & 73.91 
            & 89.00 & 68.32
            & 74.85 \\
            
            GPT-4o  &{77.00}&{61.00}&{82.61}&{66.29}&{39.02}&{82.42}&{67.47}&{44.44}&{75.79}&\underline{83.56}&{78.26}&{86.00}&{61.39}&{72.27}\\
            GPT-4o (20 Frames)  &{78.00}&{58.00}&{81.52}&{60.67}&{36.59}&{80.22}&{67.47}&{61.11}&{72.63}&{75.34}&{78.26}&{85.00}&{66.34}&{70.97}\\
            \bottomrule
        \end{tabular}
    }
    \caption{\textbf{Evaluation Results on \data}. B, MC, and O denote binary, multiple-choice, and ordering questions, respectively. \textbf{Bold} and \underline{underline} stands for the best and second. EE indicates entity existence; AC, OC, refer to entity action, outfit, scene changes, respectively; EA denotes entity ambiguity dimension in CRP. We evaluate all open-source MLLMs using 20 frames per video, while proprietary MLLMs are evaluated using 128 frames by default.}
    \label{tab:main_eval}
\end{table*}

\subsection{Experimental Settings}
\paragraph{\textbf{Raw Video Sources.}}
We construct \data using raw video sources from three widely adopted video datasets: AVA~\citep{gu2018ava}, VideoMME~\citep{fu2025video}, and LVBench~\citep{wang2024lvbench}. To ensure suitable entity-centric evaluation, we filter out dark or low-quality clips and videos containing only a single character without meaningful attribute changes. From the selected videos, we extract entity representations using our automated pipeline (\S\ref{sec:pipeline}) and generate QA pairs (\S\ref{sec:qa_gen}).

\paragraph{\textbf{Model Baselines.}} 
We evaluate 13 open-source and 7 proprietary MLLMs on our benchmark.
The open-source models include both general-purpose and video-specialized MLLMs of varying scales. We refer to open-source general-purpose MLLMs as \textit{OGP-MLLMs} and open-source video-specialized MLLMs as \textit{OVS-MLLMs}. The OGP-MLLMs includes Qwen-2.5-VL-7B, 32B~\citep{bai2025qwen2} and InternVL3-8B, 38B~\citep{zhu2025internvl3}; the OVS-MLLMs include mPLUG-Owl3-7B~\citep{ye2023mplug}, VideoChat2-7B~\citep{videochat2}, Video-LLaMA2-7B, 72B~\citep{cheng2024videollama}, Video-LLaVA-7B~\citep{videollava}, LLaVA-Video-7B~\citep{zhang2024llava}, LLaVA-NeXT-Video-7B, 34B~\citep{zhang2024llavanextvideo}, and VILA-8B~\citep{lin2024vila}. We use Gemini-2.0-Flash, Gemini-2.5-Flash, Gemini-2.5-Pro, Gemini-3-Flash, Gemini-3.1-Pro, GPT-4o, and GPT-4.1 for a proprietary model. All open-source model evaluations are conducted by sampling 20 frames per video following prior work~\citep{zhang2024eventhallusion}, as this setting achieves the best performance across frame densities (\S\ref{sec:pos_bias}). For proprietary models, we use 128 frames per video to take advantage of its larger visual context capacity, while we further evaluate GPT-4o with 20 frames per video for fair comparison.
\paragraph{\textbf{Evaluation Protocol.}}
Our evaluation does not rely on LLM-based judging or open-ended response generation. All questions are formulated as binary-choice, multiple-choice, or option-ordering tasks. Model predictions are scored deterministically using exact-match accuracy against the ground-truth answers, eliminating ambiguity from natural-language phrasing, lexical variation, or subjective assessment.

\subsection{Evaluation Results}

\subsubsection{Quantitative Results}
Table~\ref{tab:main_eval} presents the evaluation results on our benchmark, revealing a substantial performance gap across open-source and proprietary MLLMs. Among OGP-MLLMs, Qwen-2.5-VL-32B achieves the highest accuracy (56.96\%), surpassing the best OVS-MLLMs, Video-LLaMA2-72B (49.70\%). Yet, both remain far behind proprietary MLLMs (e.g., Gemini-2.5-Pro) in maintaining consistent entity tracking. Across open-source models, performance drops sharply on tasks requiring dynamic attribute reasoning, such as action and outfit changes or entity disambiguation, while scenes change tasks show relatively better accuracy due to their lower visual variability. This pattern suggests that current open-source MLLMs struggle to integrate temporal coherence with fine-grained perceptual grounding. Scaling trends further reveal that larger models generally perform better (e.g., InternVL3-38B vs. 8B: +6.66\%; Video-LLaMA2-72B vs. 7B: +5.66\%), though gains are inconsistent, as seen in LLaVA-NeXT-Video-34B, which slightly underperforms its 7B version, indicating that size alone does not guarantee improved entity tracking.

In contrast, GPT-4o attains the highest overall performance (70.97\%) under the same frame density, consistently outperforming all open-source baselines. Increasing the number of input frames to 128 further improves its average performance. Yet, even this strong proprietary model shows limitations in maintaining robust entity-level continuity over extended narratives. Further analysis confirms that \data requires genuine multimodal grounding rather than reliance on language priors: removing visual inputs reduces GPT-4o accuracy by 30.52\%, approaching the random baseline, while reversing video frames drastically lowers performance on the entity change ordering task from 51.2\% to 6.1\% (Table~\ref{tab:baseline}), demonstrating the necessity of correct temporal reasoning. These findings confirm that narrative understanding is a core unsolved challenge, emphasizing the need for MLLMs that can jointly model fine-grained visual perception, temporal consistency, and entity-centric reasoning over extended video contexts. 
\begin{figure*}[t]
    \centering
    \begin{subfigure}[t]{0.32\linewidth}
        \centering
        \includegraphics[width=\textwidth]{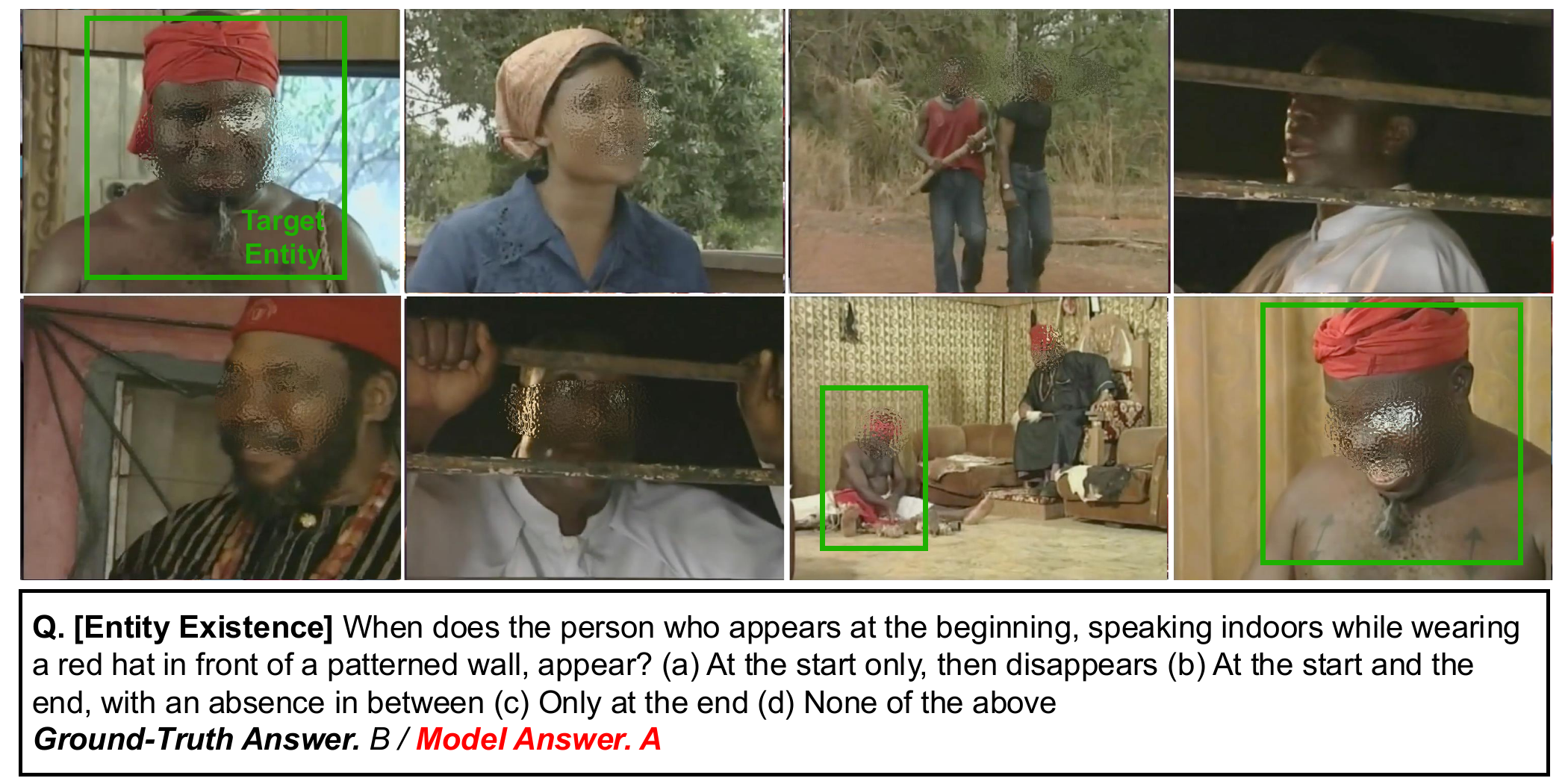}
        \caption{Entity Existence}
        \label{fig:err_ee}
    \end{subfigure}
    \begin{subfigure}[t]{0.32\linewidth}
        \centering
        \includegraphics[width=\textwidth]{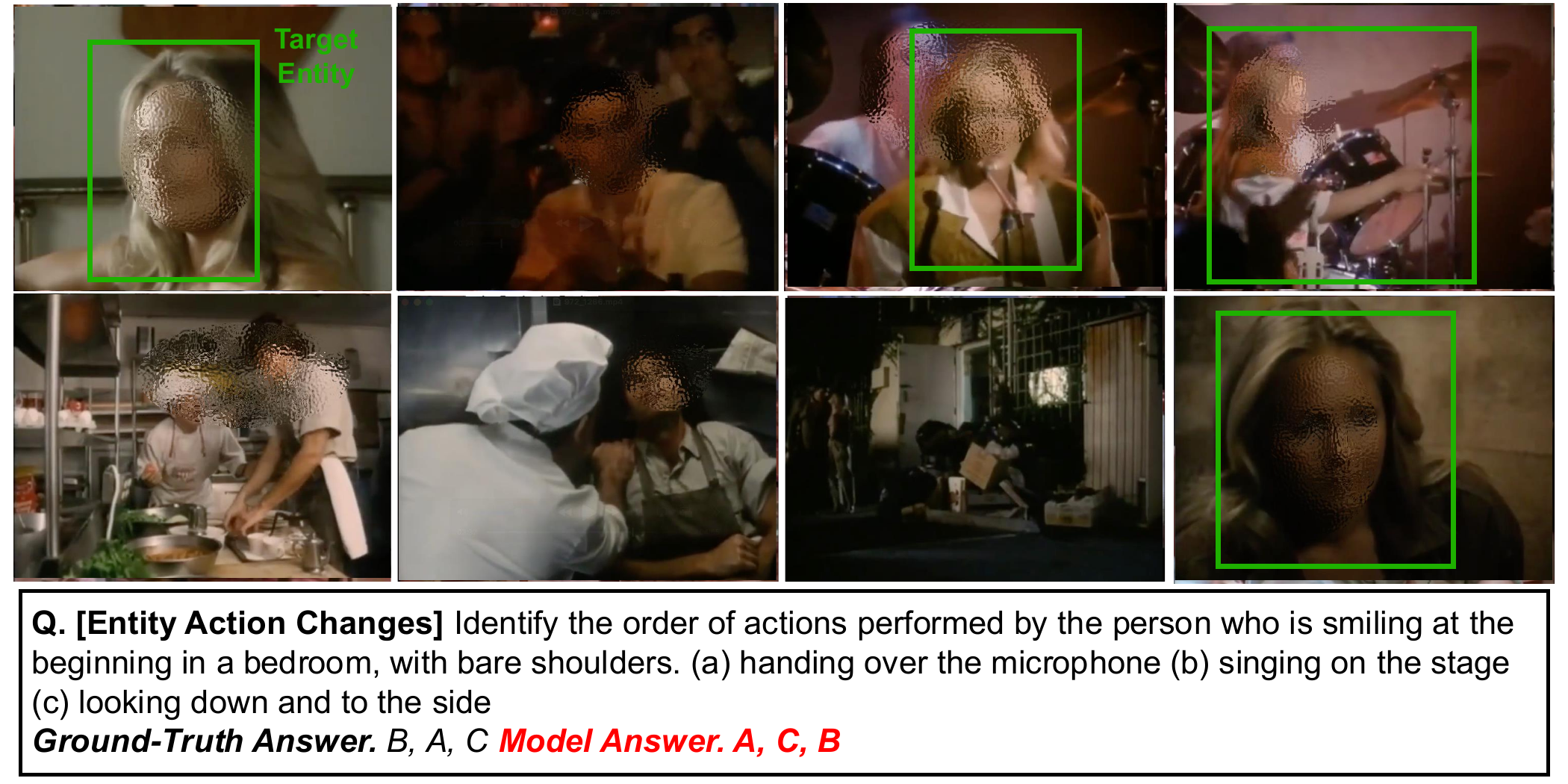}
        \caption{Entity Changes}
        \label{fig:err_ec}
    \end{subfigure}
    \begin{subfigure}[t]{0.32\linewidth}
        \centering
        \includegraphics[width=\textwidth]{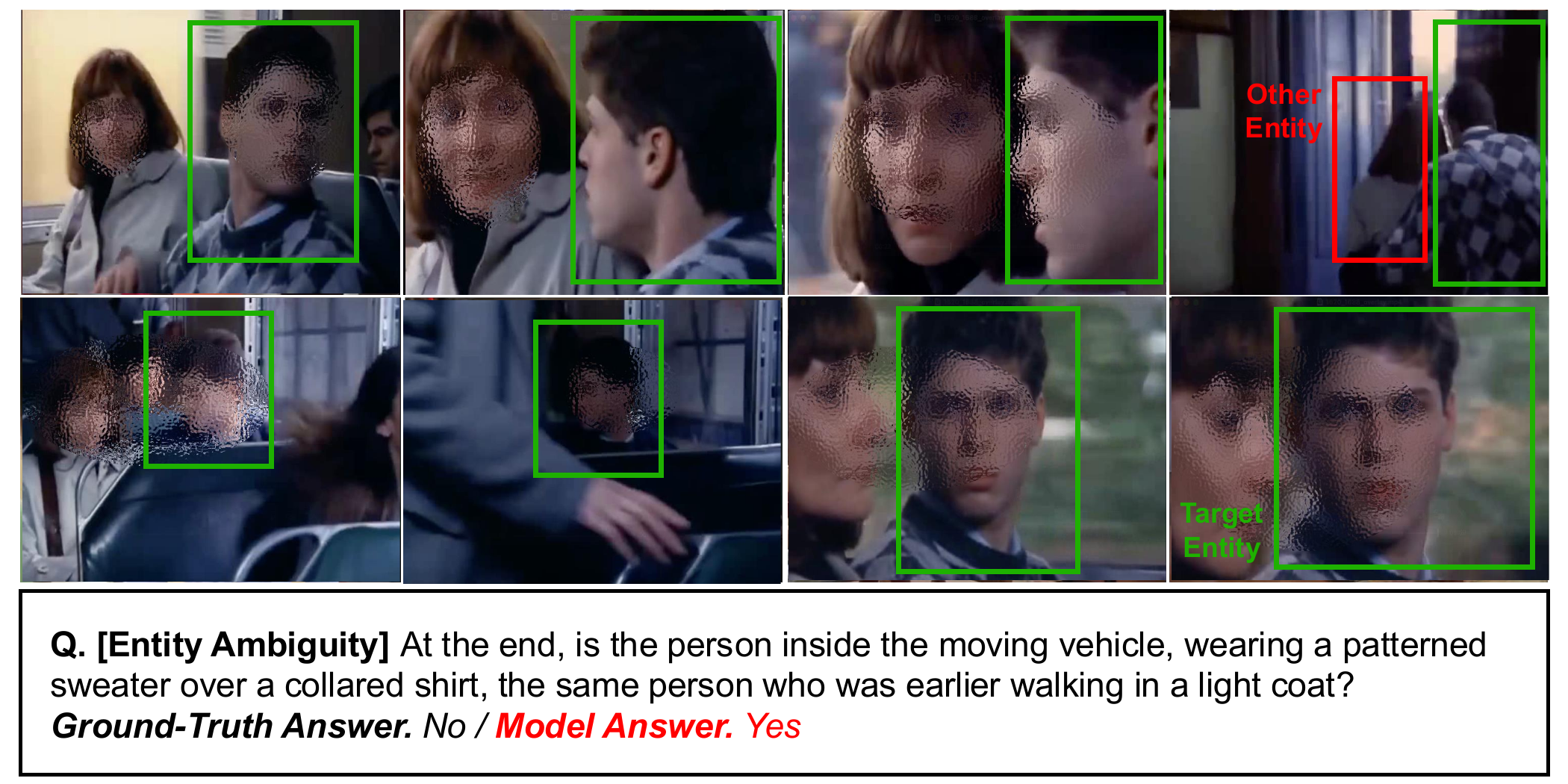}
        \caption{Entity Ambiguity}
        \label{fig:err_ea}
    \end{subfigure}
    \caption{\textbf{Examples of Model Failure in \data.}Video sources are extracted from AVA~\citep{gu2018ava} and the output is generated from Video-LLaMA2-7B~\citep{cheng2024videollama}.}
    \label{fig:error_example}
\end{figure*}



\subsubsection{Qualitative Results}
We conduct a qualitative analysis of representative failure cases to better understand model limitations across reasoning levels. At the entity existence level, models frequently overestimate continuity, predicting that an entity persists throughout the video even when it appears only briefly at the beginning (Fig.~\ref{fig:err_ee}). At the entity change level, errors compound: when entities undergo subtle attribute transitions, models often produce semantically plausible yet visually ungrounded responses (e.g., singing on the stage, handing over a microphone, and looking down), exposing deficiencies in fine-grained perception and temporal reasoning (Fig.~\ref{fig:err_ec}). The most severe failures occur at the entity ambiguity level, where multiple entities interact or reappear. Even strong proprietary models struggle here, often confusing identities or hallucinating entity presence (Fig.~\ref{fig:err_ea}). These error patterns reveal hierarchical failure cascades, where misperception at lower levels propagates upward, underscoring the persistent challenge of modeling entity continuity and temporal dependencies essential for coherent narrative understanding. \looseness=-1

\section{Analysis}
All analyses are conducted using 13 MLLMs: 12 open-source models from the Qwen2.5-VL, InternVL3, Video-LLaMA2, and LLaVA-NeXT-Video families, along with mPLUG-Owl3-7B, VideoChat2-7B, Video-LLaVA-7B and VILA-8B, plus GPT-4o. Following prior works~\citep{fu2025video, wang2024lvbench}, we group them by training objective: OGP-MLLMs emphasize visual grounding through image-text alignment, whereas OVS-MLLMs model temporal dynamics through video-text alignment. We therefore analyze trends within each category rather than treating cross-category differences as direct measures of overall model quality.

\begin{figure*}[t]
    \centering
    \begin{subfigure}[t]{0.32\linewidth}
        \centering
        \includegraphics[width=\textwidth]{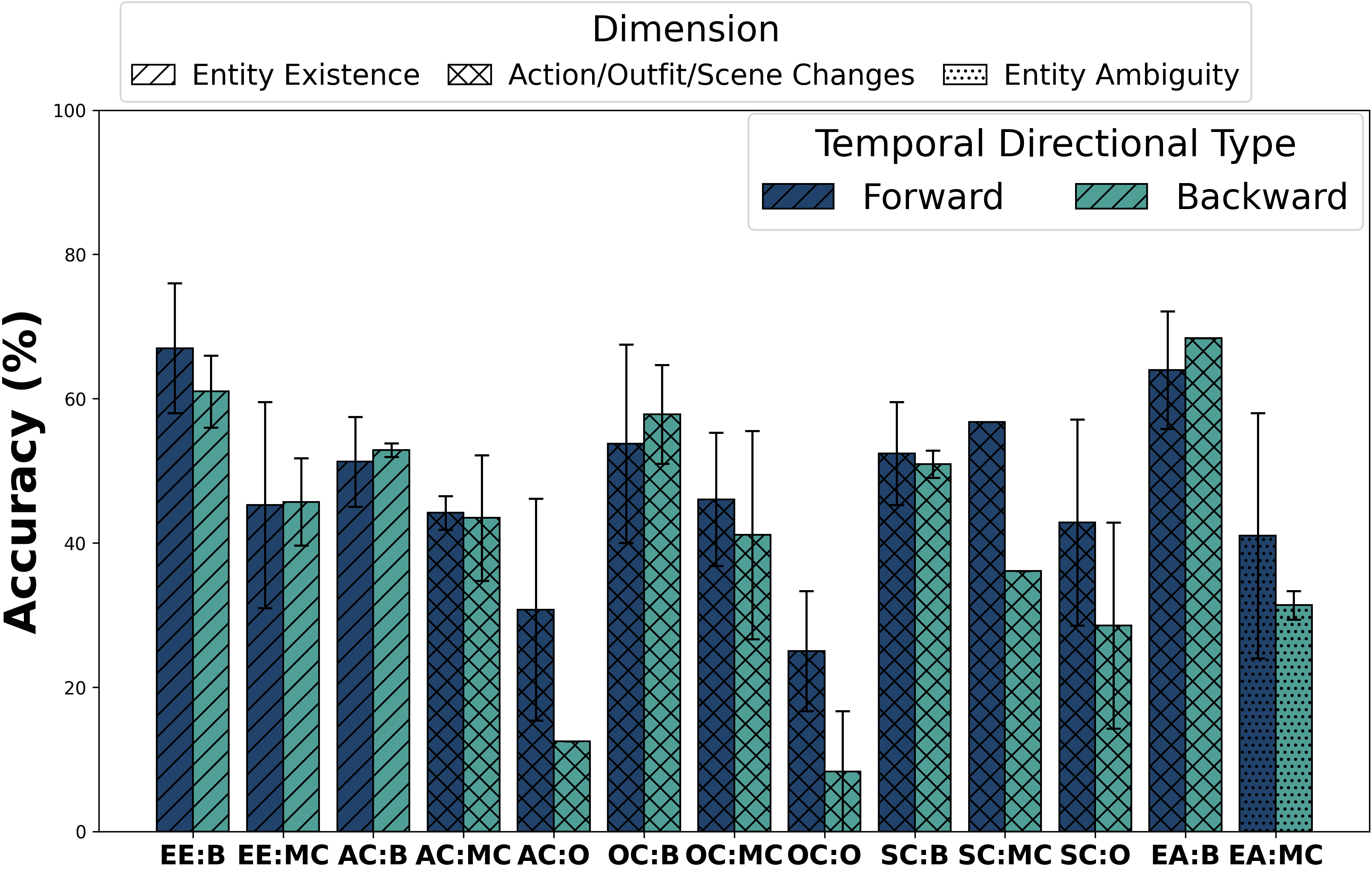}
        \caption{OGP-MLLMs.}
        \label{fig:pos_image}
    \end{subfigure}
    \begin{subfigure}[t]{0.32\linewidth}
        \centering
        \includegraphics[width=\textwidth]{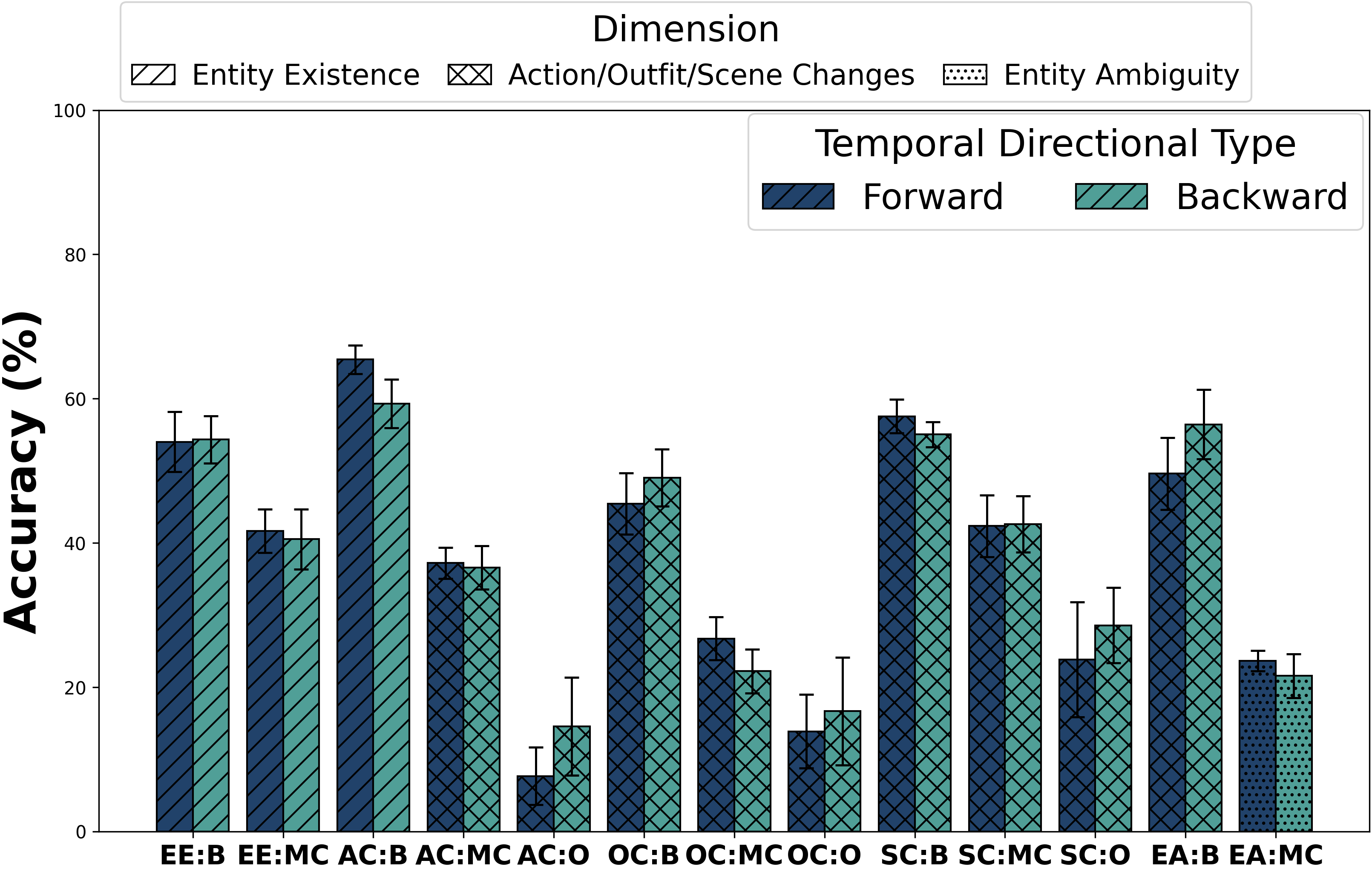}
        \caption{OVS-MLLMs.}
        \label{fig:pos_video}
    \end{subfigure}
    \begin{subfigure}[t]{0.32\linewidth}
        \centering
        \includegraphics[width=\textwidth]{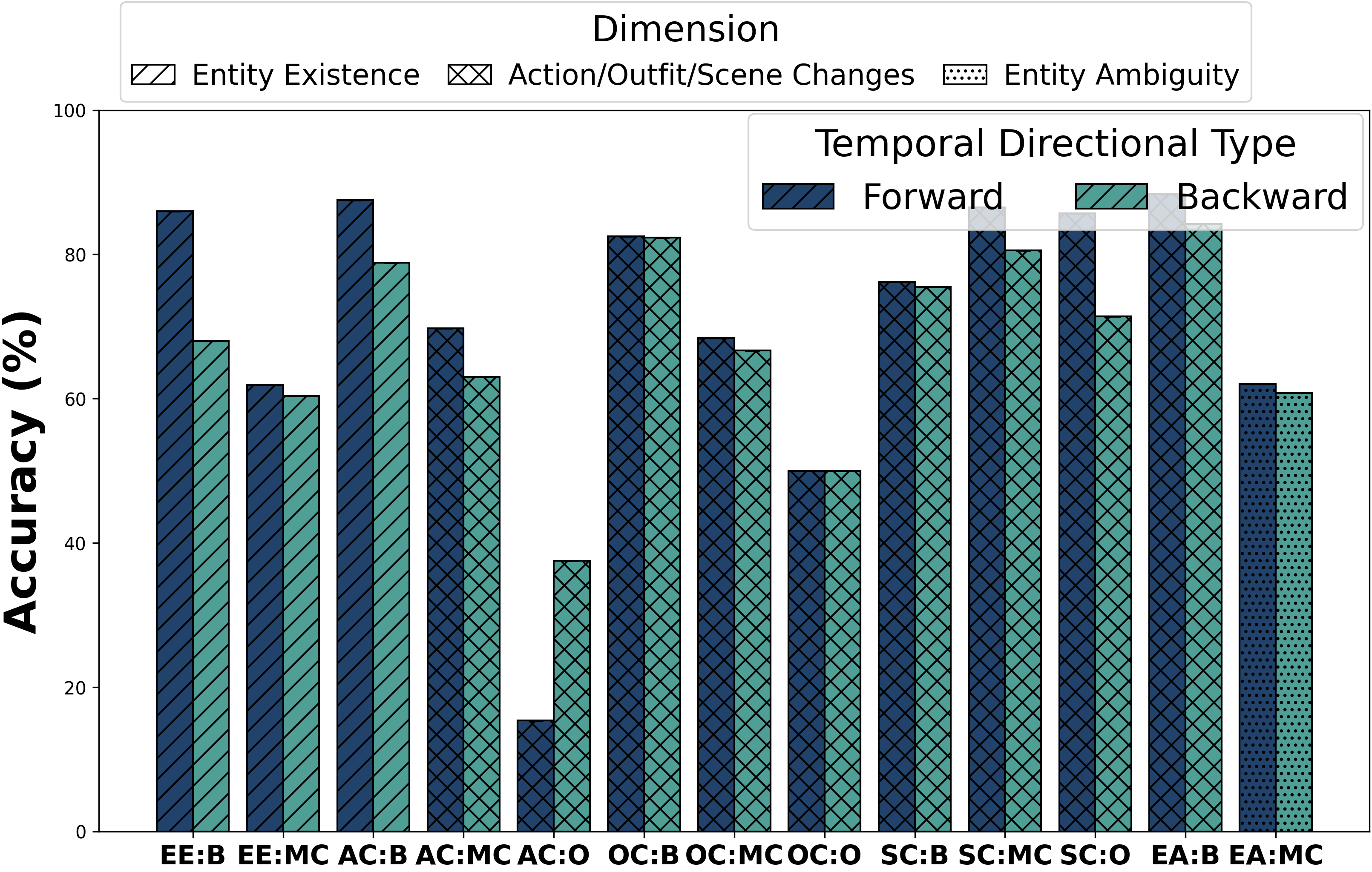}
        \caption{Proprietary MLLMs.}
        \label{fig:pos_prop}
    \end{subfigure}
    \caption{\textbf{Temporal Directional Bias of MLLMs.} MLLMs encode temporal relations in a forward-only manner and fail to generalize to reversed or bidirectional temporal contexts.}
    \label{fig:position_bias}
\end{figure*}
\subsection{Temporal Directional Bias.}
\label{sec:pos_bias}

Our benchmark includes two reasoning types: \textit{forward} and \textit{backward} reasoning as described in \S\ref{sec:qa_gen}. Across model types, we observe a consistent advantage in forward reasoning (Fig.~\ref{fig:position_bias}), revealing a strong directional bias in temporal reasoning. The performance gap between forward and backward reasoning reaches 20.65\%, 9.96\%, and 17.55\% for OGP-, OVS-, and proprietary MLLMs, respectively. This suggests that models can effectively propagate entity states along the video timeline but struggle to infer them in reverse.
This phenomenon parallels the Reversal Curse in LLMs~\citep{berglund2024reversalcursellmstrained}, where models trained on “A is B” fail to generalize to “B is A”. Similarly, MLLMs encode temporal relations directionally, binding entities to sequentially observed events without learning an invertible mapping between earlier and later states. In essence, models can extend a narrative but cannot rewind it, rooted in the causal, left-to-right decoding paradigm inherited from their LLM backbones. We further introduce an \textit{agnostic reasoning} in ordering questions, requiring bidirectional inference of entity states from a middle point. This condition yields the lowest accuracy (Fig.~\ref{fig:pos_ordering_agnostic}), underscoring that current MLLMs exhibit a forward bias and lack coherent temporal grounding mechanisms for reasoning across non-sequential or bidirectional contexts. Addressing this limitation may require bidirectional temporal modeling or contrastive reversal objectives that explicitly enforce symmetry between forward and backward temporal reasoning over entity states.
\begin{figure*}[t]
    \centering
    \begin{subfigure}[t]{0.32\linewidth}
        \centering
        \includegraphics[width=\textwidth]{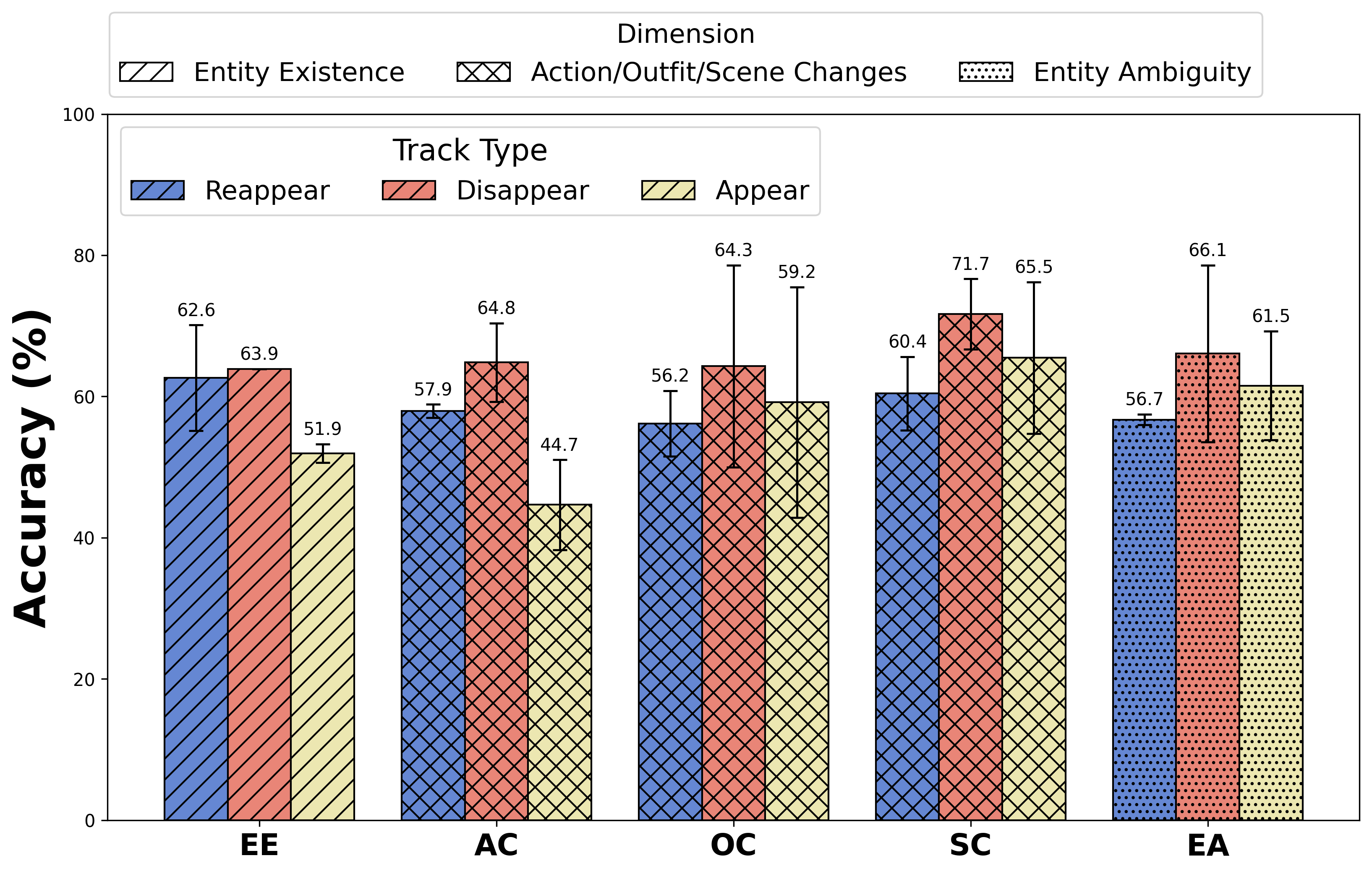}
        \caption{OGP-MLLMs.}
        \label{fig:track_type_image}
    \end{subfigure}
    \begin{subfigure}[t]{0.32\linewidth}
        \centering
        \includegraphics[width=\textwidth]{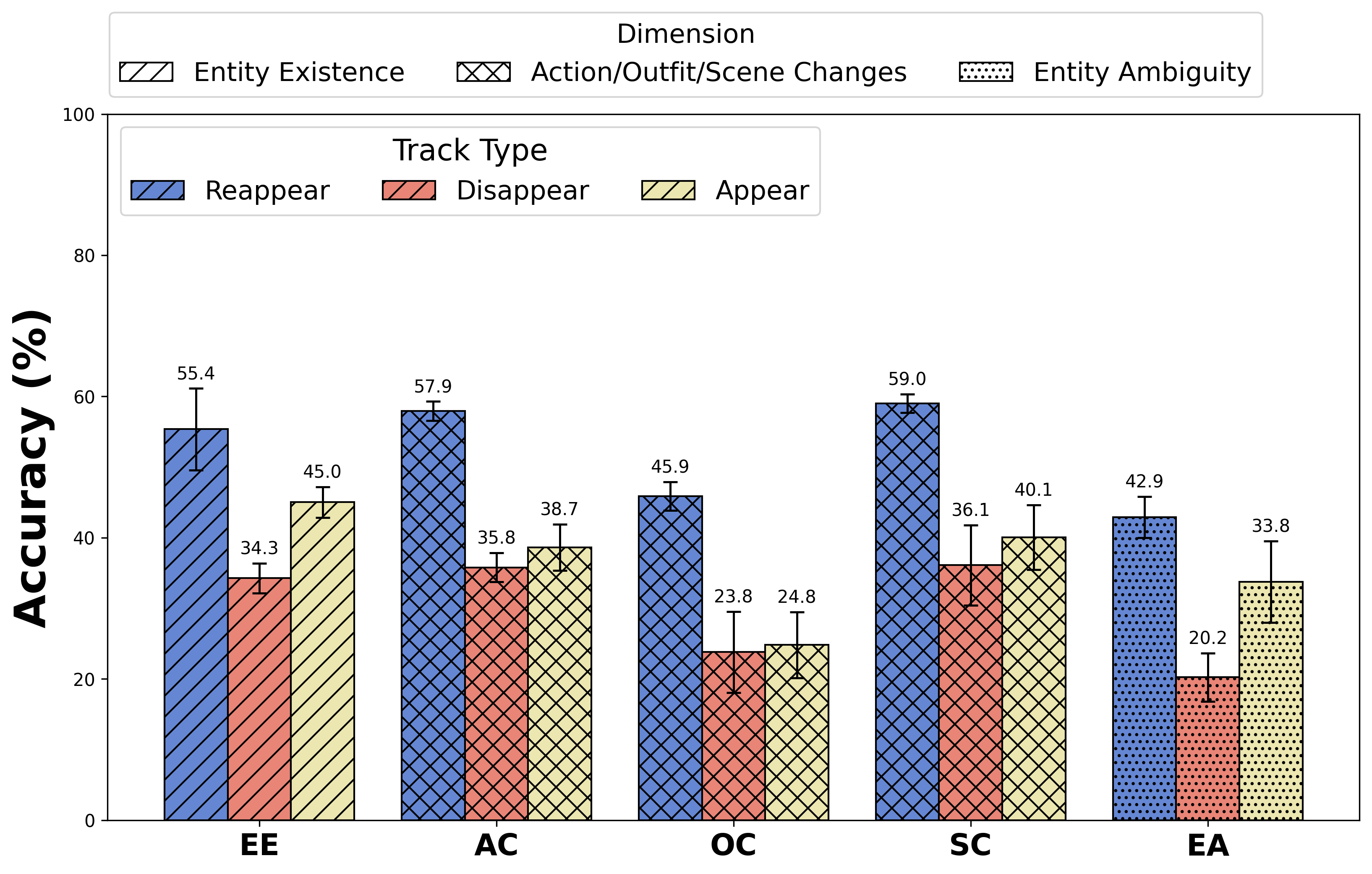}
        \caption{OVS-MLLMs.}
        \label{fig:track_type_video}
    \end{subfigure}
    \begin{subfigure}[t]{0.32\linewidth}
        \centering
        \includegraphics[width=\textwidth]{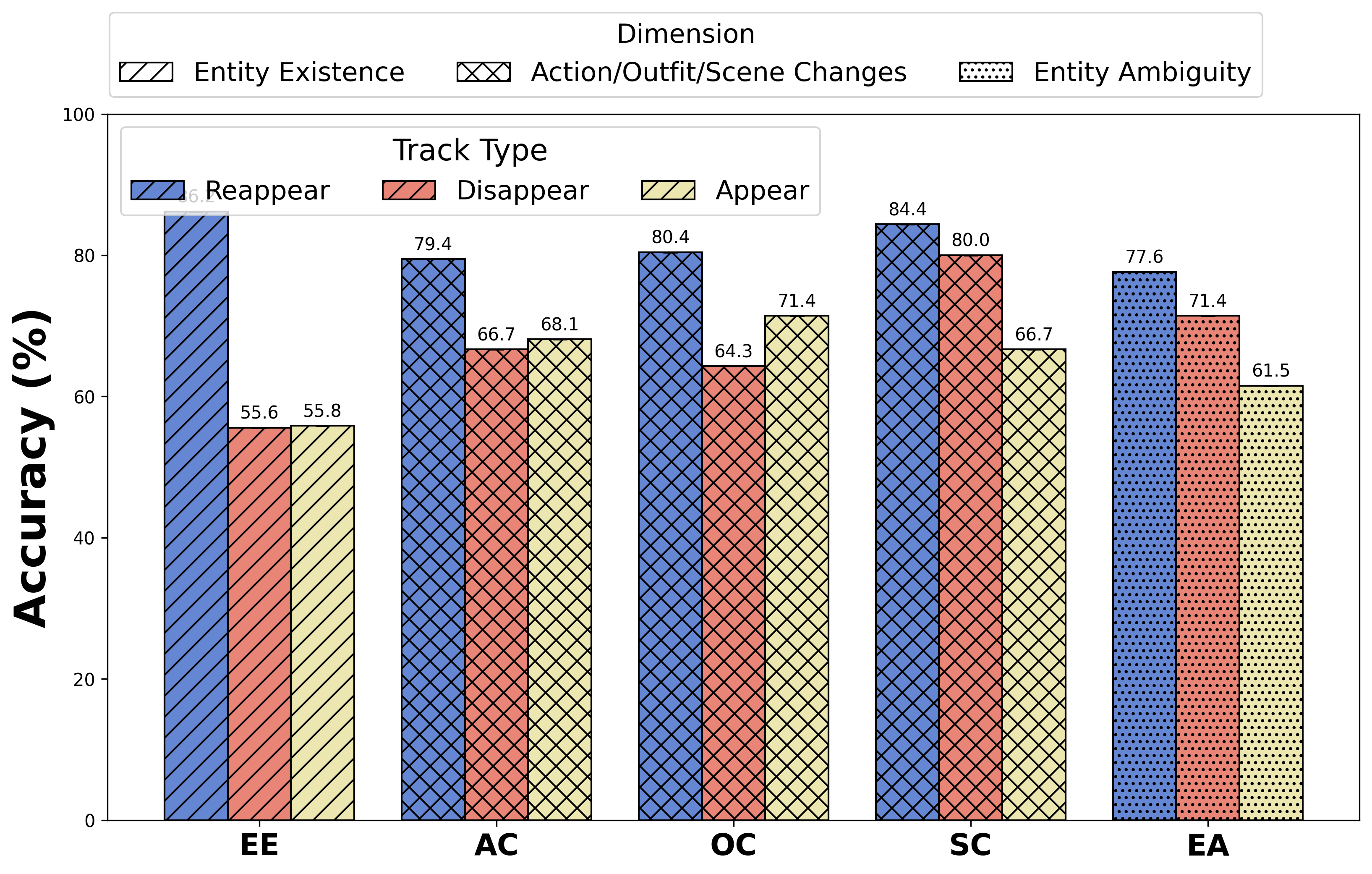}
        \caption{Proprietary MLLMs.}
        \label{fig:track_type_prop}
    \end{subfigure}
    \caption{\textbf{Performance Across Entity Continuity Types.} OGP-MLLMs perform best on \textit{disappear} cases, relying on static visual cues, whereas OVS- and proprietary MLLMs excel on \textit{reappear} cases, reflecting stronger temporal integration but a higher tendency to hallucinate visual details.}
    \label{fig:track_type}
\end{figure*}

\begin{figure*}[t]
    \centering
    \begin{minipage}[t]{0.75\textwidth}
        \centering
        \begin{subfigure}[t]{0.32\linewidth}
            \centering
            \includegraphics[width=\textwidth]{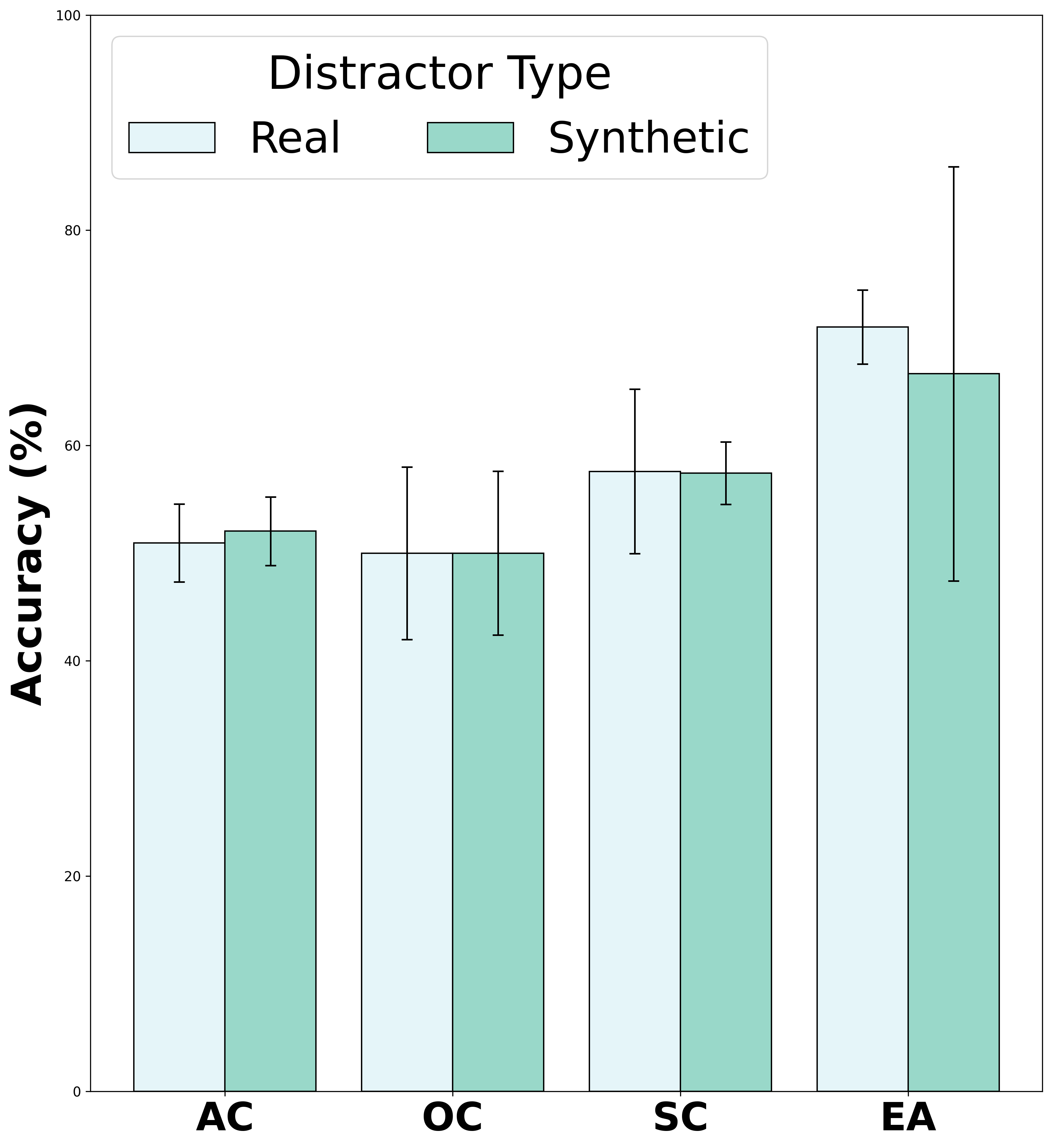}
            \caption{OGP-MLLMs.}
            \label{fig:distractor_type_image}
        \end{subfigure}
        \hfill
        \begin{subfigure}[t]{0.32\linewidth}
            \centering
            \includegraphics[width=\textwidth]{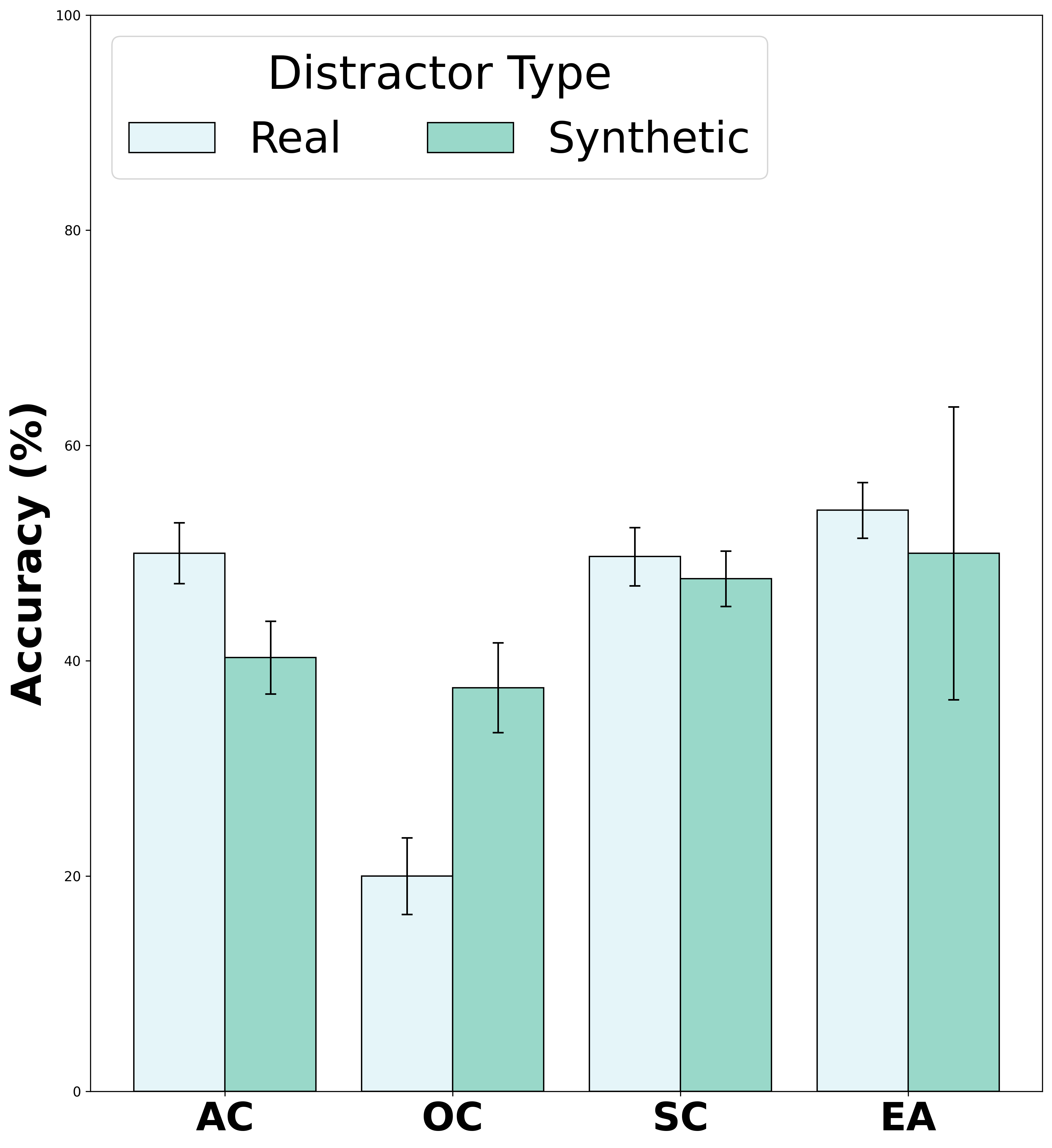}
            \caption{OVS-MLLMs.}
            \label{fig:distractor_type_video}
        \end{subfigure}
        \hfill
        \begin{subfigure}[t]{0.32\linewidth}
            \centering
            \includegraphics[width=\textwidth]{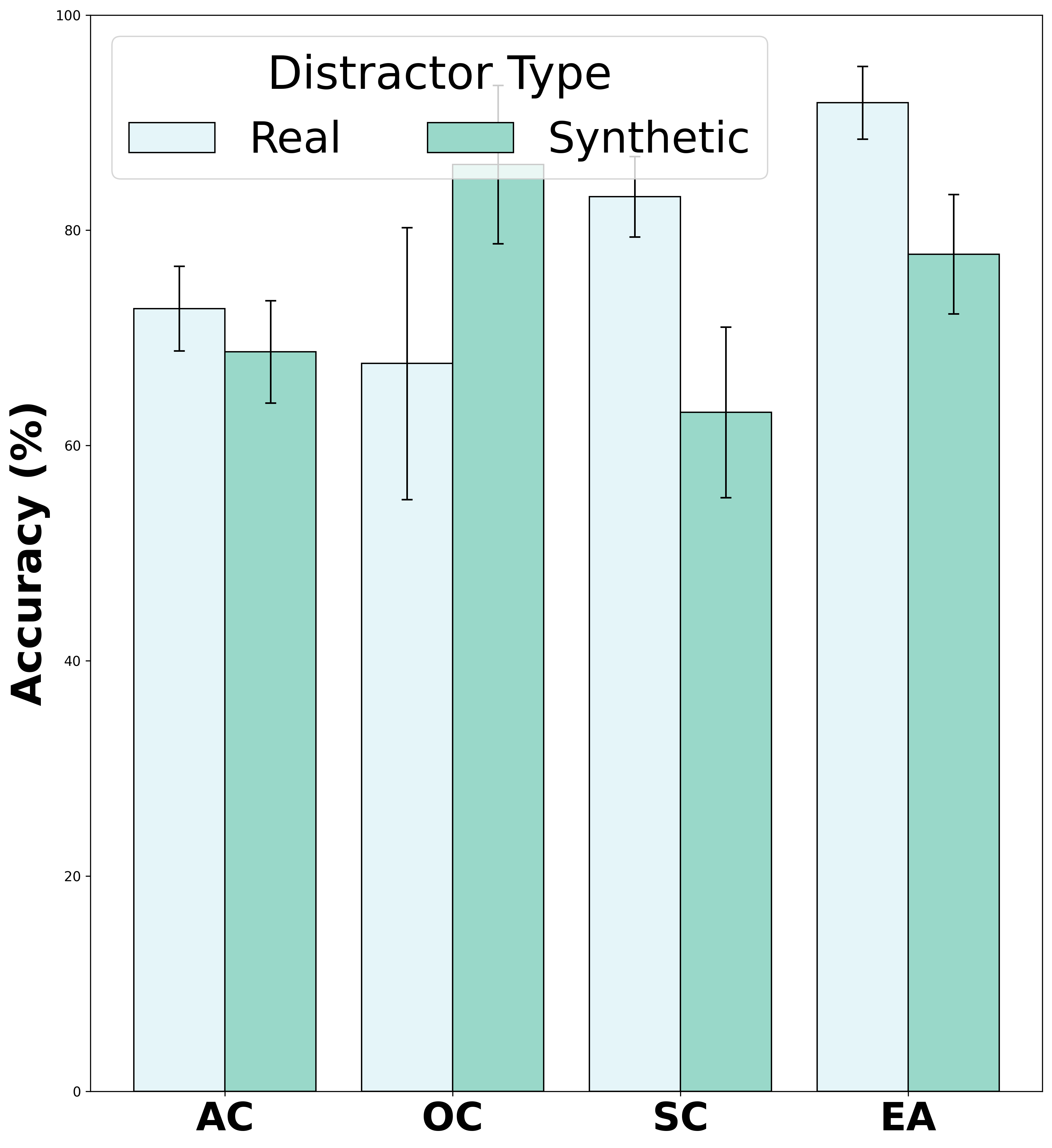}
            \caption{Proprietary MLLMs.}
            \label{fig:distractor_type_prop}
        \end{subfigure}
        \caption{\textbf{Performance Across Distractor Types.} OGP-MLLMs show stable performance across real and synthetic distractors, whereas OVS-MLLMs and proprietary MLLMs exhibit notable drops with synthetic distractors, revealing stronger hallucination tendencies.}
        \label{fig:distrator_type}
    \end{minipage}%
    \hfill
    \begin{minipage}[t]{0.23\textwidth}
        \centering
        \includegraphics[width=\linewidth]{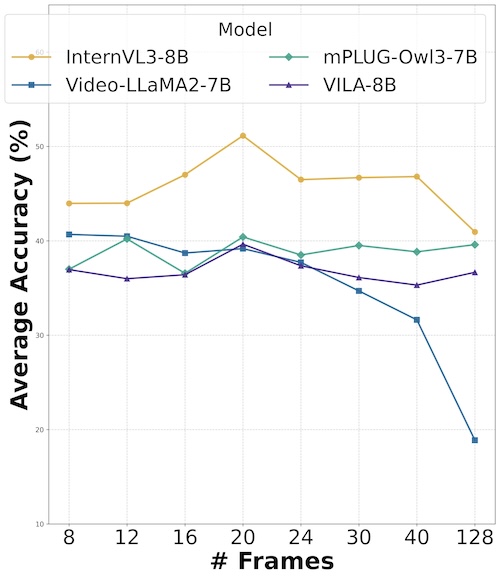}
        \caption{\textbf{Performance Across Frame Densities.} Scaling temporal coverage does not guarantee performance gain.}
        \label{fig:num_frames}
    \end{minipage}
\end{figure*}

\subsection{Entity Continuity Types.}
We categorize entity continuity into three types: \textit{appear}, \textit{disappear}, and \textit{reappear}. In appear and disappear cases, the target entity is visible in only one segment, entering or leaving the scene once. Conversely, reappear cases are the most temporally dynamic, requiring models to maintain identity across multiple disjoint segments. As shown in Fig.~\ref{fig:track_type}, OGP-MLLMs perform best on \textit{disappear} cases, suggesting reliance on static visual evidence rather than reasoning over extended temporal sequences. In contrast, OVS-MLLMs achieve higher accuracy on \textit{reappear} cases, reflecting stronger temporal integration but also a tendency to overpredict reappearances, often hallucinating entity reappearances or misattributing visual attributes. Proprietary MLLMs exhibit a similar behavior, suggesting that stronger temporal integration does not necessarily ensure precise entity grounding, particularly when identity must be maintained across temporally disjoint observations.

To verify that this pattern is not driven solely by basic entity-detection failures, we conduct a model-specific analysis conditioned on correct entity-existence predictions (Tab.~\ref{tab:real_vs_synth}). For each model, we evaluate entity-state changes and entity ambiguity only on videos in which the target entity is correctly identified as present or absent. Even under this controlled setting, OVS-MLLMs show larger performance drops on real (-11.5\%) and synthetic (-14.9\%) distractors, while remaining competitive on temporally driven tasks such as action changes. These results reveal a temporal--perceptual trade-off: OVS-MLLMs better capture temporal continuity but are more susceptible to perceptual confusion, whereas OGP-MLLMs provide stronger visual grounding but weaker temporal integration.




\subsection{Distractor Types.}
\label{sec:distractor}
We compare model performance across two distractor types: \textit{real} and \textit{synthetic} as described in \S\ref{sec:qa_gen}. Ideally, synthetic distractors should be easier since they are visually unrelated to the video. However, model behaviors diverge notably (Fig~\ref{fig:distrator_type}). OGP-MLLMs perform almost identically across both types, where they work slightly better on synthetic ones (0.44\% higher in average), suggesting reliance on localized visual cues and effective rejection of irrelevant attributes. In contrast, OVS-MLLMs exhibit an accuracy drop on synthetic distractors (up to 9.69\%), indicating stronger hallucination tendencies and weaker visual grounding.
Proprietary MLLMs achieve the highest overall performance, yet demonstrate the largest real-synthetic gap (up to 20.03\%), implying that even advanced models struggle to suppress contextually irrelevant generations. These results reveal a fundamental limitation of current MLLMs: while temporal modeling broadens contextual understanding, it also amplifies dependence on textual or global priors, undermining fine-grained visual discrimination required for robust multimodal reasoning. \looseness=-1

\subsection{Frame Density.}
Prior work shows that denser frame sampling improves long-video reasoning by enriching global context~\citep{wang2024lvbench}. To examine whether greater temporal coverage similarly benefits entity-centric reasoning, we vary the number of input frames for open-source MLLMs, $k \in \{8, 12, 16, 20, 24, 28, 30, 40, 128\}$. 
The accuracy peaks $k=20$ but degrades beyond that point, contrasting the monotonic gains seen in long-video reasoning tasks (Fig.~\ref{fig:num_frames}). The drop is particularly pronounced for Video-LLaMA2, which is trained on only 8 frames and becomes unstable when supplied with 128. These results indicate that the bottleneck in entity tracking is not the amount of temporal coverage, but the model's ability to maintain temporal coherence and perceptual grounding as more frames are introduced. Simply increasing frame density adds redundant information rather than improving entity-centric performance.

\section{Conclusion}
We introduced \data, the first benchmark for evaluating narrative understanding from a bottom-up entity-centric perspective via fine-grained entity tracking. Grounded in a Compositional Reasoning Progression spanning entity existence, changes, and ambiguity with a fully automated pipeline, \data provides a scalable and diagnostic framework that systematically measures how MLLMs reason about entities and their temporal evolution. Our results reveal that current MLLMs excel at capturing static visual cues but fail to maintain coherent entity representations under temporal dynamics and visual ambiguity. This exposes a fundamental trade-off between temporal integration and perceptual precision: models can aggregate global context yet often lose fine-grained visual grounding, leading to hallucinated or inconsistent entity representations. Despite scaling and architectural advances, they remain limited by directional bias and weak cross-frame coherence, highlighting the need for entity-centric learning and bidirectional temporal modeling to move MLLMs beyond surface-level recognition toward fine-grained narrative reasoning. \looseness=-1

\clearpage

\bibliographystyle{plainnat}
\bibliography{main}
\clearpage
\appendix

\section{Existing Benchmarks}
\label{app:existing_benchmark}
Existing VideoLLM evaluation benchmarks mostly focus on semantic understanding, where the questions often can be answered from a single frame without requiring the composition of multiple frames. Below are examples from existing benchmarks that do not require true temporal reasoning.

\begin{figure}[h]
    \centering
    \includegraphics[width=0.5\linewidth]{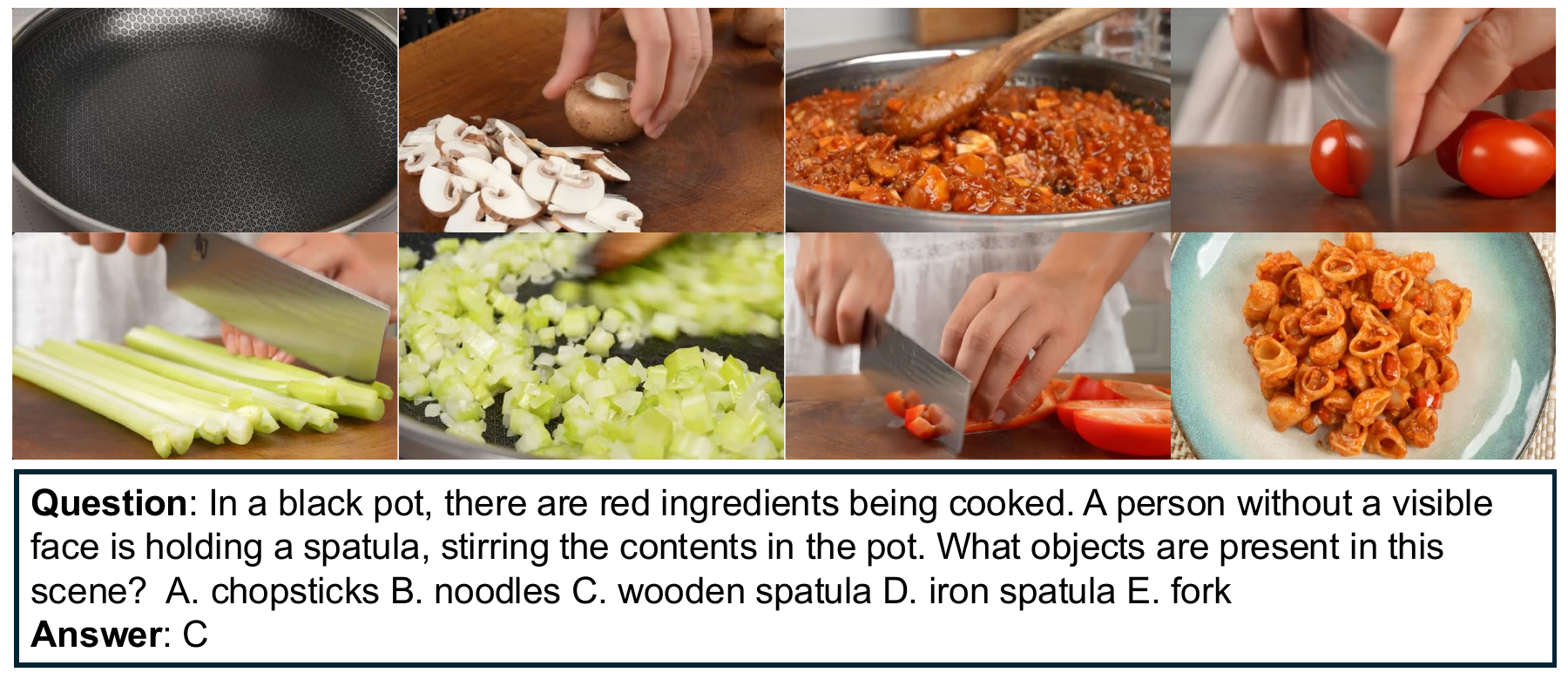}
    \includegraphics[width=0.5\linewidth]{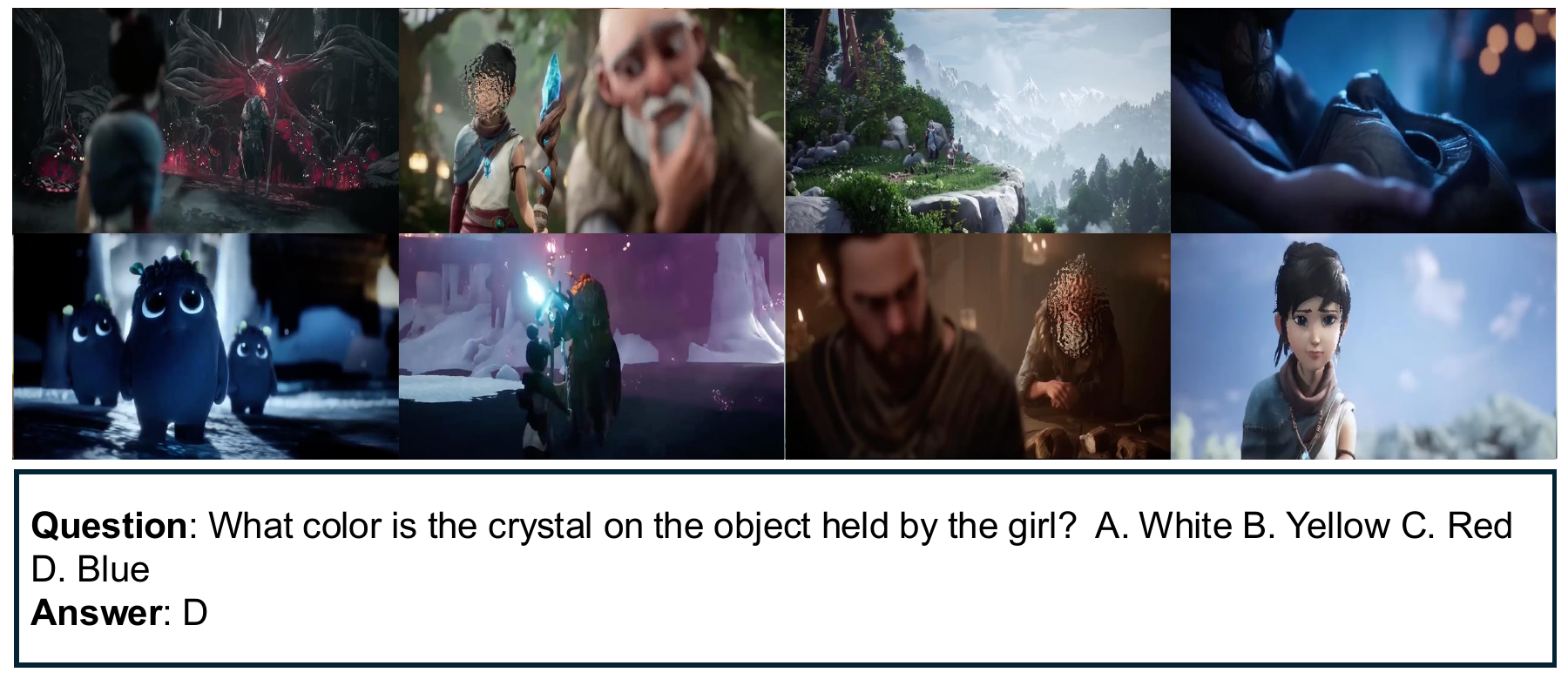}
    \includegraphics[width=0.5\linewidth]{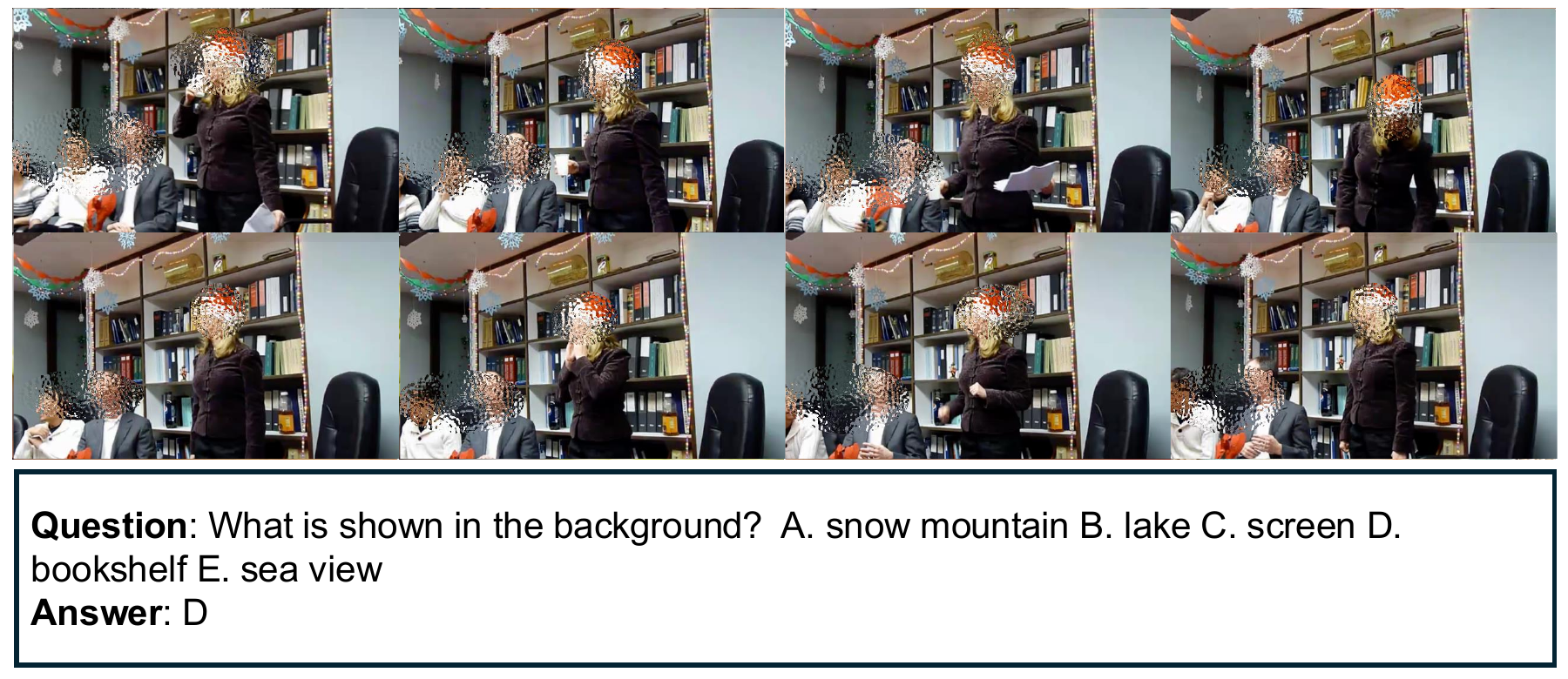}
    \includegraphics[width=0.5\linewidth]{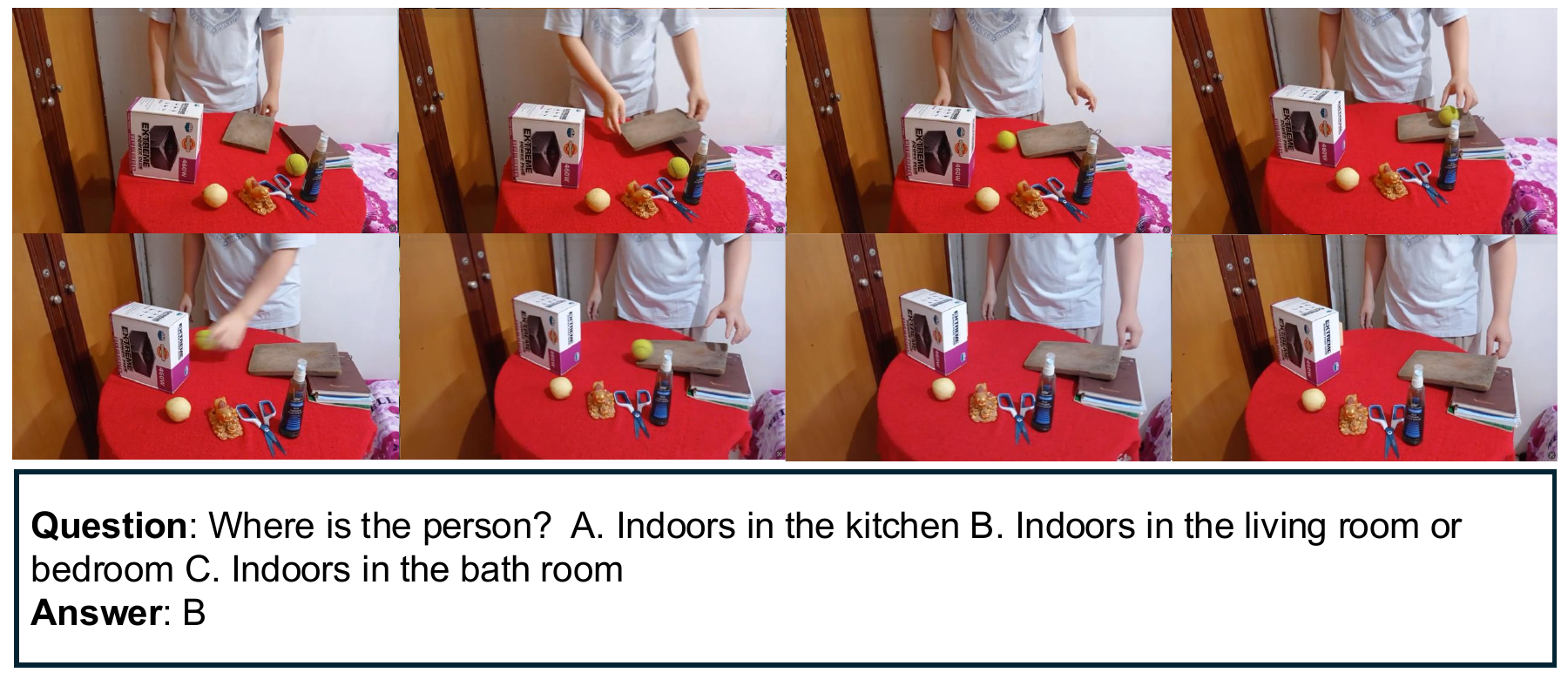}
    \caption{Examples of existing benchmarks in order of LongVideoBench~\cite{wu2024longvideobench}, LVBench~\cite{wang2024lvbench}, NeXT-QA~\cite{xiao2021next}, PerceptionTest~\cite{patraucean2023perception}. Existing benchmarks focused on semantic understanding that can be answered even from a single frame or scene, neglecting true temporal reasoning.}
    \label{fig:existing_examples}
\end{figure}

\section{Video Sources of \data}
\data leverages three widely adapted video sources: AVA~\citep{gu2018ava}, Video-MME~\citep{fu2025video}, and LVBench~\citep{wang2024lvbench}. The test set in the AVA dataset contains 64 movie clips, each 15 minutes long. The Video-MME dataset spans six genres and includes 900 videos with an average length of 1017.9 seconds. The LVBench dataset is designed to evaluate long-video understanding and covers six genres with an average length of 4,101 seconds. A breakdown of video lengths across different data sources is provided in Table~\ref{tab:dataset}.

\begin{table}[h]
\centering
\normalsize
\resizebox{\linewidth}{!}{
    \begin{tabular}{lccccc|c}
    \toprule
    \textbf{Source} 
    & \textbf{$\leq$p50 ($\leq$32s)} 
    & \textbf{p50–p75 (32–57s)} 
    & \textbf{p75–p90 (57–116s)} 
    & \textbf{p90–p95 (116–146s)} 
    & \textbf{$>$p95 ($\geq$146s)} 
    & \textbf{Total} \\
    \midrule
    AVA        & 127 & 54  & 44  & 35 & 22 & 282 \\
    LVBench    & 130 & 41  & 34  & 4  & 6  & 215 \\
    VideoMME   & 256 & 146 & 73  & 15 & 19 & 509 \\
    \midrule
    \textbf{Total} & 513 & 241 & 151 & 54 & 47 & 1006 \\
    \bottomrule
    \end{tabular}
}
\caption{Breakdown by Video Source and Length Percentile.}
\label{tab:dataset}
\end{table}

\subsection{Scalability of \data}
While our pipeline generates 3.8k QA pairs in total, we curate a 1k subset to ensure high quality and balanced coverage across reasoning dimensions. A triple-annotator evaluation finds that 70\% of automatically generated pairs are valid, with most errors stemming from weak distractors rather than errors in the pipeline. Scalability is not a limitation, where our automated pipeline can be extended to larger corpora. 

\section{Details on Automated Pipeline}
\subsection{Models} To identify the main characters, we use the ReID model\footnote{\url{https://kaiyangzhou.github.io/deep-person-reid/MODEL\_ZOO.html}}, especially osnet\_x1\_0, to extract embeddings of detected bounding boxes and select the four largest clusters based on embedding clustering. For entity detection, we employ Detectron2\footnote{\url{https://huggingface.co/spaces/lkeab/transfiner/blob/749f060b6553585cd858b890648a25af83828550/configs/COCO-Detection/faster\_rcnn\_R\_50\_FPN\_3x.yaml}} and Owlv2 base\footnote{\url{https://huggingface.co/google/owlv2-base-patch16-ensemble}} with a confidence threshold of 0.3. Entity tracking is performed using face recognition\footnote{\url{https://github.com/ageitgey/face\_recognition}} to assign identity labels to form a consistent entity trajectory across frames. To refine these trajectories and remove false assignments, we apply majority voting among Gemini family models using the prompt described in \S\ref{appendix:mv_prompt}. Finally, we elaborate on the detailed prompt for contextual recognition with Gemini-2.5-Pro in \S\ref{appendix:recog_prompt}.

\subsection{Majority Voting}
\label{appendix:mv_prompt}
\begin{tcolorbox}[
  enhanced,
  breakable,
  width=0.98\linewidth,
  colback=tzBlueFill,
  colframe=tzBlueBorder,
  boxrule=1.2pt,
  arc=6pt,
  left=5pt,right=5pt,top=4pt,bottom=2pt,
  title={\small Majority Voting},
  coltitle=white,
  colbacktitle=tzBlueHeader2,
  fonttitle=\bfseries,
]
\small
\begin{lstlisting}[style=jsonTiny]
You are given a reference image of a person, followed by a set of {n} face crop images. Your task is to determine, for each face crop, whether it depicts the same person as in the reference image.
Do not evaluate the reference image itself. Only compare the face crops to the reference image.

### Important:
- You should analyze all face crop images together, not in isolation.
- Use visual context from the entire set of face crops to help inform your decision for each one.
  For example, if several face crops share similar features or accessories, use those patterns to better judge each one.
- This helps make more consistent and accurate identity decisions.

### Evaluation Criteria:
- Facial features such as eyes, nose, mouth, and face shape
- Hair style and color (if visible)
- Accessories (e.g., glasses, earrings) if consistent across both images
- Contextual clues such as clothing or background, but only if the face is partially visible

### Output Instructions:
- Respond with a JSON array (no explanation, no markdown), where each object corresponds to one face crop, **in the exact order they are given** (excluding the reference).
- For each face crop, include:
  - "same_identity": true, false, or unsure
  - "justification": A brief explanation for your decision, including any uncertainties

### Output format:
[
  {{"same_identity": true, "justification": "Brief reason for your decision"}},
  {{"same_identity": false, "justification": "Brief reason for your decision"}},
  ...
]

Be conservative in high-confidence matches. Clearly explain uncertainty (e.g., blur, occlusion).

Images:
- First image: reference.jpg (do not evaluate this)
- Following images: [face_crop1.jpg to face_crop{n}.jpg] (evaluate these)
\end{lstlisting}
\end{tcolorbox}

\subsection{Contextual Recognition}
\label{appendix:recog_prompt}
\begin{tcolorbox}[
  enhanced,
  breakable,
  width=0.98\linewidth,
  colback=tzBlueFill,
  colframe=tzBlueBorder,
  boxrule=1.2pt,
  arc=6pt,
  left=5pt,right=5pt,top=4pt,bottom=2pt,
  title={\small First Step in Contextual Recognition},
  coltitle=white,
  colbacktitle=tzBlueHeader2,
  fonttitle=\bfseries,
]
\small
\begin{lstlisting}[style=jsonTiny]
You are an expert video understanding model.

You will be shown a short video clip where a single person is clearly highlighted using a green bounding box. This person is the main subject of interest.

### Your Task:
Your goal is to extract the following information strictly based on visual evidence inside the green bounding box and the visible background:

1. Action --- one fine-grained action the person is performing.
2. Outfit --- describe what the person is wearing (clothing type, color, accessories).
3. Scene --- briefly describe the background environment or setting (e.g., "kitchen", "forest trail", "office room").

If there is no bounding box visible throughout the video, return the message `"INVALID"` for all three fields.

### Requirements:
- Focus only on the person inside the green bounding box.
- Use precise and visually grounded descriptions.
- If the person is interacting with a visible item or person inside the bounding box, name the item or describe the outfit of the person specifically ---do not use vague terms like "object" or "item". (e.g., say "raising a suitcase" instead of "raising an object").
- Please focus on observable behavior, not inferred intentions or emotions. 
- If the person transitions between multiple actions, pick the most dominant or meaningful action visible in the clip.

### Output Format (JSON):
Return exactly one of the following:

#### If bounding box is visible:
{
  "action": "cutting vegetables",
  "outfit": "blue short-sleeve shirt and white apron",
  "scene": "modern kitchen"
}

#### If no bounding box is visible:
{
  "action": "INVALID",
  "outfit": "INVALID",
  "scene": "INVALID"
}

Return only the raw JSON object --- do NOT include any commentary, markdown, or explanation.
\end{lstlisting}
\end{tcolorbox}

\begin{tcolorbox}[
  enhanced,
  breakable,
  width=0.98\linewidth,
  colback=tzBlueFill,
  colframe=tzBlueBorder,
  boxrule=1.2pt,
  arc=6pt,
  left=5pt,right=5pt,top=4pt,bottom=2pt,
  title={\small Second Step in Contextual Recognition},
  coltitle=white,
  colbacktitle=tzBlueHeader2,
  fonttitle=\bfseries,
]
\small
\begin{lstlisting}[style=jsonTiny]
You are an expert video understanding assistant.

You will be given a video that includes the target entity consistently highlighted by green bounding boxes, along with summarized information about action, outfit, and scene changes involving a target entity.
Your task is to generate structured metadata based on the following criteria:
1. Determine whether the target entity shows significant action transitions only based on the provided action change description.
2. Determine whether the target entity shows significant outfit transitions only based on the provided outfit change description.
3. Determine whether the scene changes significantly involving the target entity only based on the provided outfit change description.
4. Determine whether any similar-looking entity (i.e., someone with a similar outfit) appears or not in the video, based on the provided video.
5. Describe the single action and outfit (e.g., clothes, color, accessories) of the other entities (not describe the changes) that are not highlighted by the bounding box, based on the provided video.
6. Describe whether the target entity shows over three action transitions (e.g., talking -> walking -> talking -> crying) --- only based on the provided action change description.
7. Describe whether the target entity shows over three outfit transitions (e.g., blue t-shirt -> white t-shirt -> pink dress -> black coat) --- only based on the provided outfit change description.
8. Describe whether the target entity shows over three scene transitions (e.g., church -> stadium -> park -> indoor room) --- only based on the provided scene change description.

In each case, return a binary decision as `true` or `false`, and provide a clear justification.
If there is only a single element in the changes, you should return `false` for the corresponding field.
If a similar-looking entity is present, also describe their outfit and actions in the justification.

Here are some examples of target entity information and generated structured metadata:
[Example 1] Target Entity Information in the video:
- Action Changes: ["sitting", "singing", "walking", "singing"]
- Outfit Changes: ["red shirt", "red striped shirt", "red shirt", "red shirt"]
- Scene Changes: ["park", "bench", "park", "park"]

[Example 1] Output Metadata:
```json
{{
  "significant_action_transition": true,
  "significant_scene_transition": false,
  "significant_outfit_transition": false,
  "similar_looking_existence": true,
  "justification": {{
    "significant_action_transition": "The target entity changes actions from sitting on the bench to singing, walking, and singing.",
    "significant_scene_transition": "The target entity remains in the park and bench, where the bench might be located in the park.",
    "significant_outfit_transition": "The target entity wears a red shirt and never changes outfits.",
    "similar_looking_existence": "There are other entities wearing red t-shirts and blue pants in the video."
  }},
  "other_entities": [
    {{
      "actions": "talking to someone sitting on the bench",
      "outfits": "red t-shirts and blue pants"
    }}, 
    {{
      "actions": "walking around the bench",
      "outfits": "yellow hat and green jacket"
    }}, 
  ],
  "over_three_action_changes": true,
  "over_three_outfit_changes": false,
  "over_three_scene_changes": false
}}
---
[Example 2] Target Entity Information in the Video:
- Action Changes: ["talking"]
- Outfit Changes: ["red shirt, black jeans, blue hat"]
- Scene Changes: ["beach"]

[Example 2] Output Metadata:
```json
{{
  "significant_action_transition": false,
  "significant_scene_transition": false,
  "significant_outfit_transition": false,
  "similar_looking_existence": false,
  "justification": {{
    "significant_action_transition": "The target entity only appears during one segment, where the given information has one action ("talking").",
    "significant_scene_transition": "The target entity only appears during one segment, where the given information has one scene ("beach").",
    "significant_outfit_transition": ""The target entity only appears during one segment, where the given information has one outfit ("red shirt, black jeans, and blue hat").",
    "similar_looking_existence": "There are no entities wearing a similar outfit to red shirt, black jeans, and blue hat in the video."
  }},
  "other_entities": [
    {{
      "actions": "crying",
      "outfits": "white dress and black shoes"
    }}
  ],
  "over_three_action_changes": false,
  "over_three_outfit_changes": false,
  "over_three_scene_changes": false
}}

### Output Format:
1. Return only the raw JSON object; do NOT include any commentary, markdown, or explanation.
2. Your output should follow the below structured format (JSON):
```json
{{
  "significant_action_transition": true or false,
  "significant_scene_transition": true or false,
  "significant_outfit_transition": true or false,
  "similar_looking_existence": true or false,
  "justification": {{
    "significant_action_transition": "Your explanation here.",
    "significant_scene_transition": "Your explanation here.",
    "significant_outfit_transition": "Your explanation here.",
    "similar_looking_existence": "If true, describe how the other entity looks and behaves. If false, justify why no such entity appears."
  }},
  "other_entities": [
    {{
      "actions": "Describe actions of other entities in the video.",
      "outfits": "Describe outfits of other entities in the video."
    }}
  ]
}}

Please generate structured metadata for the following target entity information in the video:
Target Entity Information in the Video:
- Action Changes: {action_changes}
- Outfit Changes: {outfit_changes}
- Scene Changes: {scene_changes}
\end{lstlisting}
\end{tcolorbox}

\section{Compositional Reasoning Progression}
\label{app_sec:taxonomy}

\begin{table}[h]
\centering
\footnotesize
\setlength{\tabcolsep}{4pt}
\renewcommand{\arraystretch}{1.15}
\begin{tabular}{p{0.28\linewidth} p{0.68\linewidth}}
\toprule
\textbf{Type} & \textbf{Definition} \\
\midrule
\multicolumn{2}{l}{\textbf{Level 1: Entity Existence; Evaluate the basic \textit{temporal continuity}}} \\
\midrule
Entity Existence & Track the \textit{existence} of an entity across appearance, disappearance, and reappearance over time.\\
\midrule
\multicolumn{2}{l}{
  \parbox{\linewidth}{
    \textbf{Level 2: Entity Changes; Evaluate integration of \textit{temporal continuity}}\\
    \textbf{and \textit{perceptual grounding}}
  }
} \\
\midrule
Action Changes & Track how an entity’s \textit{actions} evolve across time. \\
\cmidrule{1-2}
Outfit Changes & Track identity consistency despite \textit{outfit variations}. \\
\cmidrule{1-2}
Scene Changes & Track an entity across different \textit{scenes or contexts}.\\
\midrule
\multicolumn{2}{l}{
  \parbox{\linewidth}{
    \textbf{Level 3: Entity Ambiguity; Evaluate temporal continuity} \\
    \textbf{and \textit{fine-grained perceptual discrimination}}
  }
} \\
\midrule
Entity Ambiguity & Distinguish the target entity when there is a \textit{visually similar entity} or the target is \textit{partially occluded}. \\
\bottomrule
\end{tabular}
\caption{\textbf{Compositional Reasoning Progression.} Our Compositional Reasoning Progression (CRP) defines three ascending levels of narrative understanding: (1) basic entity existence, (2) dynamic state changes, and (3) ambiguity, which requires both temporal and perceptual reasoning.}
\label{tab:taxonomy}
\end{table}

\vspace{-0.2in}
\subsection{Template for QA Generation}
To automatically generate QA pairs whose answers are grounded in the extracted entity representations from our annotation pipeline, we design template-based questions aligned with the CRP defined for entity-centric narrative understanding (Table~\ref{tab:template_ee}-\ref{tab:template_ea}). For each reasoning dimension and temporal direction type, we construct five template variants, from which questions are randomly sampled per clip to ensure diversity, coverage, and factual correctness. 

For multiple-choice QA pairs, we select real distractors corresponding to other entities' attributes within the same clip and synthetic distractors sampled from different clips. For ordering questions, all options are drawn from the target entity's attributes to ensure temporal consistency. The ground-truth answer is always placed in option (a) for multiple-choice questions, and in the correct temporal order (a), (b), (c) for ordering questions. After generation, both question and answer distributions are distributed, where answers are randomly permuted, to achieve an approximately uniform spread across answer values and question types, minimizing potential sampling and positional biases.

\begin{table*}[p]
\centering

\scriptsize
\begin{tabularx}{\linewidth}{>{\raggedright\arraybackslash}p{0.11\linewidth} X}
\toprule
\textbf{Type} & \textbf{Definition \& Example}\\
\midrule
\multicolumn{2}{l}{\textbf{Question Type: Binary}} \\
\midrule
Appear & "Is the person [action] in [scene] with [outfit] only seen at the end?"
\newline "Does the person [action] and wearing [outfit] in [scene] first appear at the end?"
\newline "Is the person [action] and wearing [outfit] in [scene] not seen earlier but appears at the end?"
\newline "Does the person who is [action] and wearing [outfit] in [scene] appear only at the end?"
\newline "At the end, is the person [action] and in [scene] while wearing [outfit] seen for the first time?"\\
\midrule
Reappear (Start-to-later) & "Does the person [action] and wearing [outfit] in [scene] at the beginning disappear during the video and reappear at the end?"
\newline "Is the person [action] and wearing [outfit] in [scene] at the beginning gone for a while and then present again at the end?"
\newline "Does the person [action] and wearing [outfit] in [scene] at the beginning disappear and later show up at the end?"
\newline "Does the person [action] and wearing [outfit] in [scene] at the beginning appear again at the end?"
\newline "After appearing [action] and wearing [outfit] in [scene] at the beginning, does the person reappear at the end?"\\
\midrule
Reappear (Later-to-start) & "Does the person [action] and wearing [outfit] in [scene] at the end also appear at the beginning, and then disappear for a while?"
\newline "Does the person [action] and wearing [outfit] in [scene] at the end also appear at the beginning, and then go missing for a while?"
\newline "Does the person [action] and wearing [outfit] in [scene] at the end also show up earlier at the beginning?"
\newline "Does the person [action] and wearing [outfit] in [scene] at the end appear again at the beginning?"
\newline "After appearing [action] and wearing [outfit] in [scene] at the end, does the person reappear at the beginning?"\\
\midrule
Disappear & "Does the person who is [action] and wearing [outfit] in [scene] appear at the beginning and then remain unseen afterward?"
\newline "Is the person [action] in [scene] with [outfit] seen only at the beginning and then gone?"
\newline "Does the person [action] in [scene] with [outfit] disappear after the beginning and not come back?"
\newline "Is the person [action] and wearing [outfit] in [scene] only seen at the beginning and never again?"
\newline "Does the person who is [action] and wearing [outfit] in [scene] appear at the beginning, then leave and never come back?"\\
\midrule
\multicolumn{2}{l}{\textbf{Question Type: Multiple-choice}} \\
\midrule
Appear & "Which best describes the person {action} and wearing {outfit} in {scene} at the end? (a) Appear only at the end (b) Appear at the start, missing for a while, then back (c) Appear at the start, missing until the end (d) None of the above"
\newline "Which best describes the person in {scene} {action} and wearing {outfit} at the end? (a) Appear only at the end (b) Appear at the start, missing for a while, then back (c) Appear at the start, missing until the end (d) None of the above"
\newline "Which best describes the person {action} in {scene} while wearing {outfit} at the end? (a) Appears at the end (b) Appears at start, disappears, then back(c) Appears at start, disappears, and never back (d) None of the above"
\newline "When is the person appearing at the end while {action} and wearing {outfit} in {scene} seen? (a) Only at the end (b) At start and end with absence in between (c) At start only, then disappears (d) None of the above"
\newline "When is the person who appears at the end, while {action} in {scene} and wearing {outfit} visible? (a) Only at the end (b) At start and end with absence in between (c) At start only, then disappears (d) None of the above" \\
\midrule
Reappear (Start-to-later) & "Which best describes the person {action} and wearing {outfit} in {scene} at the beginning? (a) Appear only at the end (b) Appear at the start, missing for a while, then back (c) Appear at the start, missing until the end (d) None of the above"
\newline "Which best describes the person in {scene} {action} and wearing {outfit} at the beginning? (a) Appear only at the end (b) Appear at the start, missing for a while, then back (c) Appear at the start, missing until the end (d) None of the above"
\newline "Which best describes the person {action} in {scene} while wearing {outfit} at the beginning? (a) Appears at the end (b) Appears at start, disappears, then back(c) Appears at start, disappears, and never back (d) None of the above"
\newline "When is the person who appears at the beginning, while {action} and wearing {outfit} in {scene} seen? (a) Only at the end (b) At start and end with absence in between (c) At start only, then disappears (d) None of the above"
\newline "When is the person appearing at the beginning while {action} in {scene} and wearing {outfit} visible? (a) Only at the end (b) At start and end with absence in between (c) At start only, then disappears (d) None of the above" \\
\midrule
Reappear (Later-to-start) &"Which best describes the person {action} and wearing {outfit} in {scene} at the beginning? (a) Appear only at the end (b) Appear at the start, missing for a while, then back (c) Appear at the start, missing until the end (d) None of the above"
\newline "Which best describes the person in {scene} {action} and wearing {outfit} at the beginning? (a) Appear only at the end (b) Appear at the start, missing for a while, then back (c) Appear at the start, missing until the end (d) None of the above"
\newline "Which best describes the person who is {action} in {scene} while wearing {outfit} at the beginning? (a) Appears at the end (b) Appears at start, disappears, then back(c) Appears at start, disappears, and never back (d) None of the above"
\newline "When is the person who appears at the beginning while {action} and wearing {outfit} in {scene} visible? (a) Only at the end (b) At start and end with absence in between (c) At start only, then disappears (d) None of the above"
\newline "When is the person appearing at the beginning while {action} in {scene} and wearing {outfit} seen? (a) Only at the end (b) At start and end with absence in between (c) At start only, then disappears (d) None of the above"\\
\midrule
Disappear & "Which best describes the person {action} and wearing {outfit} in {scene} at the beginning? (a) Appear only at the end (b) Appear at the start, missing for a while, then back (c) Appear at the start, missing until the end (d) None of the above"
\newline "Which best describes the person in {scene} {action} and wearing {outfit} at the beginning? (a) Appear only at the end (b) Appear at the start, missing for a while, then back (c) Appear at the start, missing until the end (d) None of the above"
\newline "Which best describes the person who is {action} in {scene} while wearing {outfit} at the beginning? (a) Appears at the end (b) Appears at start, disappears, then back(c) Appears at start, disappears, and never back (d) None of the above"
\newline "When is the person appearing at the beginning with {action} and wearing {outfit} in {scene} visible? (a) Only at the end (b) At start and end with absence in between (c) At start only, then disappears (d) None of the above"
\newline "When is the person who appears at the beginning while {action} in {scene} and wearing {outfit} seen? (a) Only at the end (b) At start and end with absence in between (c) At start only, then disappears (d) None of the above"\\
\bottomrule
\caption{\textbf{Question Template of Entity Existence Dimension.}}
\label{tab:template_ee}
\end{tabularx}
\end{table*}

\begin{table*}[p]
\centering

\scriptsize
\begin{tabularx}{\linewidth}{>{\raggedright\arraybackslash}p{0.11\linewidth} X}
\toprule
\textbf{Type} & \textbf{Definition \& Example} \\
\midrule
\multicolumn{2}{l}{\textbf{Question Type: Binary}} \\
\midrule
Start-to-later & "Is the person [action1] in [scene] while wearing [outfit] at the beginning later performing [action2]?"
\newline "Is the person who is [action1] and wearing [outfit] in [scene] at the beginning later performing [action2]?"
\newline "Is the person who is [action1] and while wearing [outfit] in [scene] at the beginning later performing [action2]?"
\newline "At the beginning, the person [action1] in [scene] with [outfit] is visible — do they perform [action2] later?"
\newline "Does the person who is [action1] and wearing [outfit] in [scene] at the beginning later perform [action2]?"\\
\midrule
Later-to-start & "Is the person [action1] in [scene] with [outfit] at the end seen performing [action2] earlier?"
\newline "Is the person [action1] and wearing [outfit] in [scene] at the end, earlier performing [action2]?"
\newline "Is the person who is [action1] and with [outfit] in [scene] at the end seen performing [action2] earlier?"
\newline "At the end, the person [action1] in [scene] with [outfit] is visible — do they perform [action2] earlier?"
\newline "Does the person who is [action1] and wearing [outfit] in [scene] at the end perform [action2] earlier?"\\
\midrule
\multicolumn{2}{l}{\textbf{Question Type: Multiple-choice}} \\
\midrule
Start-to-later & "What action is the person [action] in [scene] with [outfit] at the beginning performing later? (a) [action1] (b) [action2] (c) [action3] (d) None of the above"
\newline "What action does the person who is [action] and wearing [outfit] in [scene] at the beginning perform later? (a) [action1] (b) [action2] (c) [action3] (d) None of the above"
\newline "What action is performed later by the person who is [action] and wearing [outfit] in [scene] at the beginning? (a) [action1] (b) [action2] (c) [action3] (d) None of the above"
\newline "At the beginning, the person [action] in [scene] with [outfit] is visible — what action do they perform later? (a) [action1] (b) [action2] (c) [action3] (d) None of the above"
\newline "What action does the person with [action] and [outfit] in [scene] at the beginning perform later? (a) [action1] (b) [action2] (c) [action3] (d) None of the above"\\
\midrule
Later-to-start & "What action does the person [action] in [scene] with [outfit] at the end perform earlier? (a) [action1] (b) [action2] (c) [action3] (d) None of the above"
\newline "What action does the person who is [action] and wearing [outfit] in [scene] at the end perform earlier? (a) [action1] (b) [action2] (c) [action3] (d) None of the above"
\newline "What action is performed earlier by the person who is [action] and wearing [outfit] in [scene] at the end? (a) [action1] (b) [action2] (c) [action3] (d) None of the above"
\newline "At the end, the person [action] in [scene] with [outfit] is visible — what action do they perform earlier? (a) [action1] (b) [action2] (c) [action3] (d) None of the above"
\newline "What action does the person with [action] and [outfit] in [scene] at the end perform earlier? (a) [action1] (b) [action2] (c) [action3] (d) None of the above" \\
\midrule
\multicolumn{2}{l}{\textbf{Question Type: Ordering}} \\
\midrule
Start-to-later & "What is the chronological order of the following actions performed by the person who was [action] and wearing [outfit] in [scene] at the beginning? (a) [action1] (b) [action2] (c) [action3] (d) None of the above"
\newline "Arrange the following actions in the order performed by the person seen [action] and wearing [outfit] in [scene] at the beginning. (a) [action1] (b) [action2] (c) [action3] (d) None of the above"
\newline "In what order did the person [action] and wearing [outfit] in [scene] at the beginning, perform these actions? (a) [action1] (b) [action2] (c) [action3] (d) None of the above"
\newline "Put the following actions in the order performed by the person [action] and wearing [outfit] in [scene] at the beginning. (a) [action1] (b) [action2] (c) [action3] (d) None of the above"
\newline "Identify the order of actions performed by the person [action] and wearing [outfit] in [scene] at the beginning. (a) [action1] (b) [action2] (c) [action3] (d) None of the above"\\
\midrule
Later-to-start & "What is the chronological order of the following actions performed by the person who was [action] and wearing [outfit] in [scene] at the end? (a) [action1] (b) [action2] (c) [action3] (d) None of the above"
\newline "Arrange the following actions in the order performed by the person seen [action] and wearing [outfit] in [scene] at the end. (a) [action1] (b) [action2] (c) [action3] (d) None of the above"
\newline "In what order did the person [action] and wearing [outfit] in [scene] at the end, perform these actions? (a) [action1] (b) [action2] (c) [action3] (d) None of the above"
\newline "Put the following actions in the order performed by the person [action] and wearing [outfit] in [scene] at the end. (a) [action1] (b) [action2] (c) [action3] (d) None of the above"
\newline "Identify the order of actions performed by the person [action] and wearing [outfit] in [scene] at the end. (a) [action1] (b) [action2] (c) [action3] (d) None of the above" \\
\midrule
Agnostic & "What is the chronological order of the following actions performed by the person who was [action] and wearing [outfit] in [scene] in the video? (a) [action1] (b) [action2] (c) [action3] (d) None of the above"
\newline "Arrange the following actions in the order performed by the person seen [action] and wearing [outfit] in [scene] in the video. (a) [action1] (b) [action2] (c) [action3] (d) None of the above"
\newline "In what order did the person [action] and wearing [outfit] in [scene] in the video, perform these actions? (a) [action1] (b) [action2] (c) [action3] (d) None of the above"
\newline "Put the following actions in the order performed by the person [action] and wearing [outfit] in [scene] in the video. (a) [action1] (b) [action2] (c) [action3] (d) None of the above"
\newline "Identify the order of actions performed by the person [action] and wearing [outfit] in [scene] in the video. (a) [action1] (b) [action2] (c) [action3] (d) None of the above" \\
\bottomrule
\caption{\textbf{Question Template of Entity Action Change Dimension.}}
\label{tab:template_ac}
\end{tabularx}
\end{table*}

\begin{table*}[p]
\centering

\scriptsize
\begin{tabularx}{\linewidth}{>{\raggedright\arraybackslash}p{0.11\linewidth} X}
\toprule
\textbf{Type} & \textbf{Definition \& Example} \\
\midrule
\multicolumn{2}{l}{\textbf{Question Type: Binary}} \\
\midrule
Start-to-later & "Is the person who is [action] and wearing [outfit1] in [scene] at the beginning later shown wearing [outfit2]?"
\newline "At the beginning, the person is [action] in [scene] with [outfit1] — do they wear [outfit2] later?"
\newline "Is the person [action] and wearing [outfit1] in [scene] at the beginning seen later wearing [outfit2]?"
\newline "After being in [scene] [action] and wearing [outfit1] at the beginning, is the person later seen wearing [outfit2]?"
\newline "Does the person [action] and wearing [outfit1] in [scene] at the beginning, later appear wearing [outfit2]?" \\
\midrule
Later-to-start & "Is the person who is [action] and wearing [outfit1] in [scene] at the end shown earlier wearing [outfit2]?"
\newline "At the end, the person is [action] in [scene] while wearing [outfit1] — do they wear [outfit2] earlier?"
\newline "Is the person [action] and wearing [outfit1] in [scene] at the end seen earlier in [outfit2]?"
\newline "Before being in [scene] [action] and wearing [outfit1] at the end, is the person seen earlier wearing [outfit2]?"
\newline "Does the person [action] and wearing [outfit1] in [scene] at the end appear earlier wearing [outfit2]?" \\
\midrule
\multicolumn{2}{l}{\textbf{Question Type: Multiple-choice}} \\
\midrule
Start-to-later & "What outfit is later worn by the person who is [action] and wearing [outfit] in [scene] at the beginning? (a) [outfit1] (b) [outfit2] (c) [outfit3] (d) None of the above"
\newline "At the beginning, the person is [action] in [scene] with [outfit] — what outfit does the person wear later? (a) [outfit1] (b) [outfit2] (c) [outfit3] (d) None of the above"
\newline "What outfit does the person [action] in [scene] while wearing [outfit] at the beginning wear later? (a) [outfit1] (b) [outfit2] (c) [outfit3] (d) None of the above"
\newline "After being seen [action] and wearing [outfit] in [scene] at the beginning, what outfit is the person later seen wearing? (a) [outfit1] (b) [outfit2] (c) [outfit3] (d) None of the above"
\newline "What outfit is later worn by the person who is [action] and wearing [outfit] in [scene] at the beginning? (a) [outfit1] (b) [outfit2] (c) [outfit3] (d) None of the above" \\
\midrule
Later-to-start & "What outfit is earlier worn by the person who is [action] and wearing [outfit] in [scene] at the end? (a) [outfit1] (b) [outfit2] (c) [outfit3] (d) None of the above"
\newline "At the end, the person is [action] in [scene] while wearing [outfit] — what outfit does the person wear earlier? (a) [outfit1] (b) [outfit2] (c) [outfit3] (d) None of the above"
\newline "What outfit is the person [action] in [scene] while wearing [outfit] at the end wearing earlier? (a) [outfit1] (b) [outfit2] (c) [outfit3] (d) None of the above"
\newline "Before being seen [action] and wearing [outfit] in [scene] at the end, what outfit is the person seen wearing earlier? (a) [outfit1] (b) [outfit2] (c) [outfit3] (d) None of the above"
\newline "What outfit is earlier worn by the person [action] and wearing [outfit] in [scene] at the end? (a) [outfit1] (b) [outfit2] (c) [outfit3] (d) None of the above"\\
\midrule
\multicolumn{2}{l}{\textbf{Question Type: Ordering}} \\
\midrule
Start-to-later & "What is the chronological order of the following outfits worn by the person who was [action] and wearing [outfit] in [scene] at the beginning? (a) [outfit1] (b) [outfit2] (c) [outfit3] (d) None of the above"
\newline "Arrange the following outfits in the order worn by the person [action] and wearing [outfit] in [scene] at the beginning. (a) [outfit1] (b) [outfit2] (c) [outfit3] (d) None of the above"
\newline "In what order did the person [action] and wearing [outfit] in [scene] at the beginning, wear the following outfits during the video? (a) [outfit1] (b) [outfit2] (c) [outfit3] (d) None of the above"
\newline "Put the following outfits in the order worn by the person who is [action] and wearing [outfit] in [scene] at the beginning. (a) [outfit1] (b) [outfit2] (c) [outfit3] (d) None of the above"
\newline "Identify the order of outfits worn by the person who is [action] and wearing [outfit] in [scene] at the beginning. (a) [outfit1] (b) [outfit2] (c) [outfit3] (d) None of the above" \\
\midrule
Later-to-start & "What is the chronological order of the following outfits worn by the person who was [action] and wearing [outfit] in [scene] at the end? (a) [outfit1] (b) [outfit2] (c) [outfit3] (d) None of the above"
\newline "Arrange the following outfits in the order worn by the person [action] and wearing [outfit] in [scene] at the end. (a) [outfit1] (b) [outfit2] (c) [outfit3] (d) None of the above"
\newline "In what order did the person [action] and wearing [outfit] in [scene] at the end wear the following outfits during the video? (a) [outfit1] (b) [outfit2] (c) [outfit3] (d) None of the above"
\newline "Put the following outfits in the order worn by the person who is [action] and wearing [outfit] in [scene] at the end. (a) [outfit1] (b) [outfit2] (c) [outfit3] (d) None of the above"
\newline "Identify the order of outfits worn by the person who is [action] and wearing [outfit] in [scene] at the end. (a) [outfit1] (b) [outfit2] (c) [outfit3] (d) None of the above" \\
\midrule
Agnostic & "What is the chronological order of the following outfits worn by the person who is [action] and wearing [outfit] in [scene] in the video? (a) [outfit1] (b) [outfit2] (c) [outfit3] (d) None of the above"
\newline "Arrange the following outfits in the order worn by the person who is [action] and wearing [outfit] in [scene] in the video. (a) [outfit1] (b) [outfit2] (c) [outfit3] (d) None of the above"
\newline "In what order did the person [action] and wearing [outfit] in [scene] in the video wear the following outfits? (a) [outfit1] (b) [outfit2] (c) [outfit3] (d) None of the above"
\newline "Put the following outfits in the order worn by the person who is [action] and wearing [outfit] in [scene] in the video. (a) [outfit1] (b) [outfit2] (c) [outfit3] (d) None of the above"
\newline "Identify the order of outfits worn by the person who is [action] and wearing [outfit] in [scene] in the video. (a) [outfit1] (b) [outfit2] (c) [outfit3] (d) None of the above"\\
\bottomrule
\caption{\textbf{Question Template of Entity Outfit Change Dimension.}}
\label{tab:template_oc}
\end{tabularx}
\end{table*}

\begin{table*}[p]
\centering

\scriptsize
\begin{tabularx}{\linewidth}{>{\raggedright\arraybackslash}p{0.11\linewidth} X}
\toprule
\textbf{Type} & \textbf{Definition \& Example} \\
\midrule
\multicolumn{2}{l}{\textbf{Question Type: Binary}} \\
\midrule
Start-to-later & Does the person who is [action] and wearing [outfit] in [scene1] at the beginning appear later in [scene2]?"
\newline "Is the person [action] and wearing [outfit] in [scene1] at the beginning later seen in [scene2]?"
\newline "Is the person [action] and in [scene1] while wearing [outfit] at the beginning later shown in [scene2]?"
\newline "After [action] and wearing [outfit] in [scene1] at the beginning, does the person later show up in [scene2]?"
\newline "Is the person [action] while wearing [outfit] in [scene1] at the beginning later present in [scene2]?" \\
\midrule
Later-to-start & "Does the person who is [action] and wearing [outfit] in [scene1] at the end appear earlier in [scene2]?"
\newline "Is the person [action] and wearing [outfit] in [scene1] at the end seen earlier in [scene2]?"
\newline "Is the person [action] and in [scene1] while wearing [outfit] at the end shown earlier in [scene2]?"
\newline "Before [action] and wearing [outfit] in [scene1] at the end, does the person appear in [scene2] earlier?"
\newline "Is the person [action] while wearing [outfit] in [scene1] at the end present earlier in [scene2]?"\\
\midrule
\multicolumn{2}{l}{\textbf{Question Type: Multiple-choice}} \\
\midrule
Start-to-later & "In which scene does the person who is [action] and wearing [outfit] in [scene] at the beginning appear later? (a) [scene1] (b) [scene2] (c) [scene3] (d) None of the above"
\newline "In which scene is the person [action] and wearing [outfit] in [scene] at the beginning seen later? (a) [scene1] (b) [scene2] (c) [scene3] (d) None of the above"
\newline "In which scene is the person [action] and in [scene] while wearing [outfit] at the beginning shown later? (a) [scene1] (b) [scene2] (c) [scene3] (d) None of the above"
\newline "After [action] and wearing [outfit] in [scene] at the beginning, in which scene does the person appear later? (a) [scene1] (b) [scene2] (c) [scene3] (d) None of the above"
\newline "In which scene is the person [action] while wearing [outfit] in [scene] at the beginning later present? (a) [scene1] (b) [scene2] (c) [scene3] (d) None of the above" \\
\midrule
Later-to-start & "In which scene does the person who is [action] and wearing [outfit] in [scene] at the end appear earlier? (a) [scene1] (b) [scene2] (c) [scene3] (d) None of the above"
\newline "In which scene is the person [action] and wearing [outfit] in [scene] at the end seen earlier? (a) [scene1] (b) [scene2] (c) [scene3] (d) None of the above"
\newline "In which scene is the person [action] and in [scene] while wearing [outfit] at the end shown earlier? (a) [scene1] (b) [scene2] (c) [scene3] (d) None of the above"
\newline "After [action] and wearing [outfit] in [scene] at the end, in which scene does the person appear earlier? (a) [scene1] (b) [scene2] (c) [scene3] (d) None of the above"
\newline "In which scene is the person [action] while wearing [outfit] in [scene] at the end present earlier? (a) [scene1] (b) [scene2] (c) [scene3] (d) None of the above"\\
\midrule
\multicolumn{2}{l}{\textbf{Question Type: Ordering}} \\
\midrule
Start-to-later & "What is the chronological order of the following scenes involving the person who was [action] and wearing [outfit] in [scene] at the beginning? (a) [scene1] (b) [scene2] (c) [scene3] (d) None of the above",
\newline "Arrange the following scenes in the order the person seen [action] and wearing [outfit] in [scene] at the beginning appears in them. (a) [scene1] (b) [scene2] (c) [scene3] (d) None of the above",
\newline "In what order did the person [action] and wearing [outfit] in [scene] at the beginning, move through these scenes? (a) [scene1] (b) [scene2] (c) [scene3] (d) None of the above",
\newline "Put the following scenes in the order in which the person seen [action] and wearing [outfit] in [scene] at the beginning appears. (a) [scene1] (b) [scene2] (c) [scene3] (d) None of the above",
\newline "Identify the order of scenes in which the person who is [action] and wearing [outfit] in [scene] at the beginning appears. (a) [scene1] (b) [scene2] (c) [scene3] (d) None of the above" \\
\midrule
Later-to-start & "What is the chronological order of the following scenes involving the person who was [action] and wearing [outfit] in [scene] at the end? (a) [scene1] (b) [scene2] (c) [scene3] (d) None of the above"
\newline "Arrange the following scenes in the order the person seen [action] and wearing [outfit] in [scene] at the end appears. (a) [scene1] (b) [scene2] (c) [scene3] (d) None of the above"
\newline "In what order did the person [action] and wearing [outfit] in [scene] at the end, move through these scenes? (a) [scene1] (b) [scene2] (c) [scene3] (d) None of the above"
\newline "Put the following scenes in the order the person seen [action] and wearing [outfit] in [scene] at the end appears. (a) [scene1] (b) [scene2] (c) [scene3] (d) None of the above"
\newline "Identify the order of scenes in which the person who is [action] and wearing [outfit] in [scene] at the end appears. (a) [scene1] (b) [scene2] (c) [scene3] (d) None of the above"\\
\midrule
Agnostic & "What is the chronological order of the following scenes involving the person who is [action] and wearing [outfit] in [scene] in the video? (a) [scene1] (b) [scene2] (c) [scene3] (d) None of the above"
\newline "Arrange the following scenes in the order in which the person seen [action] and wearing [outfit] in [scene] in the video appears. (a) [scene1] (b) [scene2] (c) [scene3] (d) None of the above"
\newline "In what order did the person [action] and wearing [outfit] in [scene] in the video, move through these scenes? (a) [scene1] (b) [scene2] (c) [scene3] (d) None of the above"
\newline "Put the following scenes in the order in which the person seen [action] and wearing [outfit] in [scene] in the video appears. (a) [scene1] (b) [scene2] (c) [scene3] (d) None of the above"
\newline "Identify the order of scenes in which the person who is [action] and wearing [outfit] in [scene] in the video appears. (a) [scene1] (b) [scene2] (c) [scene3] (d) None of the above"\\
\bottomrule
\caption{\textbf{Question Template of Entity Scene Change Dimension.}}
\label{tab:template_sc}
\end{tabularx}
\end{table*}

\begin{table*}[p]
\centering

\scriptsize
\begin{tabularx}{\linewidth}{>{\raggedright\arraybackslash}p{0.11\linewidth} X}
\toprule
\textbf{Type} & \textbf{Definition \& Example} \\
\midrule
\multicolumn{2}{l}{\textbf{Question Type: Binary}} \\
\midrule
Start-to-later & "Is the person [action1] and wearing [outfit1] in [scene1] at the beginning the same person seen later with [action2], [outfit2], and [scene2]?"
\newline "Does the person who is [action1] and wearing [outfit1] in [scene1] at the beginning match the same person seen later with [action2], [outfit2], and [scene2]?"
\newline "At the beginning, the person is [action1] in [scene1] with [outfit1] — is the same person seen later with [action2], [outfit2], and [scene2]?"
\newline "After appearing while [action1] and wearing [outfit1] in [scene1] at the beginning, is it the same person shown later with [action2], [outfit2], and [scene2]?"
\newline "Is the person [action1] in [scene1] while wearing [outfit1] at the beginning identical to the person later seen [action2] in [scene2] wearing [outfit2]?"\\
\midrule
Later-to-start & "Is the person who is [action1] and wearing [outfit1] in [scene1] at the end the same person seen earlier with [action2], [outfit2], and [scene2]?"
\newline "Does the person who is [action1] and wearing [outfit1] in [scene1] at the end match the same person seen earlier with [action2], [outfit2], and [scene2]?"
\newline "At the end, the person is [action1] in [scene1] with [outfit1] — is the same person seen earlier with [action2], [outfit2], and [scene2]?"
\newline "Before appearing [action1] and wearing [outfit1] in [scene1] at the end, is the same person shown earlier with [action2], [outfit2], and [scene2]?"
\newline "Is the person [action1] in [scene1] with [outfit1] at the end identical to the person earlier [action2] in [scene2] with [outfit2]?"\\
\midrule
\multicolumn{2}{l}{\textbf{Question Type: Multiple-choice}} \\
\midrule
Start-to-later & "Later in the video, which person is most likely the same one who was [action] and wearing [outfit] in the [scene] at the beginning? (a) [option1] (b) [option2] (c) [option3] (d) None of the above"
\newline "Later in the video, who is the same person who was [action] and wearing [outfit] in [scene] at the beginning? (a) [option1] (b) [option2] (c) [option3] (d) None of the above"
\newline "Which person seen later matches the one from the beginning who was [action] and wearing [outfit] in [scene]? (a) [option1] (b) [option2] (c) [option3] (d) None of the above"
\newline "Later in the video, who is the same person seen at the beginning while wearing [outfit] in [scene] [action]? (a) [option1] (b) [option2] (c) [option3] (d) None of the above"
\newline "Which person seen later matches the one seen at the beginning who was [action] and wearing [outfit] in the [scene]? (a) [option1] (b) [option2] (c) [option3] (d) None of the above"\\
\midrule
Later-to-start & "Earlier in the video, which person is most likely the same one who is [action] and wearing [outfit] in [scene] at the end? (a) [option1] (b) [option2] (c) [option3] (d) None of the above"
\newline "Earlier in the video, who is the same person who is [action] and wearing [outfit] in [scene] at the end? (a) [option1] (b) [option2] (c) [option3] (d) None of the above"
\newline "Which person seen earlier matches the one from the end who is [action] and wearing [outfit] in [scene]? (a) [option1] (b) [option2] (c) [option3] (d) None of the above"
\newline "Earlier in the video, who is the same person who was at the end while wearing [outfit] in [scene] [action]? (a) [option1] (b) [option2] (c) [option3] (d) None of the above"
\newline "Which person seen earlier matches the one seen at the end who is [action] and wearing [outfit] in the [scene]? (a) [option1] (b) [option2] (c) [option3] (d) None of the above"\\
\bottomrule
\caption{\textbf{Question Template of Entity Ambiguity Dimension.}}
\label{tab:template_ea}
\end{tabularx}
\end{table*}

\clearpage
\subsection{QA Filtering}
\label{sec:qa_quality}
To ensure QA quality, we apply additional verification using GPT-4o. We first refine question phrasing for grammatical correctness (\S\ref{sec:grammar}). For multiple-choice and ordering QA pairs with options, we apply a two-stage safeguard to maintain distractor quality: (1) option-level filtering based on pairwise similarity to remove duplicates or near-duplicates (\S\ref{sec:option}), and (2) a manual verification pass to ensure that synthetic distractors are plausible, visually confusable, and do not contain lexical or structural hints. This process ensures that the QA pairs genuinely probe the model's ability to differentiate between visually similar entities rather than exploit superficial cues.

\subsubsection{Grammar Check}
\label{sec:grammar}

\begin{tcolorbox}[
  enhanced,
  breakable,
  width=0.98\linewidth,
  colback=tzBlueFill,
  colframe=tzBlueBorder,
  boxrule=1.2pt,
  arc=6pt,
  left=5pt,right=5pt,top=4pt,bottom=2pt,
  title={\small Grammar Check},
  coltitle=white,
  colbacktitle=tzBlueHeader2,
  fonttitle=\bfseries,
]
\small
\begin{lstlisting}[style=jsonTiny]
You will be given a template and a question.  
Your task is to determine whether the question:  
1. Is grammatically correct.  
2. Is easy to understand.  

While doing this, ensure the question's intent remains consistent with the given template.  

Rules:  
- If the question is grammatically correct, set "grammar" to "yes".  
- If the question is not grammatically correct, set "grammar" to a corrected version that preserves its meaning.  
- If the question is easy to understand, set "understandable" to "yes".  
- If the question is not easy to understand, set "understandable" to a corrected version that is easier for the model to understand (without changing the meaning).  

Generate an answer in JSON format with the following fields:
```json 
{{
  "justification": "Brief explanation of the grammar correctness and understandability, including any changes made",
  "grammar": "yes" or "corrected question",
  "understandable": "yes" or "corrected question"
}}

### Template: {template}
### Question: {question}
\end{lstlisting}
\end{tcolorbox}



\subsubsection{Option Similarity Check}
\label{sec:option}
\begin{tcolorbox}[
  enhanced,
  breakable,
  width=0.98\linewidth,
  colback=tzBlueFill,
  colframe=tzBlueBorder,
  boxrule=1.2pt,
  arc=6pt,
  left=5pt,right=5pt,top=4pt,bottom=2pt,
  title={\small Option Similarity Check},
  coltitle=white,
  colbacktitle=tzBlueHeader2,
  fonttitle=\bfseries,
]
\small
\begin{lstlisting}[style=jsonTiny]
You will be given two options.  
Determine whether the two options are semantically similar or whether one option is a higher-level (more general or superset) of the other.
- If the two options are similar (e.g., office vs office with a picture), return "yes".
- If one option is a higher-level that is more general than the other (e.g., indoor room vs office), return "yes".
- If one option is a subset of the other (e.g., black jacket, red long sleeve button-up shirt, and light-colored pants vs red shirt and black jacket), return "yes".
- If two options have some overlapping elements but not exactly similar (e.g., The person entering a coffee shop and standing, wearing a black leather jacket over a dark shirt and jeans vs. the person entering a coffee shop and standing, wearing a red coat over a black top), return "no".
- If neither of the above cases applies, return "no".

Generate an answer in JSON format with the following fields:
```json
{{
  "justification": "Brief explanation describing which options are similar (if any) and why",
  "answer": "yes" or "no"
}}

### Option1: {option1}
### Option2: {option2}
\end{lstlisting}
\end{tcolorbox}

\section{Additional Experimental Results}
\subsection{Impact of Model Sizes}
We further analyze how scaling model size influences performance on \data. As shown in Table~\ref{tab:main_eval}, increasing model size generally improves performance within each model family. For instance, InternVL3-38B surpasses its 8B counterpart across nearly all question types and dimensions, except for ordering questions in the entity action changes dimension, achieving a 6.66\% average gain. This trend indicates that larger OGP-MLLMs capture richer multimodal correspondences and maintain more stable entity representations. In the video domain, Video-LLaMA2-72B outperforms the 7B variant by 5.66\% on average, suggesting that scaling can enhance temporal and perceptual grounding. However, LLaVA-NeXT-Video-34B slightly underperforms its 7B counterpart, revealing that larger parameter counts do not necessarily translate into better entity tracking capabilities. This inconsistency suggests that while scaling may improve general temporal reasoning, it remains insufficient to resolve the fine-grained, entity-centric grounding required by \data. Overall, even the largest OVS-MLLMs lag behind comparably sized OGP-MLLMs, implying that true progress in narrative understanding requires architectural or training-level advances, particularly those enforcing temporal alignment and identity consistency, beyond simple model scaling while preserving the perceptual grounding.

\subsection{Temporal Directional Bias}

\begin{figure}[h]
    \centering
    \includegraphics[width=0.4\linewidth]{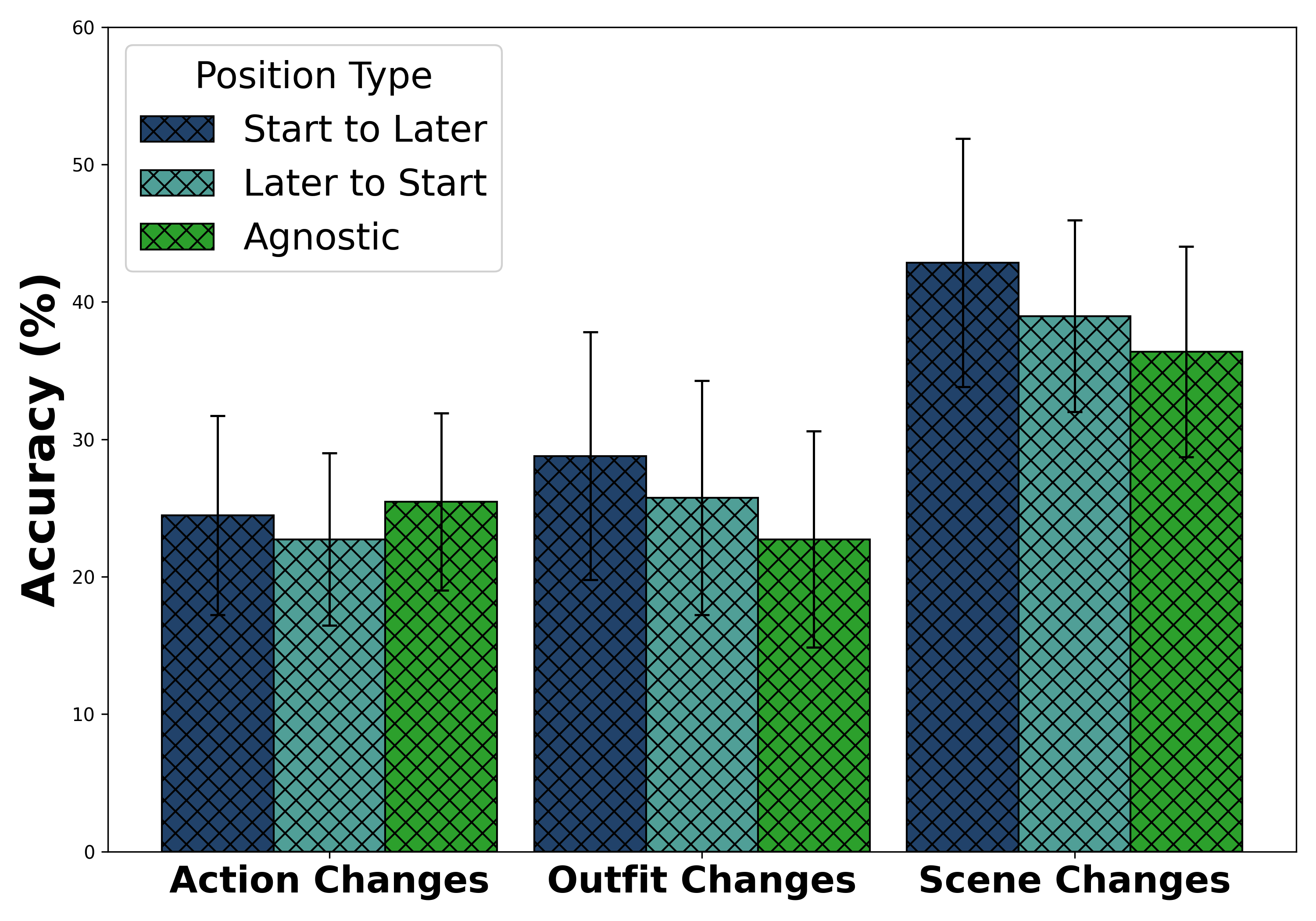}
    \caption{Temporal Directional Bias in Agnostic Reasoning.}
    \label{fig:pos_ordering_agnostic}
\end{figure}

We discussed the temporal directional bias in Section~\ref{sec:pos_bias}. In addition to forward and backward reasoning, we introduce an agnostic reasoning type for ordering questions, where the target entity is defined by its contextual attributes appearing in the middle of the video. The model must then chronologically arrange the entity's attributes from start to end, requiring bidirectional temporal understanding. Notably, models achieve the lowest performance on agnostic reasoning compared to forward and backward reasoning, particularly in outfit-change and scene-change dimensions that demand fine-grained perceptual grounding (Fig.~\ref{fig:pos_ordering_agnostic}). These results highlight that integrating temporal reasoning introduces a trade-off with perceptual precision: current MLLMs struggle to maintain both simultaneously, revealing a fundamental limitation in achieving narrative understanding that jointly requires temporal and perceptual reasoning.

\begin{figure*}[t]
    \centering
    \begin{subfigure}[t]{0.32\linewidth}
        \centering
        \includegraphics[width=\textwidth]{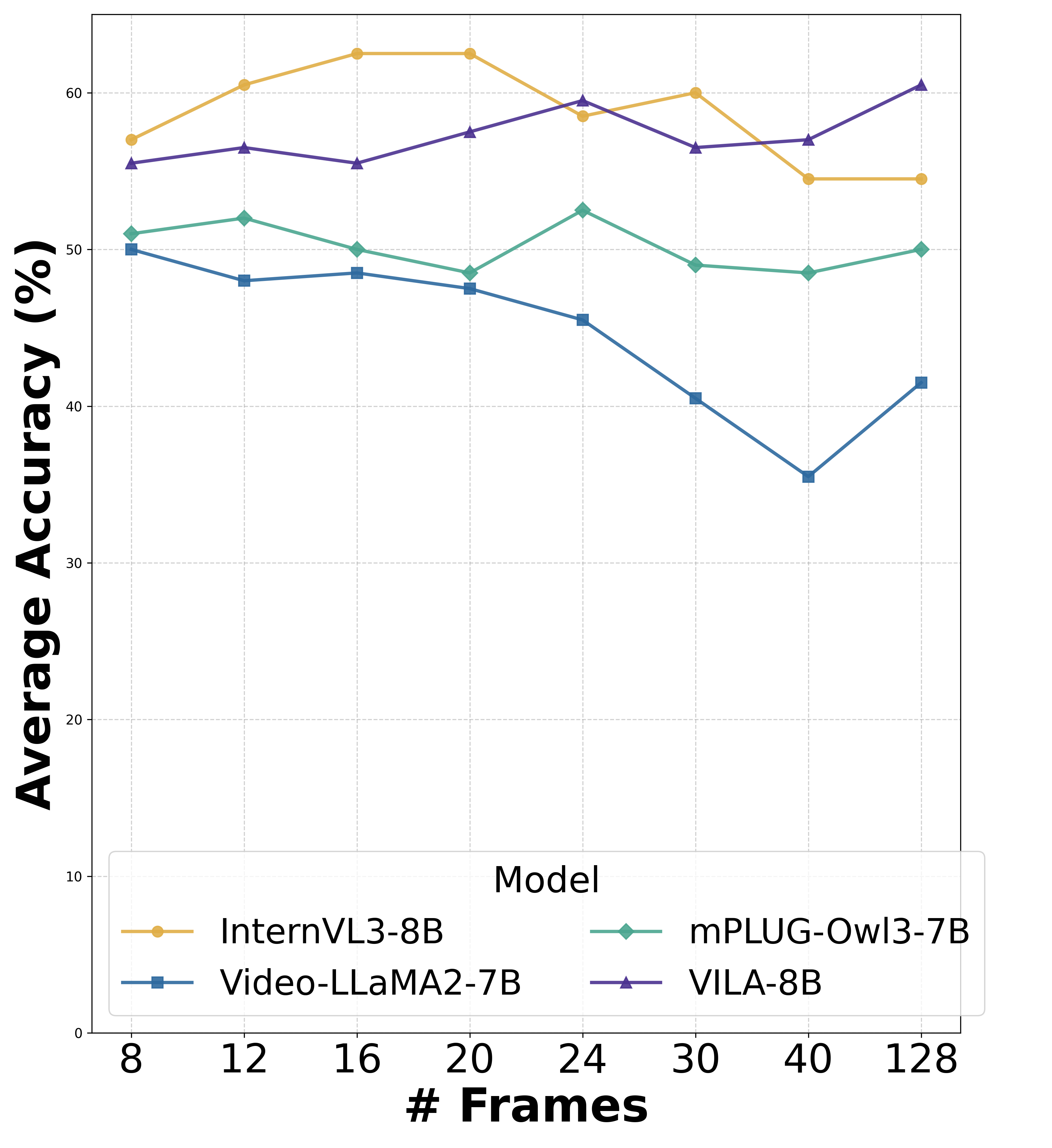}
        \caption{Entity Existence.}
        \label{fig:nf_ee}
    \end{subfigure}
    \begin{subfigure}[t]{0.32\linewidth}
        \centering
        \includegraphics[width=\textwidth]{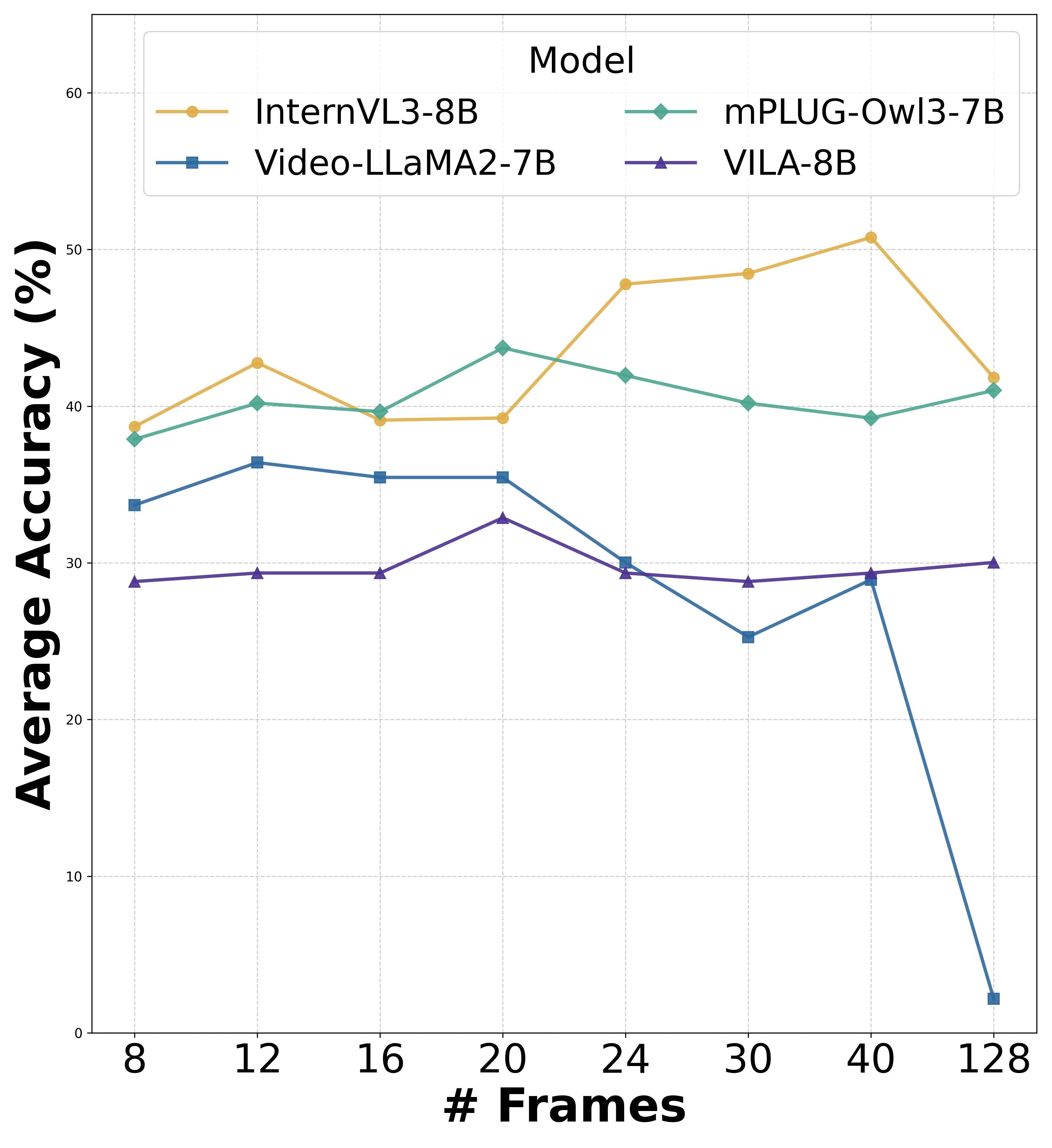}
        \caption{Entity Action Changes.}
        \label{fig:nf_ac}
    \end{subfigure}
    \begin{subfigure}[t]{0.32\linewidth}
        \centering
        \includegraphics[width=\textwidth]{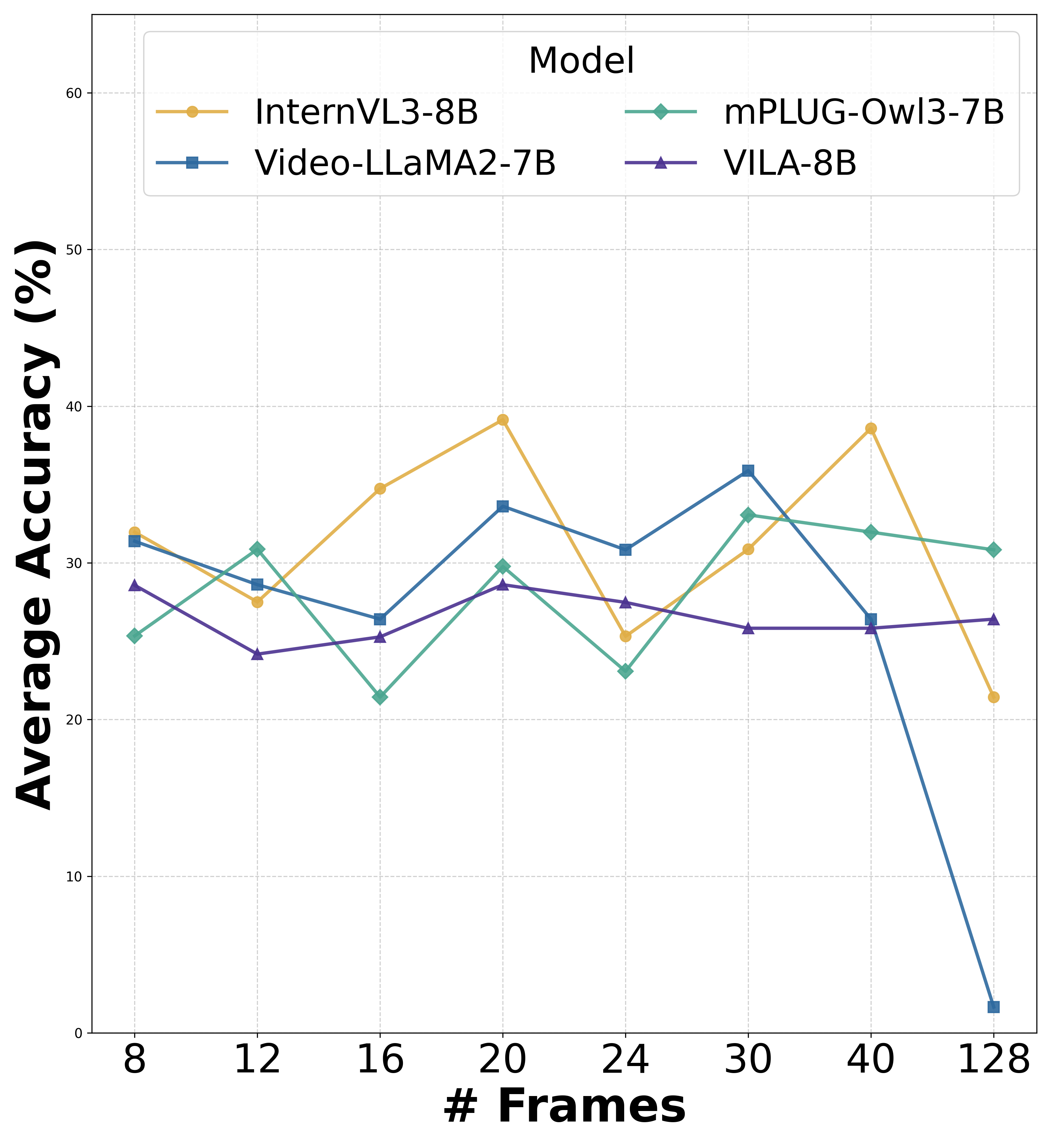}
        \caption{Entity Outfit Changes.}
        \label{fig:nf_oc}
    \end{subfigure}
    \begin{subfigure}[t]{0.32\linewidth}
        \centering
        \includegraphics[width=\textwidth]{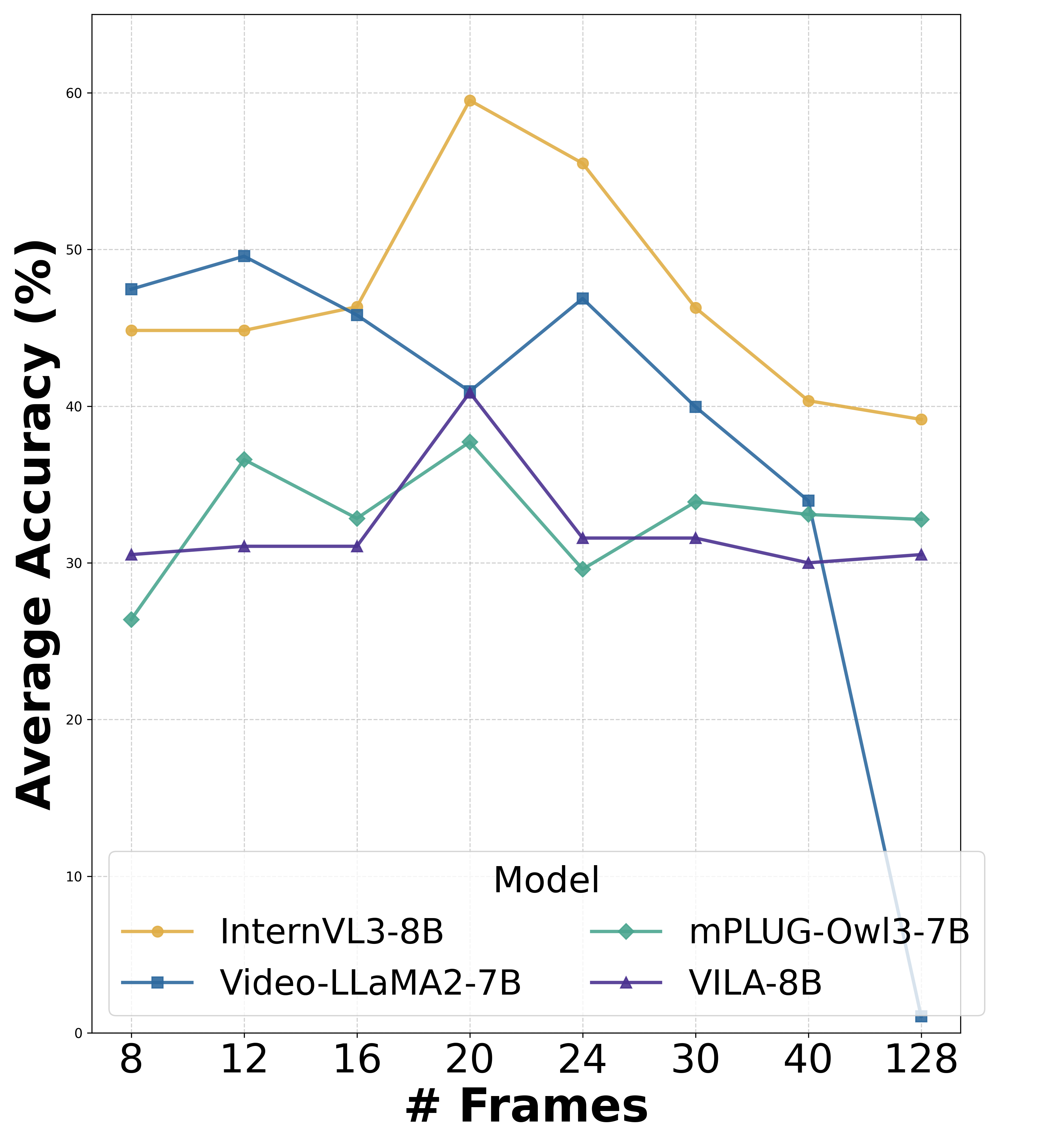}
        \caption{Entity Scene Changes.}
        \label{fig:nf_sc}
    \end{subfigure}
    \begin{subfigure}[t]{0.32\linewidth}
        \centering
        \includegraphics[width=\textwidth]{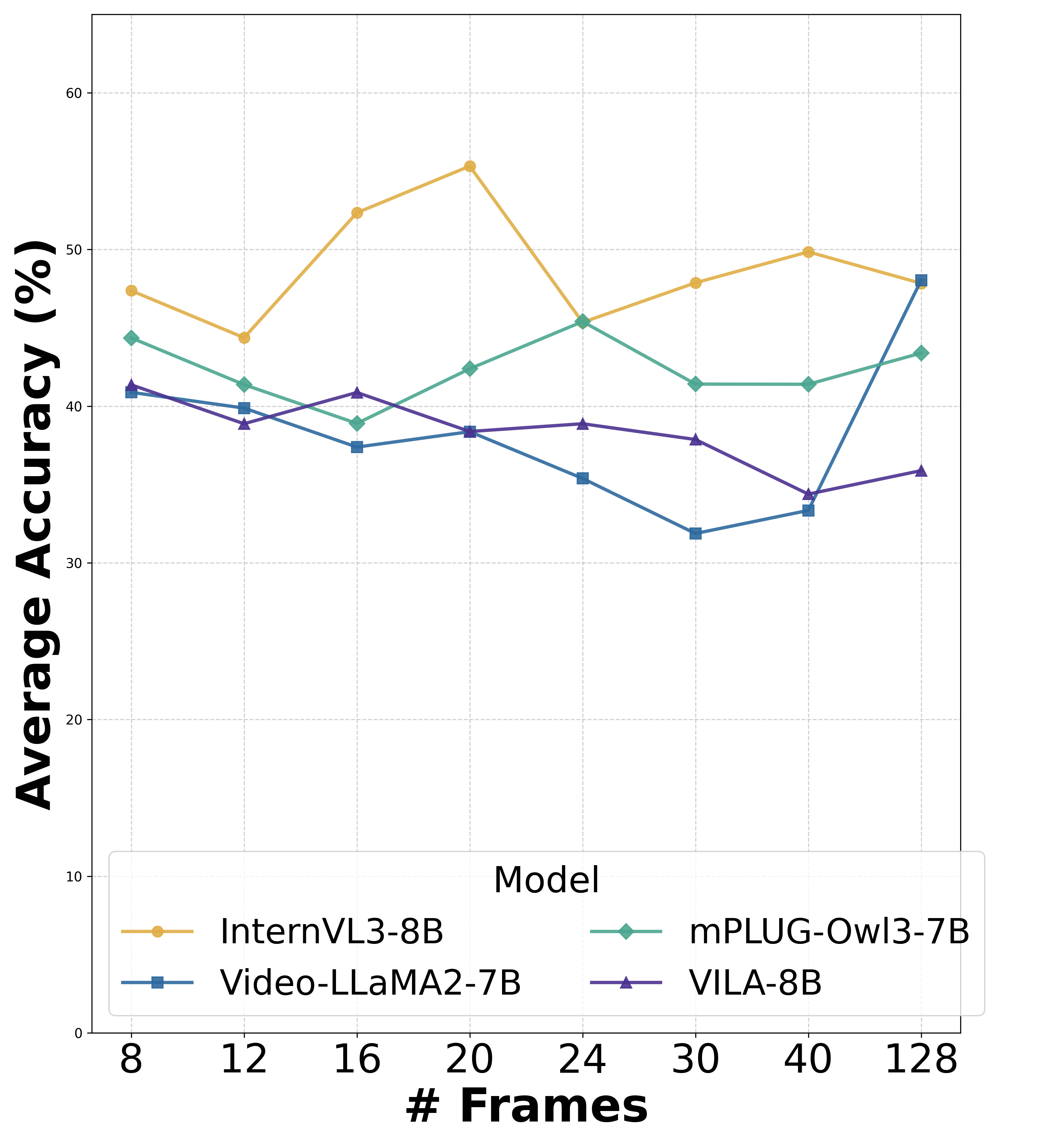}
        \caption{Entity Ambiguity.}
        \label{fig:nf_ea}
    \end{subfigure}
    \caption{Ablation study on frame density.}
    \label{fig:nf_app_dim}
\end{figure*}

\subsection{Ablation on Frame Density}
\label{subsec:abl_frame_density}
We further investigate the effect of frame density on reasoning performance across different dimensions in \data. For all models and reasoning types, performance tends to generally increase as the number of input frames grows, peaking around 20 frames, but drops sharply beyond this threshold (Fig.~\ref{fig:nf_app_dim}). This indicates that excessive frame sampling introduces redundant or noisy information, overwhelming the model's limited ability to capture temporal dependencies and fine-grained visual cues. The degradation suggests that current MLLMs struggle to integrate dense temporal information effectively, lacking mechanisms for selective attention and long-term temporal coherence. These findings emphasize that simply increasing frame density is insufficient; instead, entity-centric training objectives are needed to strengthen fine-grained perceptual grounding, and memory-augmented or recurrent architectures may be required to mitigate weaknesses in reverse and bidirectional reasoning relative to forward reasoning.

\subsection{Temporal-Perceptual Trade-offs}
To determine whether the observed continuity patterns are driven primarily by failures in basic entity perception, we conduct a model-specific controlled analysis conditioned on correct entity-existence predictions (Tab.~\ref{tab:real_vs_synth}). For each model, we first identify the subset of videos for which it correctly determines whether the target entity is present in the relevant frames. We then evaluate its accuracy on the remaining reasoning dimensions, including action, outfit, and scene changes, using only the corresponding subset, and compute the average performance across these dimensions. For example, if a model correctly answers 100 entity-existence questions, its controlled performance is measured using the videos associated with those 100 correct predictions. This conditioning reduces the influence of basic entity-detection errors and enables a more targeted evaluation of temporal reasoning given successful initial perceptual grounding.

Even under this controlled setting, OVS-MLLMs exhibit larger performance drops for both real (-11.5\%) and synthetic (-14.9\%) distractors, indicating that their errors cannot be attributed solely to failures in detecting the target entity. Instead, these models remain more susceptible to perceptual confusion and to selecting plausible but visually unsupported entities. At the same time, OVS-MLLMs remain competitive on tasks driven primarily by temporal dynamics, such as action changes, confirming a relative strength in temporal reasoning. Overall, these findings reveal a trade-off associated with model training objectives: OVS-MLLMs tend to capture temporal coherence more effectively but are more vulnerable to perceptual errors, whereas OGP-MLLMs generally provide stronger visual grounding but have greater difficulty integrating entity states over time.

\begin{table}[h]
\centering
\small
\resizebox{0.8\linewidth}{!}{
\begin{tabular}{lcccccccccc}
\toprule
\textbf{Model} & \textbf{Real (Avg.)} & \textbf{AC} & \textbf{OC} & \textbf{SC} & \textbf{EA}
& \textbf{Synthetic (Avg.)} & \textbf{AC} & \textbf{OC} & \textbf{SC} & \textbf{EA} \\
\midrule
OGP-MLLMs & 52.9 & 39.8 & 51.9 & 62.7 & 61.7 & 58.6& 61.8 & 85.7 & 54.7 & 50.0 \\
OVS-MLLMs & 41.4 & 38.3 & 36.3 & 42.0 & 51.7 & 43.7 & 41.2 & 37.5 & 45.7 & 45.5 \\
\midrule
$\Delta$ & 11.5 & 1.5 & 15.6 & 20.7 & 10.0 & 14.9 & 20.6 & 48.2 & 9.0 & 4.5 \\
\bottomrule
\end{tabular}
}
\caption{Isolating temporal errors with perceptual errors.}
\label{tab:real_vs_synth}
\end{table}

\subsection{True Temporal Reasoning in \data}

\begin{table*}[h]
    \centering
    \resizebox{\linewidth}{!}{%
        \begin{tabular}{l|cccccccccccccc}
            \toprule
            Model & \multicolumn{2}{c}{Existence} & \multicolumn{3}{c}{Action Changes} & \multicolumn{3}{c}{Outfit Changes} & \multicolumn{3}{c}{Scene  Changes}  & \multicolumn{2}{c}{Ambiguity} & Avg. \\
            \cmidrule(lr){2-3} \cmidrule(lr){4-6} \cmidrule(lr){7-9} \cmidrule(lr){10-12} \cmidrule(lr){13-14}
            & B & MC & B & MC & O & B & MC & O & B & MC & O & B & MC & \\
            \midrule
            Random & 50.00 & 25.00&50.00&25.00&16.67&50.00&25.00&16.67&50.00&25.00&16.67&50.00&25.00 &32.69 \\
            GPT-4o Text-Only & 57.00&27.72&53.26&29.21&19.51&50.55&32.53&0.00&55.79&23.29&30.44&57.00&27.72&41.75 \\
            GPT-4o Reverse Video & 73.00&53.00&70.65&58.43&4.88&83.52&56.63&11.11&74.74&61.64&4.35&90.00&55.45&62.92\\
            \midrule
            GPT-4o~\cite{achiam2023gpt}  &{77.00}&61.00&{82.61}&{66.29}&39.02&82.42&67.47&44.44&75.79&{83.56}&{78.26}&{86.00}&61.39&72.27\\
            \bottomrule
        \end{tabular}%
    }
    \caption{\textbf{Evaluation Results on \data}. B denotes binary, MC refers to the multiple-choice, and O indicates ordering questions.}
    \label{tab:baseline}
\end{table*}

Unlike existing benchmarks that can often be answered without visual inputs, our benchmark is explicitly designed to require true temporal reasoning, where questions cannot be solved without visual grounding or when input frames are shuffled. Because entity tracking inherently depends on referencing frames in their correct chronological order, models must integrate both temporal and perceptual cues. As shown in Table~\ref{tab:baseline}, GPT-4o exhibits a drastic performance drop when visual inputs are removed, approaching random guess accuracy, and performs substantially worse when video frames are reversed. These results demonstrate that \data effectively enforces temporally grounded reasoning and serves as a rigorous test of a model's ability to reason over time with perceptual grounding.

\subsection{Likelihood-based Evaluation}
\label{appendix:likelihood}

While we minimize generation bias by balancing answer distributions, we introduce a likelihood-based evaluation (LE) inspired by StrictVLE~\citep{saravanan2025velociti}. LE scores each answer choice by its conditional likelihood given the video $V$ and question $Q$: $e(V, Q, a_k)=p_M(a_k \mid I(V, Q)) / \sum_{j=1}^{K} p_M(a_j \mid I(V, Q))$, and select the choice $a_k$ with the highest score. Results in Table~\ref{tab:rebuttal_exp} show that LE yields lower overall accuracy, indicating that generation-based evaluation better reflects the model's reasoning behavior, while LE is more sensitive to uncertainty but limited to open-source models.
\begin{table}[h]
    \centering
    \resizebox{0.6\linewidth}{!}{%
        \begin{tabular}{l|cccccc}
            \toprule
            Model & EE & AC & OC & SC & EA & Avg. \\
            \midrule
            InternVL3-8B & 44.50 & 40.99 & 44.27& 48.17& 56.72 & 46.82 \\
            Video-LLaMA-7B & 41.50 & 32.43& 38.02& 38.74 & 39.80 & 37.97 \\
             \bottomrule
        \end{tabular}
    }
    \caption{Experimental results with likelihood-based evaluation on best-performing MLLMs.}
    \label{tab:rebuttal_exp}
\end{table}

\end{document}